\documentclass[runningheads]{llncs}

\usepackage[mobile]{eccv}

\usepackage{eccvabbrv}

\usepackage{graphicx}
\usepackage{booktabs}
\usepackage[accsupp]{axessibility}  

 \makeatletter
\def\@fnsymbol#1{\ensuremath{\ifcase#1\or \dagger\or \ddagger\or
  \mathsection\or \mathparagraph\or \|\or **\or \dagger\dagger
  \or \ddagger\ddagger \else\@ctrerr\fi}}
\makeatother

\usepackage{multirow}
\usepackage{amsmath,amssymb,lipsum}
\usepackage{mathrsfs}
\usepackage{enumerate}
\usepackage{colortbl}
\usepackage{enumitem}
\usepackage{wrapfig}
\usepackage{bbding}
\usepackage{pifont}
\usepackage{wasysym}
\usepackage{textcomp}
\usepackage{microtype}      
\usepackage[bottom]{footmisc}

\usepackage[misc]{ifsym} 

\usepackage{color}
\usepackage{tocloft}

\def\eg{{\it{e.g.}}}
\def\etal{{\it{et al.}}}
\def\ie{{\it{i.e.}}}

\def\reconone{{\scshape ReCon}}
\def\recon{{\scshape ReCon}{\raisebox{0.4ex}{\scriptsize \textbf{++}}}}
\def\shapellm{{\scshape ShapeLLM}}

\definecolor{drp-blue}{HTML}{1f77b4}
\definecolor{pretty-blue}{RGB}{0, 113, 188}
\definecolor{kaiming-green}{RGB}{57,181,74} 
\definecolor{icmlblue}{rgb}{0,0.08,0.45} 
\definecolor{linecolor}{gray}{.91} 
\definecolor{linecolor2}{gray}{.95} 
\definecolor{linecolor1}{gray}{.97} 
\definecolor{linecolor3}{HTML}{E3EFF7}
\definecolor{linecolor4}{HTML}{F1F7FB}

\definecolor{reconcolor}{HTML}{412F8A}
\definecolor{runpei_orange}{HTML}{F35F27}
\definecolor{runpei_blue}{HTML}{14294B}
\definecolor{datacolor}{HTML}{0009BF}
\definecolor{vitcolor}{HTML}{fc8e62}

\newcommand{\cmark}{\ding{51}}%
\newcommand{\xmark}{\ding{55}}%

\usepackage[
    breaklinks,
    colorlinks,
    linkcolor=blue, 
    urlcolor=blue,
    citecolor=blue,
]{hyperref}

\usepackage[capitalize]{cleveref}

\usepackage{orcidlink}

\begin{document}

\title{ShapeLLM: Universal 3D Object Understanding for Embodied Interaction}

\titlerunning{ShapeLLM: Universal 3D Object Understanding for Embodied Interaction}

\author{Zekun Qi\inst{12}\textsuperscript{\orcidlink{0009-0001-2554-5141}} \and
Runpei Dong\inst{12}\thanks{Project lead. \textsuperscript{(\Letter)}Corresponding authors. \\ \makebox[0.4cm]{} Work done during Z. Qi and R. Dong’s internships at MEGVII \& IIISCT.}\textsuperscript{\orcidlink{0000-0002-1104-7897}} \and
Shaochen Zhang\inst{1}\textsuperscript{\orcidlink{0009-0006-5550-905X}} \and
Haoran Geng\inst{3}\textsuperscript{\orcidlink{0009-0006-7828-3241}} \and\\
Chunrui Han\inst{4}\textsuperscript{\orcidlink{0000-0001-9725-280X}} \and
Zheng Ge\inst{4}\textsuperscript{\orcidlink{0000-0002-8630-8270}} \and
Li Yi\inst{567}\textsuperscript{(\Letter)}\textsuperscript{\orcidlink{0000-0002-9319-0354}} \and
Kaisheng Ma\inst{5}\textsuperscript{(\Letter)}\textsuperscript{\orcidlink{0000-0001-9226-3366}}
}

\authorrunning{Z.~Qi et al.}

\institute{Xi'an Jiaotong University\and
IIISCT \and
Peking University \and
MEGVII\\ \and
IIIS, Tsinghua University\and
Shanghai AI Laboratory\and
Shanghai Qi Zhi Institute\\
\url{https://qizekun.github.io/shapellm/}
}

\maketitle

\begin{center}
\includegraphics[width=1\linewidth]{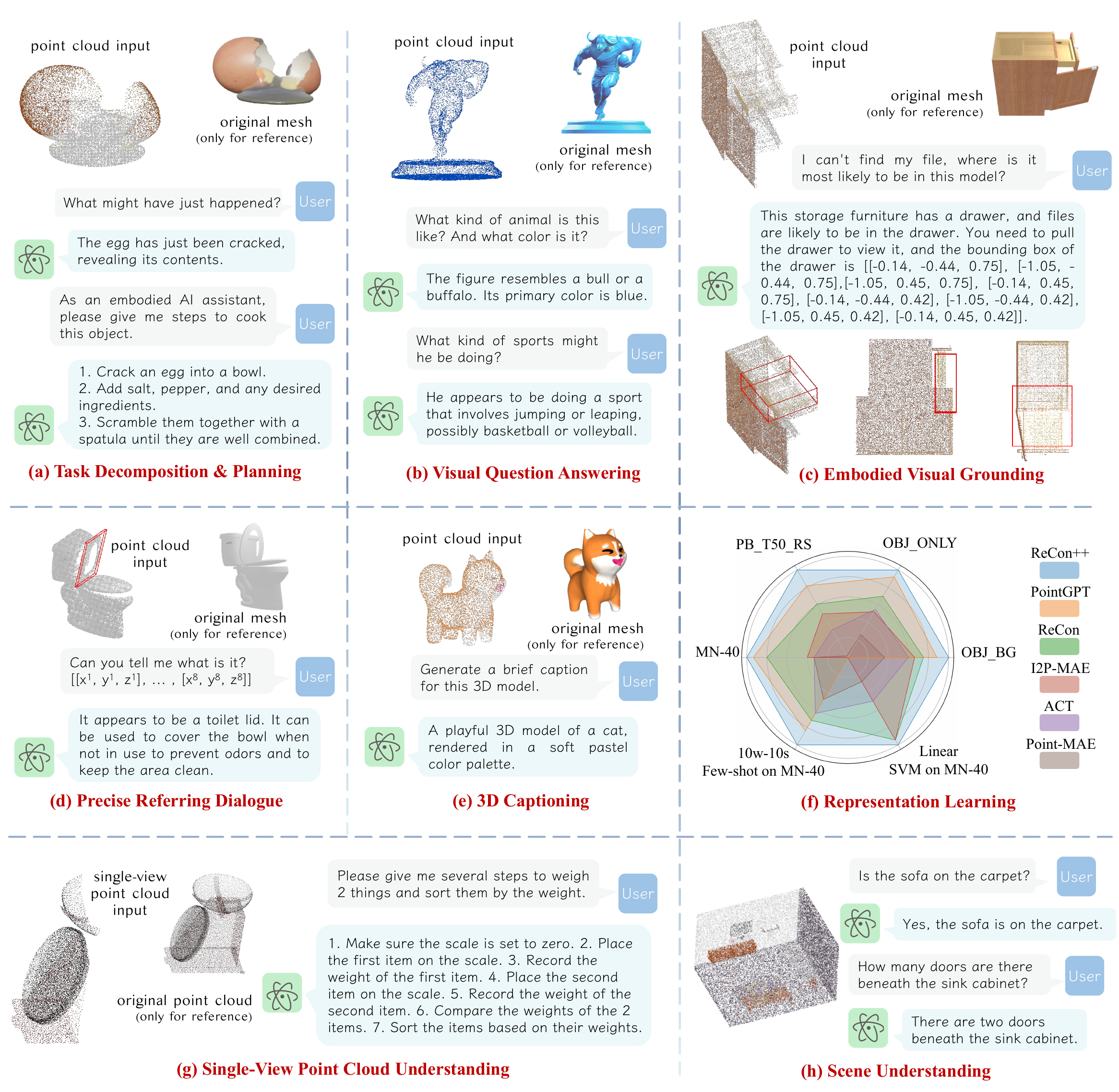}
\captionof{figure}{\textbf{Demonstrations of \shapellm
~and \recon.} We present \shapellm, the first 3D LLM designed for embodied interaction and spatial intelligence.
}
\label{fig:poster}
\end{center}

\begin{abstract}
This paper presents \shapellm, the first 3D Multimodal Large Language Model (LLM) designed for embodied interaction, exploring a universal 3D object understanding with 3D point clouds and languages.
\shapellm~is built upon an improved 3D encoder by extending \reconone~\cite{ReCon23} to \recon~that benefits from multi-view image distillation for enhanced geometry understanding.
By utilizing \recon~as the 3D point cloud input encoder for LLMs, \shapellm~is trained on constructed instruction-following data and tested on our newly human-curated benchmark, 3D MM-Vet.
\recon~and \shapellm~achieve state-of-the-art performance in 3D geometry understanding and language-unified 3D interaction tasks, such as embodied visual grounding.
\keywords{3D Point Clouds \and Large Language Models \and Embodied Intelligence \and 3D Representation Learning \and Zero-shot Learning}
\end{abstract}
\section{Introduction}\label{sec:intro}
3D shape understanding, serving as a fundamental capability for molding intelligent systems in both digital and physical worlds, has witnessed tremendous progress in graphics, vision, augmented reality, and embodied robotics.
However, to be effectively deployed by real-world agents, several critical criteria must be fulfilled:
\textbf{(i)} Sufficient 3D \textit{geometry} information needs to be captured for accurate spatial and structure processing~\cite{StereoMatching94,ShapeGoogle11,ShapeNet15,PointNet}.
\textbf{(ii)} Models should be endowed with a foundational knowledge of the \textit{embodied interaction} fashion with objects — often physically — for functional comprehension~\cite{WhatMakesAChair11,InteractionContext15,LearnObjectFunction16,SnapshotPartMobility17,Shape2Pose14,InteractionLandScapes17,BisectorInteraction3D14,CAMS23}.
\textbf{(iii)} A \textit{universal interface} is required as a bridge between information encoding and decoding, which could help translate high-order instructions for agent reactions like dialogue response and embodied feedback~\cite{VisDial19,ShareGPT23,EmbodiedReasoningLM22}.

Recent advancements in Large Language Models (LLMs)~\cite{GPT1_18,GPT2_19,GPT3_20,GPT4_23,LLaMA23} have demonstrated unprecedented success of foundational knowledge and unified reasoning capabilities across tasks~\cite{FoundationModel21,LMKnowledgeBase19,LLMKnowledge20,CommonSenseLM19,OnePolicyControlAll20,SayCan22,DreamLLM23,PaLME23,PALI-X23}.
It makes it possible to utilize language as a \textit{universal interface} that enables the comprehensive \textit{commonsense knowledge} embedded in LLMs to enhance understanding of 3D shapes. 
This is particularly evident in \textit{physically-grounded} tasks, where the wealth of commonsense knowledge simplifies the interpretation of an object's functionality, mobility, and dynamics, \etc.
However, the aforementioned challenges remain when incorporating LLMs for 3D object understanding — especially \textit{embodied interaction} that relies on precise \textit{geometry} — currently under-explored.

The question is: \textit{What makes better 3D representations that bridge language models and interaction-oriented 3D object understanding?} In this work, we introduce \shapellm~that meets the requirements, which is established based on the following three designing policies:
\begin{itemize}
    \item[\textbf{i.}] \textbf{3D Point Clouds as Inputs}~
    Some concurrent works~\cite{pointbind23} recently propose to use point cloud-rendered images~\cite{PointCLIP22} as multimodal LLMs' inputs and demonstrate effectiveness.
    However, these works fail to achieve accurate 3D geometry understanding and often suffer from a well-known visual hallucination issue~\cite{VisualHallucination18,POPE23,ZhouHall24}.
    Compared to 2D images, 3D point clouds provide a more accurate representation of the physical environment, encapsulating sparse yet highly precise geometric data~\cite{ACT23,RepGen3D18,PointNet++}.
    Moreover, 3D point clouds are crucial in facilitating embodied interactions necessitating accurate 3D structures like 6-DoF object pose estimation~\cite{NPCSCatPose20,NOCSPose19,UniDexGrasp23,UniDexGrasp23++23,CAPTRA21}. 
    \item[\textbf{ii.}] \textbf{Selective Multi-View Distillation}~
    Interacting with objects typically necessitates an intricate 3D understanding that involves knowledge at various levels and granularities.
    For instance, a whole-part \textit{high-level} semantic understanding is needed for interactions like opening a large cabinet, while detailed, \textit{high-resolution} (\ie, \textit{low-level}) semantics are crucial for smaller objects like manipulating a drawer handle~\cite{DeepPartInduction18}.
    However, existing works mainly distill single-view high-resolution object features from 2D foundation models~\cite{CLIP}, providing a complementary understanding~\cite{ACT23,ReCon23,ULIP22}.
    The potential of multi-view images, which offer abundant multi-level features due to view variation and geometry consistency~\cite{MV3DRec94,MVCNN3D15,StereoMatching94,MVTN,SyncDreamer23,Pri3D21}, is often neglected.
    \shapellm~extends \reconone~\cite{ReCon23} to \recon~ as the 3D encoder by integrating multi-view distillation. 
    To enable the model to selectively distill views that enhance optimization and generalization, inspired by DETR~\cite{DETR20}, \recon~ is optimized through adaptive selective matching using the Hungarian algorithm~\cite{HungarianAlogorithm55}.
    \item[\textbf{iii.}] \textbf{3D Visual Instruction Tuning}~
    Instruction tuning has been proven effective in improving LLMs' alignment capability~\cite{InstructGPT22,InstructGPT4_23}.
    To realize various 3D understanding tasks with a universal language interface, \shapellm~is trained through instruction-following tuning on constructed language-output data.
    However, similar to 2D visual instruction tuning~\cite{LLaVA23,LVM23}, the data-desert issue~\cite{ACT23} is even worse since no object-level VQA data is available, unlike 2D~\cite{LLaVA1.523}.
    To validate the efficacy of \shapellm, we first construct $\sim$45K instruction-following data using the advanced GPT-4V(ision)~\cite{GPT4Vision23} on the processed Objaverse dataset~\cite{objaverse23} and 30K embodied part understanding data from GAPartNet~\cite{gapartnet23} for supervised fine-tuning.
    Following MM-Vet~\cite{MMVet23}, we further develop a novel evaluation benchmark named 3D MM-Vet. 
    This benchmark is designed to assess the core vision-language capabilities, including embodied interaction in a 3D context, thereby stimulating future research.
    The 3D MM-Vet benchmark comprises 59 diverse Internet\footnote{\href{https://www.turbosquid.com/}{URL} \& \href{https://blog.turbosquid.com/turbosquid-3d-model-license/}{License}.} 3D objects and 232 human-written question-answer pairs.
    \begin{figure}[t!]
  \begin{center}
  \includegraphics[width=\linewidth]{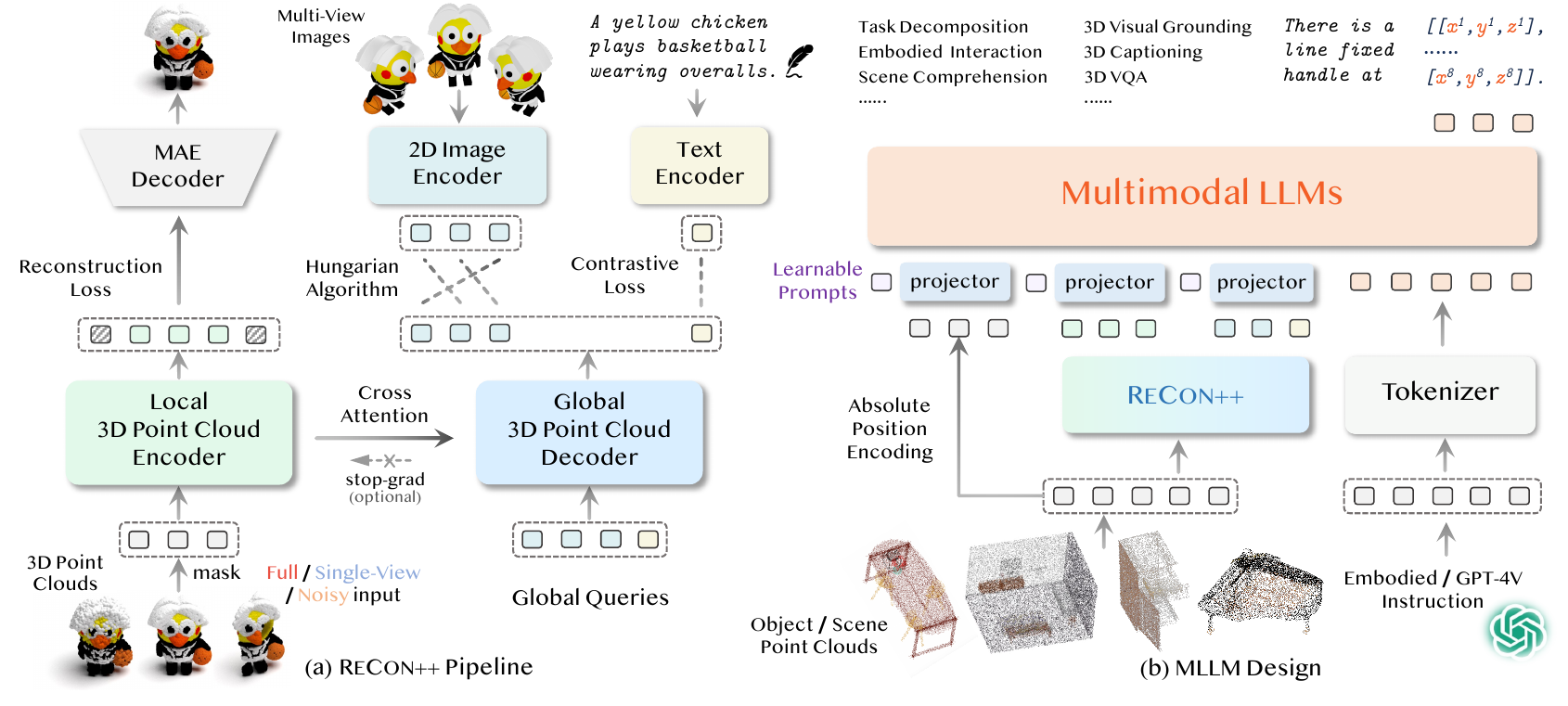}
  \vspace{-16pt}
  \caption{\textbf{Overview of our \shapellm\ framework}. 
  (a) The introduced \recon\ pipeline incorporates the required 3D encoder. (b) The comprehensive design of the MLLM, featuring an instruction-mode tokenizer and the integration of an aligned multi-modal representation, equips the MLLM with the capability to effectively handle 3D vision language tasks.}\label{fig:framework}
  \end{center}
  \vspace{-16pt}
\end{figure}

\end{itemize}

Through extensive experimentation, we first demonstrate that our improved 3D encoder \recon~sets a new state-of-the-art representation transferring on both downstream fine-tuned and zero-shot 3D object recognition.
Specifically, \recon~has obtained \textbf{95.25\%} and \textbf{95.0\%} fine-tuned accuracy on ScanObjectNN and ModelNet40,
surpassing previous best records by \textbf{+1.85\%} on the most challenging ScanObjectNN.
Besides, \recon~achieved \textbf{53.7\%} and \textbf{65.4\%} zero-shot accuracy on Objaverse-LVIS and ScanObjectNN, which is \textbf{+0.6\%} and \textbf{+1.6\%} higher than previous best.
By utilizing our \recon~as \shapellm's 3D encoder, \shapellm~successfully unifies various downstream tasks, including \textit{3D captioning}, \textit{3D VQA}, \textit{embodied task planning \& decomposition}, \textit{3D embodied visual grounding}, and \textit{3D precise referring dialogue} (See~\cref{fig:poster}).
On our newly constructed 3D MM-Vet benchmark, \textbf{42.7\%} and \textbf{49.3\%} Total accuracy have been achieved by \shapellm-7B and \shapellm-13B, surpassing previous best records~\cite{pointllm23} that also uses 3D point clouds by \textbf{+2.1\%} and \textbf{+5.1\%}, respectively.
This work initiates a first step towards leveraging LLMs for embodied object interaction, and we hope our \shapellm~and proposed 3D MM-Vet benchmark could spur more related future research.
\vspace{-2pt}
\section{\shapellm}\label{sec:method}
\vspace{-2pt}
In this section, we first introduce the overall architecture of \shapellm. Then, we delve into two critical challenges faced in interactive 3D understanding: data desert~\cite{ACT23} and representation of 3D point clouds. We present the detailed design of our method to tackle these challenges, respectively.

\vspace{-3pt}
\subsection{Overall Architecture}
\vspace{-2pt}
The main objective of this work is interactive 3D understanding by using the LLM as a universal interface. Drawing inspiration from recent work in visual understanding ~\cite{LLaVA23}, the proposed \shapellm~consists a pre-trained 3D encoder and an LLM for effective 3D representation learning and understanding, respectively. Specifically, we adopt LLaMA~\cite{LLaMA23} as our LLM, building upon the success of previous work ~\cite{LLaVA23,InstructBLIP23,DreamLLM23}. As for the 3D encoder, we propose a novel 3D model named \recon~based on the recent work \reconone~\cite{ReCon23} with multiple improvements as the 3D understanding generally demands more information, such as accurate spatial and multi-view details, etc. To ensure compatibility with the LLM inputs, the representation of a 3D object obtained from \recon~undergoes a linear projection before being fed into the LLM.
To further improve low-level geometry understanding, which benefits tasks like 6-DoF pose estimation, we append the absolute position encoding (APE) obtained by linear projection of 3D coordinates.
Besides, we use prefix-tuning with learnable prompts~\cite{VPT22,ACT23,DreamLLM23,PrefixTuning21} to adaptively modulate the different semantics of APE and \recon\ representations.
\begin{figure}[t!]
  \centering
  \subfloat[\textbf{Construction illustration of instruct-following data using GPT-4V~\cite{GPT4Vision23}.} Four perspective views are input into GPT-4V. In-context prompts focusing on different topics are explicitly incorporated to ensure data diversity.]{\includegraphics[width=0.48\textwidth]{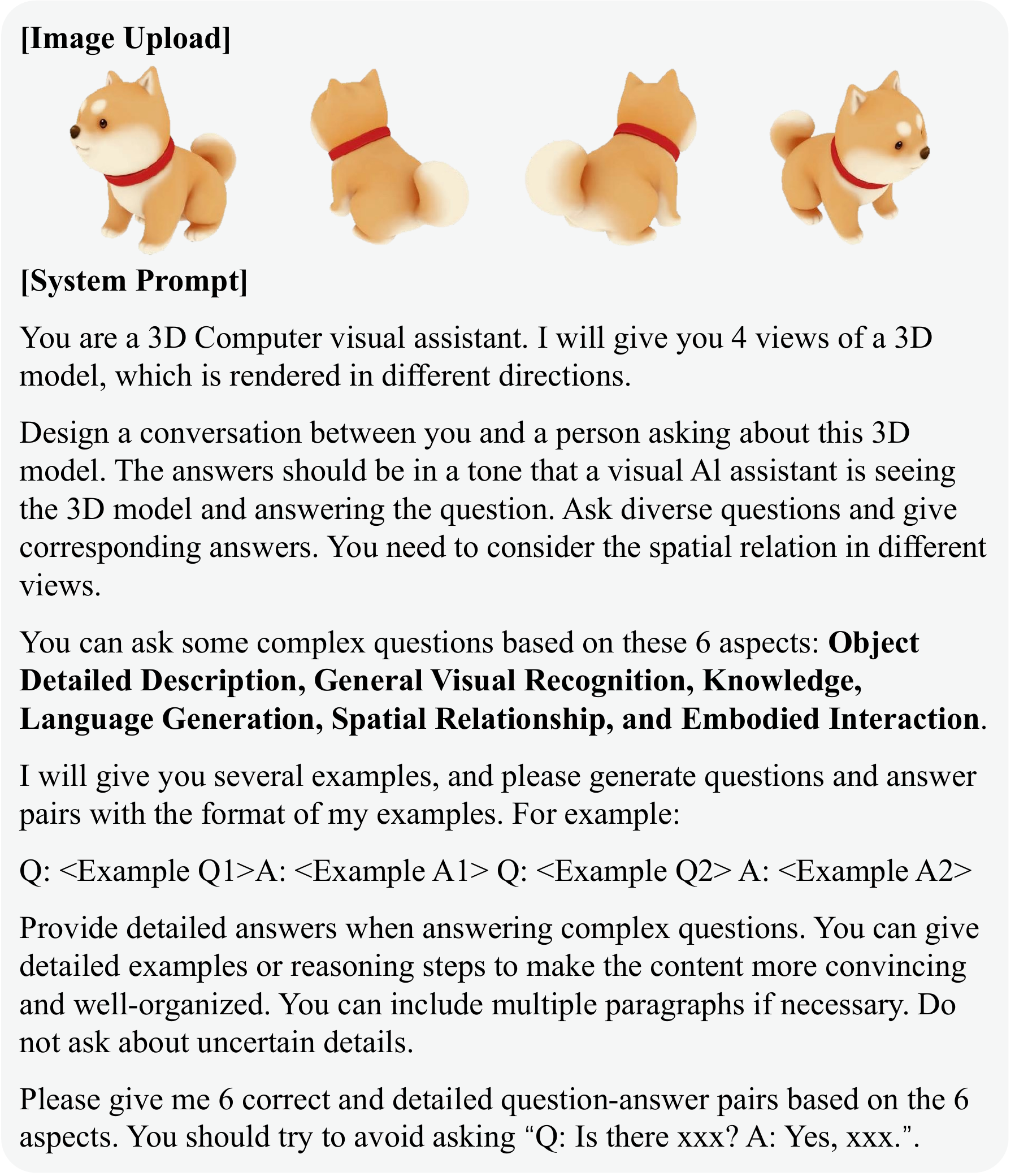}\label{fig:prompt}}
  \hfill
  \subfloat[\textbf{3D MM-Vet dataset sample.} A wealth of precise evaluation metrics enable a comprehensive assessment.]{\includegraphics[width=0.48\textwidth]{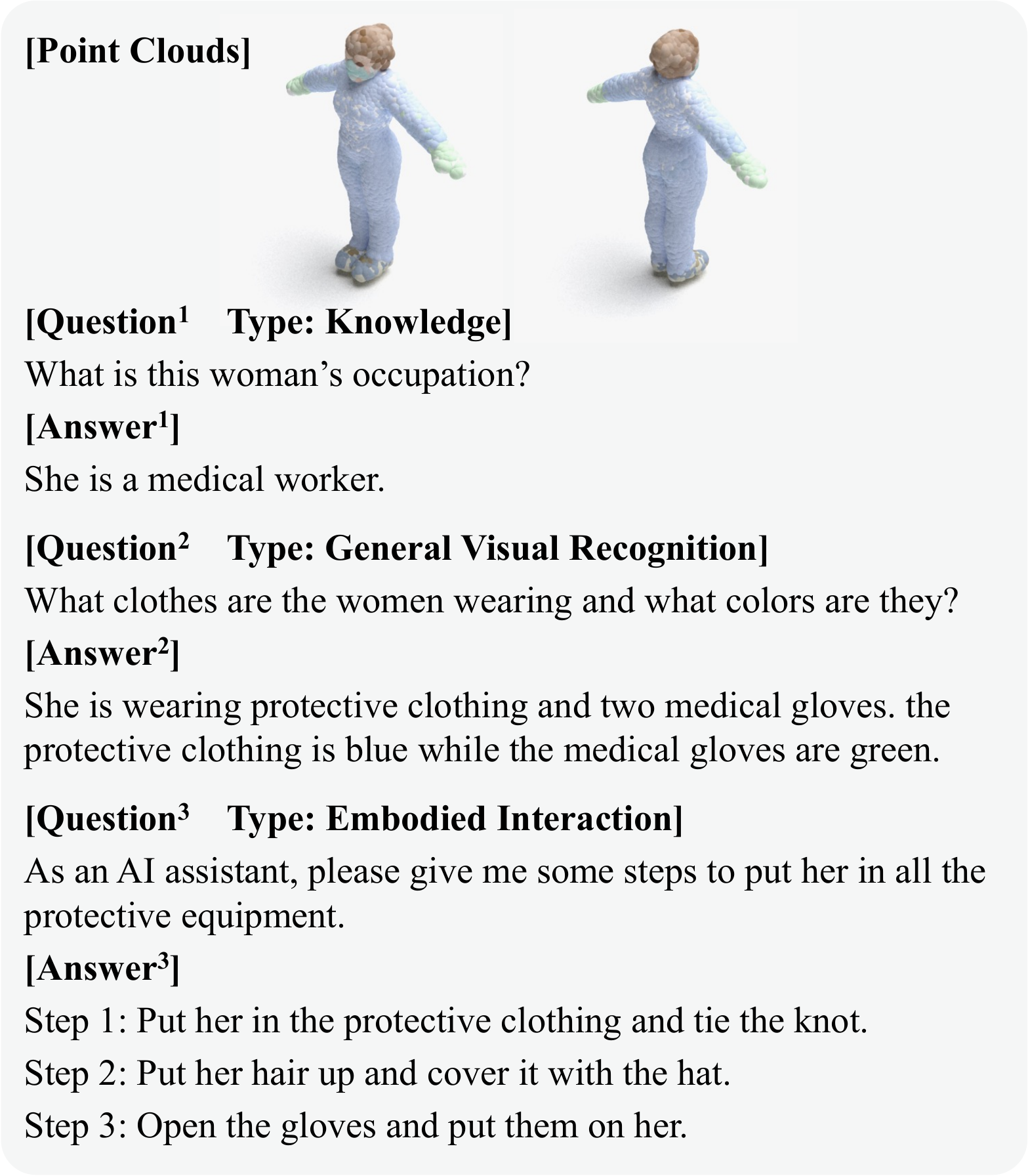}\label{fig:mmvet}}
  \vspace{-3pt}
  \caption{\textbf{Qualitative visualization} of the instruction-following and 3D MM-Vet data.}
  \label{fig:combined}
  \vspace{-10pt}
\end{figure}

\vspace{-3pt}
\subsection{How to alleviate interactive 3D understanding \emph{\textbf{Data Desert}}?}\label{sec:sft_data_gpt4v}
Most published 3D data is typically presented as 3D object-caption pairs, lacking an interactive style. Although a few concurrent works~\cite{pointllm23,3DLLM23} have attempted to construct interactive 3D understanding datasets, the questions-and-answers (Q\&As) are primarily based on annotated captions, often providing a limited perspective without sufficient details. Additionally, those works have generally been limited to semantic understanding without considering embodied interaction. To address these limitations, our work constructs question-and-answer pairs based on multi-view images of a 3D object using GPT-4V(ision)~\cite{GPT4Vision23}. For data diversity, we explicitly introduce six aspects as prompts, as illustrated \cref{fig:prompt}. 
In the following, we provide the details about data collection and construction regarding \textit{general semantic understanding} and \textit{embodied object understanding}, respectively.

\vspace{3pt}
\noindent\textbf{Data}~
Objaverse-LVIS~\cite{objaverse23,Cap3D23} and GAPartNet~\cite{gapartnet23} are data sources.
Objaverse-LVIS covers 1,156 LVIS~\cite{LVIS19} categories, and we sample Top-10 ``likes''\footnote{``Likes'' statistics can be found at \href{https://sketchfab.com/}{Sketchfab}.} 3D objects per category and generate Q\&A pairs per sample.
After filtering out noisy Q\&As, we obtain $\sim$45K instruction-following samples.
We use 12 categories from GAPartNet by removing ``Remote'' to avoid too many tiny boxes, which leads to filtered $\sim$30K Q\&A samples constructed from $\sim$8K parts of the $\sim$4K objects states covering $\sim$1.1K different objects.

\vspace{5pt}
\noindent\textbf{General Semantic Understanding}~
This aims to enhance the model's generalization abilities in visual recognition, knowledge integration, spatial understanding, and other aspects. We prompt GPT4-V to generate Q\&As in six different aspects based on images captured from four different views, as illustrated in \cref{fig:prompt}.
\begin{figure}[t]
\centering
\includegraphics[width=\linewidth]{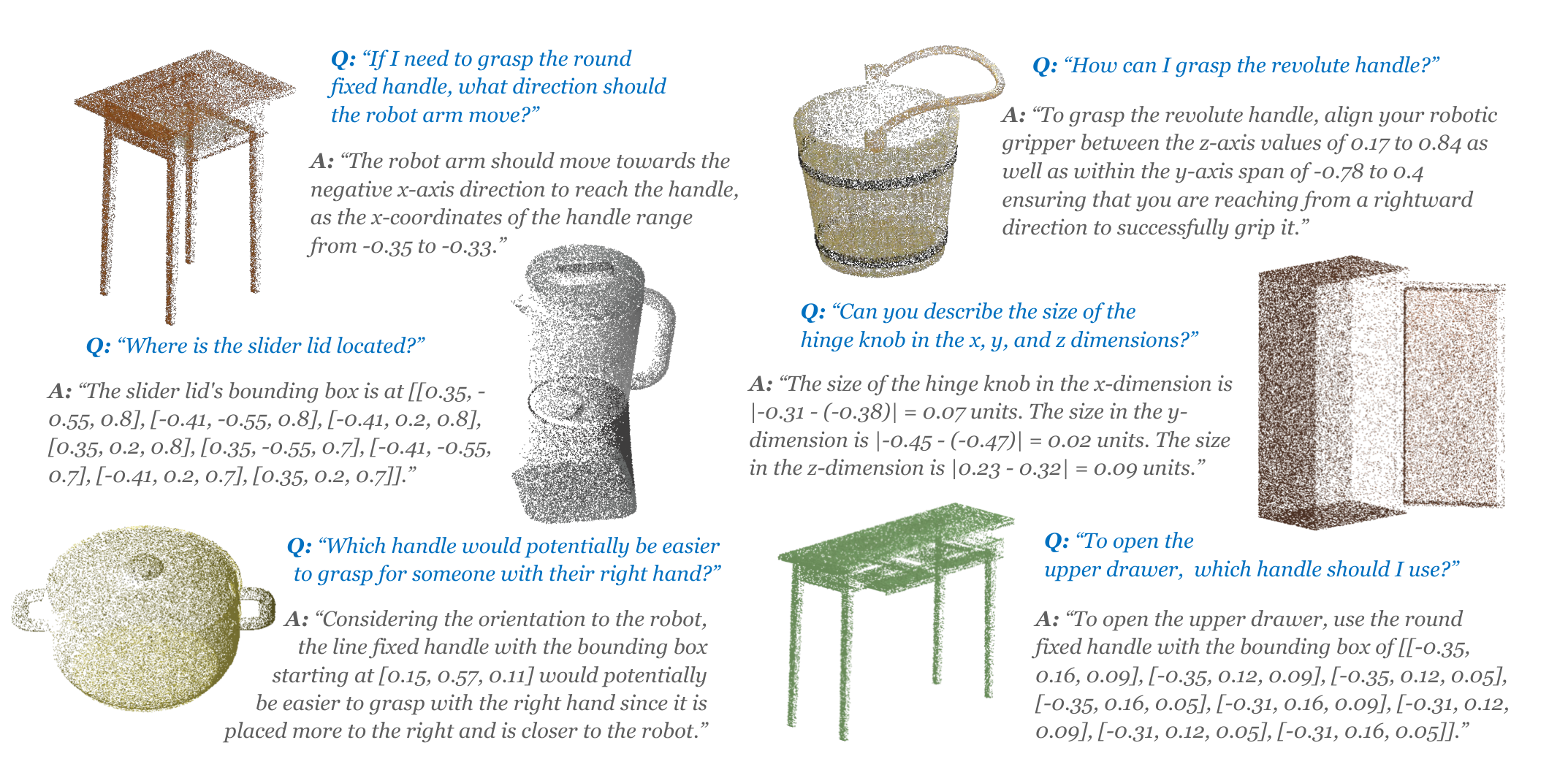}
\vspace{-20pt}
\captionof{figure}{\textbf{Qualitative examples} of the embodied interaction data.}
\vspace{-10pt}
\label{fig:gapartnet}
\end{figure}

\vspace{5pt}
\noindent\textbf{Embodied Object Understanding}~
A comprehensive understanding of the spatial positions and semantics at the part level is crucial to facilitate effective object grasping and interaction in embodied scenarios. Fortunately, the GAPartNet~\cite{gapartnet23} provides rich part annotations, including semantics and poses, which are instrumental in constructing instruction-tuning data for embodied interactive parts of a subject. Specifically, given a 3D object, questions are formulated based on the semantics of its different parts, and answers are constructed in both the semantics and 3D positions. The positions are represented as 6-DoF 3D bounding boxes in a straightened Python multidimensional list format, denoted as \textbf{\texttt{\textcolor{runpei_blue}{[[}\textcolor{runpei_orange}{x1}\textcolor{runpei_orange}{,} \textcolor{runpei_orange}{y1}\textcolor{runpei_orange}{,} \textcolor{runpei_orange}{z1}]\textcolor{runpei_blue}{, [}\textcolor{runpei_orange}{x2}\textcolor{runpei_orange}{,} \textcolor{runpei_orange}{y2}\textcolor{runpei_orange}{,} \textcolor{runpei_orange}{z2}]\textcolor{runpei_blue}{, \dots, [}\textcolor{runpei_orange}{x8}, \textcolor{runpei_orange}{y8}, \textcolor{runpei_orange}{z8}\textcolor{runpei_blue}{]]}}}, to meet characteristics of the textual dialogues response in LLMs. The canonical space of the object determines the sequence of coordinates. Using bounding box coordinates leverages the inherent spatial relationship, allowing LLMs to readily learn these patterns and generate accurate output coordinates. This approach can offer specific position information for embodied manipulation, as illustrated in \cref{fig:gapartnet}.

\vspace{-2pt}
\subsection{\textbf{\recon}: \emph{\textbf{Scaling Up}} 3D Representation Learning}
\vspace{-1pt}
Interaction with objects such as object grasping~\cite{UniDexGrasp23, UniDexGrasp23++23,GenOHDiffusion24} typically requires accurate perception of 3D shape information at multi-level and multi-granularity. This imposes heightened requirements on 3D representations, calling for a higher standard of a holistic understanding of 3D geometry.

However, existing 3D cross-modal representation learning methods~\cite{ULIP2_23,OpenShape23} mainly distill high-resolution object features from single-view 2D foundation models, resulting in a unilateral shape understanding. Besides, they generally employ multi-view images as data augmentation, imposing the learned representation to the average representation of all views. Thus, the accurate 3D shape information is missing. Recently, \reconone~\cite{ReCon23} utilizes contrast guided by reconstruction to address the pattern disparities between local masked data modeling and global cross-modal alignment. This results in remarkable performance in various tasks, including transfer learning, zero-shot classification, and part segmentation. However, its potential is hindered by the scarcity of pretraining data~\cite{ShapeNet15}.

To address the above limitations, this paper proposes \recon~with multiple improvements. First, multi-view image query tokens collaboratively comprehend the semantic information of 3D objects across different views, encompassing both RGB images and depth maps. 
Considering the disorderliness of pretraining data in terms of pose, we propose a cross-modal alignment method based on \textit{bipartite matching}, which implicitly learns the pose estimation of 3D objects. Second, we \textit{scale up} the parameters of \reconone~and broaden the scale of the pretraining dataset~\cite{objaverse23, Cap3D23, pointgpt23} for robust 3D representations.

Denote $N$ as the number of multi-view images, $I_{i}$ is the image feature from $i$-th view, and $Q_{i}$ represents the global query of $i$-th view. 
Following DETR~\cite{DETR20}, we search for an optimal permutation $\sigma$ of $N$ elements with the lowest cost:
\vspace{-3pt}
\begin{equation}
\label{eq:matching}
    \hat{\sigma} = \underset {\sigma} { \operatorname {arg\,min} }\sum_{i}^{N}\mathcal{L}_{\text{match}}(I_i, Q_{\sigma(i)}),
\end{equation}
where $\mathcal{L}_{\text{match}}(I_i, Q_{\sigma(i)})$ is a pair-wise matching cost between $i$-th view image features $I_i$ and matched query $Q_{\sigma(i)}$ with the permutation $\sigma$. 
In practice, we employ cosine similarity as the matching cost. 
In this fashion, the query of each view is learned to gather accurate 3D shape information from the 3D point clouds. 
Concatenating the features from the local 3D point cloud encoder and global 3D point cloud decoder together provides comprehensive information for 3D understanding of multimodal LLMs.

\vspace{-5pt}
\section{3D MM-Vet: Benchmarking 3D Comprehension}
\vspace{-5pt}
A wide range of diverse visual-language capabilities is essential to develop a multimodal large language model tailored for embodied scenarios, particularly addressing task and action planning. 

The model's proficiency in processing point clouds enables it to perform general recognition tasks effortlessly, demonstrating a broad understanding of colored point clouds. This capability serves as the groundwork for more intricate tasks. Beyond 3D recognition, the LLM should exhibit competence in addressing tasks in real-world embodied scenarios. This entails unifying the aforementioned abilities to generate decomposed task actions step-by-step in an instruction-following fashion, addressing specific problems.

\begin{table}[t!]
\setlength\tabcolsep{5pt}
\caption{
\textbf{Fine-tuned 3D recognition} on ScanObjectNN and ModelNet40.
Overall accuracy (\%) with voting~\cite{RSCNN} is reported.
$^\dagger$: Results with a post-pretraining stage~\cite{pointgpt23}. 
}\label{tab:cls}
\vspace{-16pt}
\begin{center}
\resizebox{0.82\linewidth}{!}{
\begin{tabular}{lcccccc}
\toprule[0.95pt]
\multirow{2}{*}[-0.5ex]{Method} & \multicolumn{3}{c}{\textbf{ScanObjectNN}} & \multicolumn{2}{c}{\textbf{ModelNet40}}\\
\cmidrule(lr){2-4}\cmidrule(lr){5-6} & OBJ\_BG & OBJ\_ONLY & PB\_T50\_RS & 1k P & 8k P\\
\midrule[0.6pt]
\multicolumn{6}{c}{\textit{Supervised Learning Only}}\\
\midrule[0.6pt]
PointNet~\cite{PointNet} &73.3 & 79.2 & 68.0 & 89.2 & 90.8\\
PointNet++~\cite{PointNet++} & 82.3 & 84.3 & 77.9 & 90.7 & 91.9\\
DGCNN~\cite{DGCNN} & 82.8 & 86.2 & 78.1 & 92.9 & -\\
PointMLP~\cite{PointMLP} & - & - & 85.4 & 94.5 & -\\
PointNeXt~\cite{PointNext} & - & - & 87.7 & 94.0 & -\\
\midrule[0.6pt]
\multicolumn{6}{c}{\textit{with Self-Supervised Representation Learning}}\\
\midrule[0.6pt]
Point-BERT~\cite{PointBERT} & 87.43 & 88.12 & 83.07 & 93.2 & 93.8\\
Point-MAE~\cite{PointMAE} & 90.02 & 88.29 & 85.18 & 93.8 & 94.0\\
Point-M2AE~\cite{PointM2AE22} & 91.22 & 88.81 & 86.43 & 94.0 & -\\
Point2Vec~\cite{point2vec23} & 91.2 & 90.4 & 87.5 & 94.8 & -\\
ACT~\cite{ACT23} & 93.29 & 91.91 & 88.21 & 93.7 & 94.0\\
TAP ~\cite{takeaphoto23} & - & - & 88.5 & 94.0 & -\\
VPP~\cite{VPP23} & 93.11 & 91.91 & 89.28 & 94.1 & 94.3\\
I2P-MAE~\cite{I2PMAE23} & 94.15 & 91.57 & 90.11 & 94.1 & -\\
ULIP-2~\cite{ULIP2_23} & - & - & 91.5 & - & -\\
\reconone~\cite{ReCon23} & 95.35 & 93.80 & 91.26 & 94.5 & 94.7\\
PointGPT-B$^\dagger$~\cite{pointgpt23} & 95.8 & 95.2 & 91.9 & 94.4 & 94.6 \\
PointGPT-L$^\dagger$~\cite{pointgpt23} & 97.2 & 96.6 & 93.4 & 94.7 & 94.9 \\
\rowcolor{linecolor4}\textbf{\recon-B}$^\dagger$ & \textbf{98.62} & \textbf{96.21} & \textbf{93.34} & \textbf{94.6} & \textbf{94.8}\\
\rowcolor{linecolor3}\textbf{\recon-L}$^\dagger$ & \textbf{98.80} & \textbf{97.59} & \textbf{95.25} & \textbf{94.8} & \textbf{95.0}\\
\bottomrule[0.95pt]
\end{tabular}
}
\end{center}
\vspace{-22pt}
\end{table}

Hence, to formulate an evaluation system aligned with the aforementioned task description, we establish a multi-level evaluation task system encompassing four-level tasks: \textbf{General Recognition}, \textbf{Knowledge and Language Generation}, \textbf{Spatial Awareness}, and \textbf{Embodied Interaction}. This framework systematically and comprehensively assesses the model's proficiency in information comprehension and language generation when processing interactive objects. The detailed descriptions of the tasks are listed as follows:
\begin{itemize}
    \item[\textbf{i}.] \textbf{General Recognition:} Following MM-Vet~\cite{MMVet23}, we assess the fundamental comprehension abilities of LLMs involving both coarse- and fine-grained aspects. Coarse-grained recognition focuses on basic object attributes such as color, shape, action, \etc.
    While fine-grained recognition delves into details like subparts and counting, \etc.

    \item[\textbf{ii}.] \textbf{Knowledge Capability \& Language Generation:} To examine the models' capacity to understand and utilize knowledge, drawing inspiration from MMBench~\cite{MMBench23}, we integrate its reasoning components. This includes knowledge spanning natural and social reasoning, physical properties, sequential prediction, math, \etc, evaluating gauges whether multimodal LLMs possess the requisite expertise and capacity to solve intricate tasks. We utilize customized prompts to stimulate models and extract detailed responses to evaluate language generation.
    
    \item[\textbf{iii}.] \textbf{Spatial Awareness:} In 3D, spatial awareness holds heightened significance compared to 2D due to the provided geometry information. The point clouds contain location information crucial for discerning spatial relationships between different parts. In 2D, achieving the same information intensity level would necessitate multi-view images. Therefore, our evaluation includes questions probing the ability of LLMs to understand spatial relations.

    \item[\textbf{iv}.] \textbf{Embodied Interaction:} The utilization scope of MLLMs extends into the field of embodied interaction, facilitated by the utilization of instruction-following data. Our evaluation system tests their capacity by formally requesting LLMs to provide execution steps toward an instruction. This approach aims to establish connections for handling Embodied Interaction tasks~\cite{OnePolicyControlAll20,PaLME23}.
\end{itemize}

 \begin{figure}[t!]
\centering
\includegraphics[width=\linewidth]{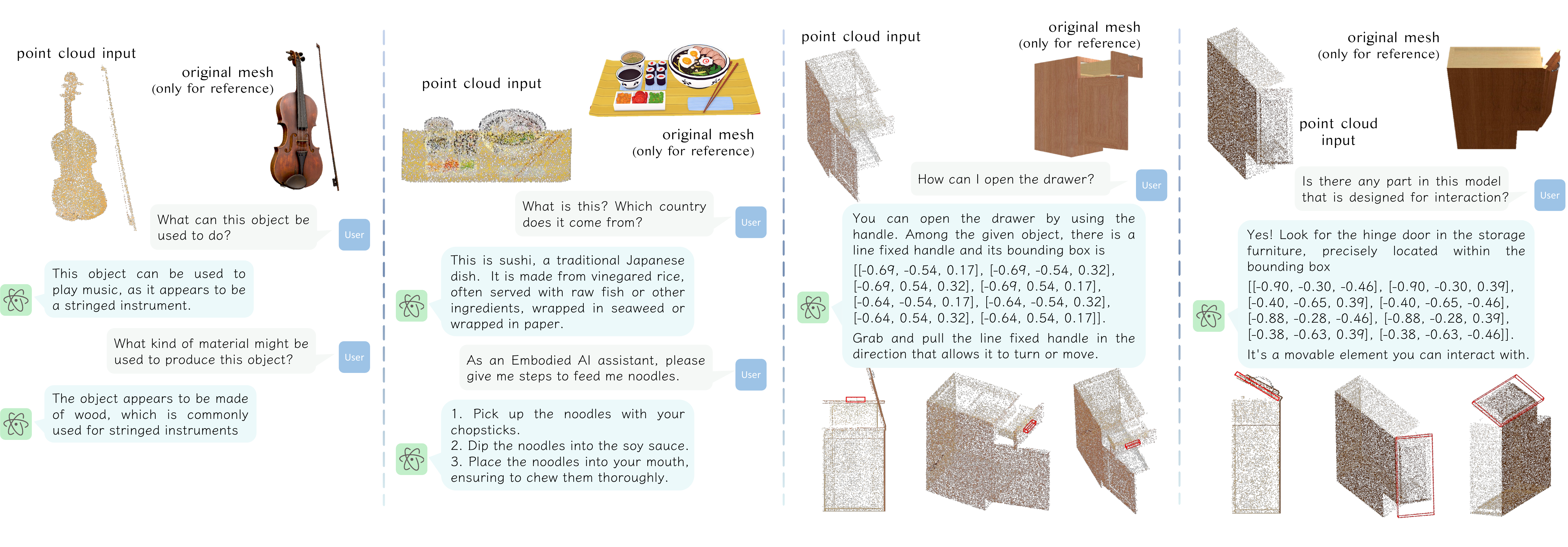}
\vspace{-21pt}
\captionof{figure}{\textbf{Selected multimodal dialogue examples.} \shapellm\ possesses robust capabilities in knowledge representation, reasoning, and instruction-following dialogue. With its powerful point cloud encoder \recon, \shapellm\ can even make accurate predictions about minute interactive components, \eg, handle. The rendered mesh images are solely for visual reference here and do not constitute input data.
}
\vspace{-12pt}
\label{fig:dialogues}
\end{figure}
 \begin{table}[t!]
\scriptsize
  \setlength{\tabcolsep}{4pt}
  \centering
  \caption{\textbf{Zero-shot 3D recognition} on Objaverse-LVIS~\cite{objaverse23}, ModelNet40~\cite{ModelNet15} and ScanObjectNN~\cite{ScanObjectNN19}. Ensembled~\cite{OpenShape23}: pretraining with four datasets, Objaverse~\cite{objaverse23}, ShapeNet~\cite{ShapeNet15}, ABO~\cite{ABO22} and 3D-FUTURE~\cite{3d-future21}. $^\dagger$: Uni3D employs a larger EVA-CLIP-E~\cite{EVACLIP23} teacher, while other methods employ OpenCLIP-bigG~\cite{OpenCLIP21}.}
  \vspace{-8pt}
  \resizebox{0.86\linewidth}{!}{
    \begin{tabular}{lcccccccccc}
    \toprule[0.95pt]
    \multirow{2}{*}[-0.5ex]{Method} &  \multicolumn{3}{c}{\textcolor{black}{\textbf{Objaverse-LVIS}}} & \multicolumn{3}{c}{\textcolor{black}{\textbf{ModelNet40}}} & \multicolumn{3}{c}{\textcolor{black}{\textbf{ScanObjectNN}}}\\
    \cmidrule{2-10} & Top1 & Top3 & Top5 & Top1 & Top3 & Top5 & Top1 & Top3 & Top5 \\
    \midrule[0.6pt]
    \multicolumn{10}{c}{\textit{2D Inference without 3D Training}}\\
    \midrule[0.6pt]
    PointCLIP~\cite{PointCLIP22} & 1.9 & 4.1 & 5.8 & 19.3 & 28.6 & 34.8 & 10.5 & 20.8 & 30.6\\
    PointCLIPv2~\cite{pointclipv223} & 4.7 & 9.5 & 12.9 & 63.6 & 77.9 & 85.0 & 42.2 & 63.3 & 74.5\\
    \midrule[0.6pt]
    \multicolumn{10}{c}{\textit{Trained on ShapeNet}}\\
    \midrule[0.6pt]
    {\scshape ReCon}~\cite{ReCon23} & 1.1 & 2.7 & 3.7 & 61.2 & 73.9 & 78.1 & 42.3 & 62.5 & 75.6\\
    CLIP2Point~\cite{CLIP2Point22} & 2.7 & 5.8 & 7.9 & 49.5 & 71.3 & 81.2 & 25.5 & 44.6 & 59.4\\
    ULIP~\cite{ULIP22} & 6.2 & 13.6 & 17.9 & 60.4 & 79.0 & 84.4 & 51.5 & 71.1 & 80.2\\
    OpenShape~\cite{OpenShape23} & 10.8 & 20.2 & 25.0 & 70.3 & 86.9 & 91.3 & 47.2 & 72.4 & 84.7\\
    TAMM~\cite{TAMM24} & 13.7 & 24.2 & 29.2 & 73.1 & 88.5 & 91.9 & 54.8 & 74.5 & 83.3\\
    MixCon3D~\cite{MixCon3D24} & 22.3& 37.5& 44.3 & 72.6&87.1  & 91.3 &52.6  & 69.9 & 78.7\\
    \midrule[0.6pt]
    \multicolumn{10}{c}{\textit{Trained on Ensembled}}\\
    \midrule[0.6pt]
    ULIP-2~\cite{ULIP2_23} & 26.8 & 44.8 & 52.6 & 75.1 & 88.1 & 93.2 & 51.6 & 72.5 & 82.3\\
    OpenShape~\cite{OpenShape23} & 46.8 & 69.1 & 77.0 & 84.4 & 96.5 & 98.0 & 52.2 & 79.7 & 88.7\\
    TAMM~\cite{TAMM24} & 50.7 & 73.2 & 80.6 & 85.0 & 96.6 & 98.1 & 55.7 & 80.7 & 88.9\\
    MixCon3D~\cite{MixCon3D24} & 52.5 & 74.5 & 81.2 & \textbf{86.8} & \textbf{96.9} & \textbf{98.3} & 58.6 & 80.3 & 89.2\\
    Uni3D-B$^\dagger$~\cite{Uni3D23} & 51.7 & 74.1 & 80.8 & 86.3 & 96.5 & 97.9 & \textbf{63.8} & \textbf{82.7} & 90.2\\
    Uni3D-L$^\dagger$~\cite{Uni3D23} & 53.1 & 75.0 & 81.5 & 86.3 & \textbf{96.8} & \textbf{98.3} & 58.2 & 81.8 & 89.4\\
    \rowcolor{linecolor4}\textbf{\recon-B} & \textbf{53.2} & \textbf{75.3} & \textbf{81.5} & 86.5 & 94.7 & 95.8 & 63.6 & 80.2 & \textbf{90.6} \\
    \rowcolor{linecolor3}\textbf{\recon-L} & \textbf{53.7} & \textbf{75.8} & \textbf{82.0} & \textbf{87.3} & 95.4 & 96.1 &
    \textbf{65.4} & \textbf{84.1} & \textbf{89.7} \\
    \bottomrule[0.95pt]
    \end{tabular}
    }
  \label{tab:zeroshot}
  \vspace{-8pt}
\end{table}

 \vspace{-2pt}
 To prevent any overlap with training data, our collection of 3D models is sourced exclusively from Turbosquid~\cite{Turbosquid}, a platform not included in the acquisition lists of Objaverse~\cite{objaverse23} and ShapeNet~\cite{ShapeNet15}. We meticulously curated a dataset of 59 3D models, generating 232 Q\&As for evaluation purposes. In our pursuit of a precise assessment of single-task capabilities, each question is designed to test only one specific capacity outlined earlier. Every question is paired with a corresponding answer tailored to the particular 3D model, serving as the ground truth. Dataset samples are illustrated in \cref{fig:mmvet}. More details and analysis can be found in the supplemental material.

\vspace{-4pt}
\section{Experiments}
\vspace{-2pt}
\subsection{3D Representation Transferring with \textbf{\recon}}
\vspace{-2pt}
\noindent\textbf{Fine-tuned 3D Object Recognition}~
In \cref{tab:cls}, we first evaluate the representation transfer learning capabilities of self-supervised \recon\ by fine-tuning on ScanObjectNN~\cite{ScanObjectNN19} and ModelNet~\cite{ModelNet15}, which are currently the two most challenging 3D object datasets.
ScanObjectNN is a collection of $\sim$15K 3D object point clouds from the real-world scene dataset ScanNet~\cite{ScanNet17}, which involves 15 categories.
ModelNet is one of the most classical 3D object datasets collected from clean 3D CAD models, which includes $\sim$12K meshed 3D CAD models covering 40 categories.
Following PointGPT~\cite{pointgpt23}, we adopt the intermediate fine-tuning strategy and use the post-pretraining stage to transfer the general semantics learned through self-supervised pretraining on ShapeNetCore~\cite{ShapeNet15}.
For a fair comparison, our Base and Large models adopt the same architecture as PointGPT regarding layers, hidden size, and attention heads.
\cref{tab:cls} shows that: (i) \recon\ exhibits representation performance significantly surpassing that of other baselines, achieving state-of-the-art results. (ii) Particularly, \recon\ achieves a remarkable accuracy of 95.25\% on the most challenging ScanObjectNN PB\_T50\_RS benchmark, boosting the Transformer baseline by +16.14\%.
\begin{table}[t!]
\setlength{\tabcolsep}{4pt}
  \caption{\textbf{Zero-shot 3D multimodal comprehension} of \textit{core VL capabilities in 3D context} on 3D MM-Vet.
  \textbf{\texttt{Rec}}: General Visual Recognition, \textbf{\texttt{Know}}: Knowledge, \textbf{\texttt{Gen}}: Language Generation, \textbf{\texttt{Spat}}: Spatial Awareness, \textbf{\texttt{Emb}}: Embodied Interaction.
  }
  \label{tab:mmvet}
  \vspace{-8pt}
  \centering
    \resizebox{0.82\linewidth}{!}{
    \begin{tabular}{lcccccccc}
    \toprule[0.95pt]
        Method & Input & \textbf{\texttt{Rec}} & \textbf{\texttt{Know}} & \textbf{\texttt{Gen}} & \textbf{\texttt{Spat}} & \textbf{\texttt{Emb}} & Total  \\ 
        \midrule[0.6pt]
        LLaVA-13B~\cite{LLaVA23} & 1-View 2D Image & 40.0 & 55.3 & 51.3 & 43.2 & 51.1 & 47.9 \\
        DreamLLM-7B~\cite{DreamLLM23} & 4-View 2D Image & 42.2 & 54.4 & 50.8 & 48.9 & 54.5 & 50.3 \\
        GPT-4V~\cite{GPT4Vision23} & 1-View 2D Image & 53.7 & 59.5 & 61.1 & 54.7 & 59.0 & 57.4 \\
        GPT-4V~\cite{GPT4Vision23} & 4-View 2D Image & 65.1 & 69.1 & 61.4 & 52.9 & 65.5 & 63.4 \\
        \midrule[0.6pt]
         PointBind\&LLM \cite{pointbind23} & 3D Point Cloud & 16.9 & 13.0 & 18.5 & 32.9 & 40.4 & 23.5 \\
        PointLLM-7B~\cite{pointllm23} & 3D Point Cloud & 40.6 & 49.5 & 34.3 & 29.1 & 48.7 & 41.2 \\
        PointLLM-13B~\cite{pointllm23} & 3D Point Cloud & 46.6 & 48.3 & 38.8 & 45.2 & 50.9 & 46.6 \\
        \rowcolor{linecolor4}\textbf{\shapellm-7B} & 3D Point Cloud& \textbf{45.7} & \textbf{42.7} & \textbf{43.4} & \textbf{39.9} & \textbf{64.5} & \textbf{47.4}\\
        \rowcolor{linecolor3}\textbf{\shapellm-13B} & 3D Point Cloud & \textbf{46.8} & \textbf{53.0} & \textbf{53.9} & \textbf{45.3} & \textbf{68.4} & \textbf{53.1}\\
        \bottomrule[0.95pt]
    \end{tabular}
    }
\vspace{-16pt}
\end{table}

\vspace{4pt}
\noindent\textbf{Zero-Shot 3D Open-World Recognition}~
Similar to CLIP~\cite{CLIP}, our model aligns the feature space of languages and other modalities, which results in a zero-shot open-world recognition capability.
In \cref{tab:zeroshot}, we compare the zero-shot 3D open-world object recognition models to evaluate the generalizable recognition capability.
Following OpenShape~\cite{OpenShape23}, we evaluate on ModelNet~\cite{ModelNet15}, ScanObjectNN~\cite{ScanObjectNN19}, and Objaverse-LVIS~\cite{objaverse23}.
Objaverse-LVIS is a benchmark involving $\sim$47K clean 3D models of 1,156 LVIS categories~\cite{LVIS19}. 
We compare \recon\ with 2D inference methods, ShapeNet pretrained methods, and ``Ensembled'' datasets-pretrained methods.
It can be concluded from \cref{tab:zeroshot}:
i) Compared to 2D inference and ShapeNet-pretrained methods, \recon\ demonstrates significantly superior performance, showing the necessity of \textit{3D point clouds as inputs} and \textit{scaling up}.
ii) Compared to state-of-the-art methods trained on ``Ensembled'' datasets, \recon\ demonstrates superior or on-par performance across all benchmarks.
Notably, \recon-L achieves a remarkable Top-1 accuracy, which is +0.6\% and +7.2\% higher than Uni3D-L on the most challenging Objaverse-LVIS and ScanObjectNN benchmarks, respectively.

\vspace{-5pt}
\subsection{Multimodal Comprehension with \textbf{\shapellm}}\label{sec:vqa_exp}
\noindent\textbf{Quantitative Analysis}~
To assess the comprehensive capabilities of \shapellm, we first quantitatively compare various baselines and our model on the proposed 3D MM-Vet using GPT-4. Following ModelNet-C~\cite{ModelNetC22} and ModelNet40-C~\cite{ModelNet40C22}, we construct 3D MM-Vet-C to benchmark the robustness against 3D corruptions.
\vspace{3pt}
\begin{wrapfigure}{r}{0.5\textwidth}
    \vspace{-0.83cm}
    \makeatletter\def\@captype{table}\makeatother
    \caption{\textbf{Zero-shot 3D multimodal comprehension} of \textit{robustness} on 3D MM-Vet-C. \textbf{\texttt{Clean}}: no corruptions. \textbf{\texttt{Single-View}}: randomly select a camera viewpoint within the unit sphere and generate a \textbf{single viewpoint} within the FoV on polar coordinates. \textbf{\texttt{Jitter}}: Gaussian jittering with noise $\epsilon\sim\mathcal{N}(0,\sigma^2)$ and $\sigma=0.01$.
    \textbf{\texttt{Rotate}}: random SO(3) rotation sampling over X-Y-Z Euler angle $(\alpha,\beta,\gamma)\sim \mathcal{U}(-\theta,\theta)$ and $\theta=\pi/6$.
  }
  \label{tab:mmvetc}
  \vspace{-6pt}
  \centering
    \resizebox{0.5\textwidth}{!}{
    \begin{tabular}{lccccc}
    \toprule[0.95pt]
        \multirow{2}{*}[-0.5ex]{Method} & \multicolumn{4}{c}{\textbf{3D MM-Vet-C Variants}}\\
        \cmidrule(lr){2-5}
        & \textbf{\texttt{Clean}} & \textbf{\texttt{Single}}-\textbf{\texttt{View}} & \textbf{\texttt{Jitter}} & \textbf{\texttt{Rotate}}\\ 
        \midrule[0.6pt]
        PointBind\&LLM~\cite{pointbind23} & 23.5 & 20.4 & 19.7 & 19.5\\
        PointLLM-7B~\cite{pointllm23} & 41.2 & 33.6 & 38.8 & 40.6\\
        PointLLM-13B~\cite{pointllm23} & 46.6 & 41.3 & 42.3 & 44.2\\
        \rowcolor{linecolor4}\textbf{\shapellm-7B} & \textbf{47.4} & \textbf{38.3} & \textbf{45.8} & \textbf{42.7}\\
        \rowcolor{linecolor3}\textbf{\shapellm-13B} & \textbf{53.1} & \textbf{43.6} & \textbf{47.8} & \textbf{49.3}\\
        \bottomrule[0.95pt]
    \end{tabular}}
\vspace{-0.6cm}
\end{wrapfigure}
\noindent\textbf{i) 3D MM-Vet.}~
\cref{tab:mmvet} shows the detailed results of \shapellm\ on different tasks of 3D MM-Vet.
It is observed that \shapellm\ significantly outperforms PointLLM~\cite{pointllm23} across various metrics, particularly in Embodied Tasks. This substantiates our model's versatile capability in addressing real-world tasks.

\vspace{3pt}
\noindent\textbf{ii) 3D MM-Vet-C.}~
Following the ModelNet-C~\cite{ModelNetC22} and ModelNet40-C~\cite{ModelNet40C22}, we construct 3D MM-Vet-C to benchmark the robustness against 3D corruptions.
\cref{tab:mmvetc} shows the comparison of robustness against ``single-view'', ``jitter'', and ``rotate'' corruptions, which are the most common in real scenarios. 
The ``single-view'' issue is the most critical challenge since obtaining the complete point clouds is non-trivial, similar to multi-view images.
Therefore, everyday real-world robots only get single-view 3D perceptions with sensors such as RGB-D~\cite{SGRGBD14}.
The results demonstrate significantly superior robustness of \shapellm, indicating stronger potential in real-world applicability. 

\noindent\textbf{Baseline Improvement}~
Can we improve the baseline to bridge the gap between PointLLM and \shapellm? In \cref{tab:improvement}, we study two technical factors that are contributed by \shapellm: 3D point cloud encoder and SFT data.
\begin{figure}[t!]
\centering
\includegraphics[width=\linewidth]{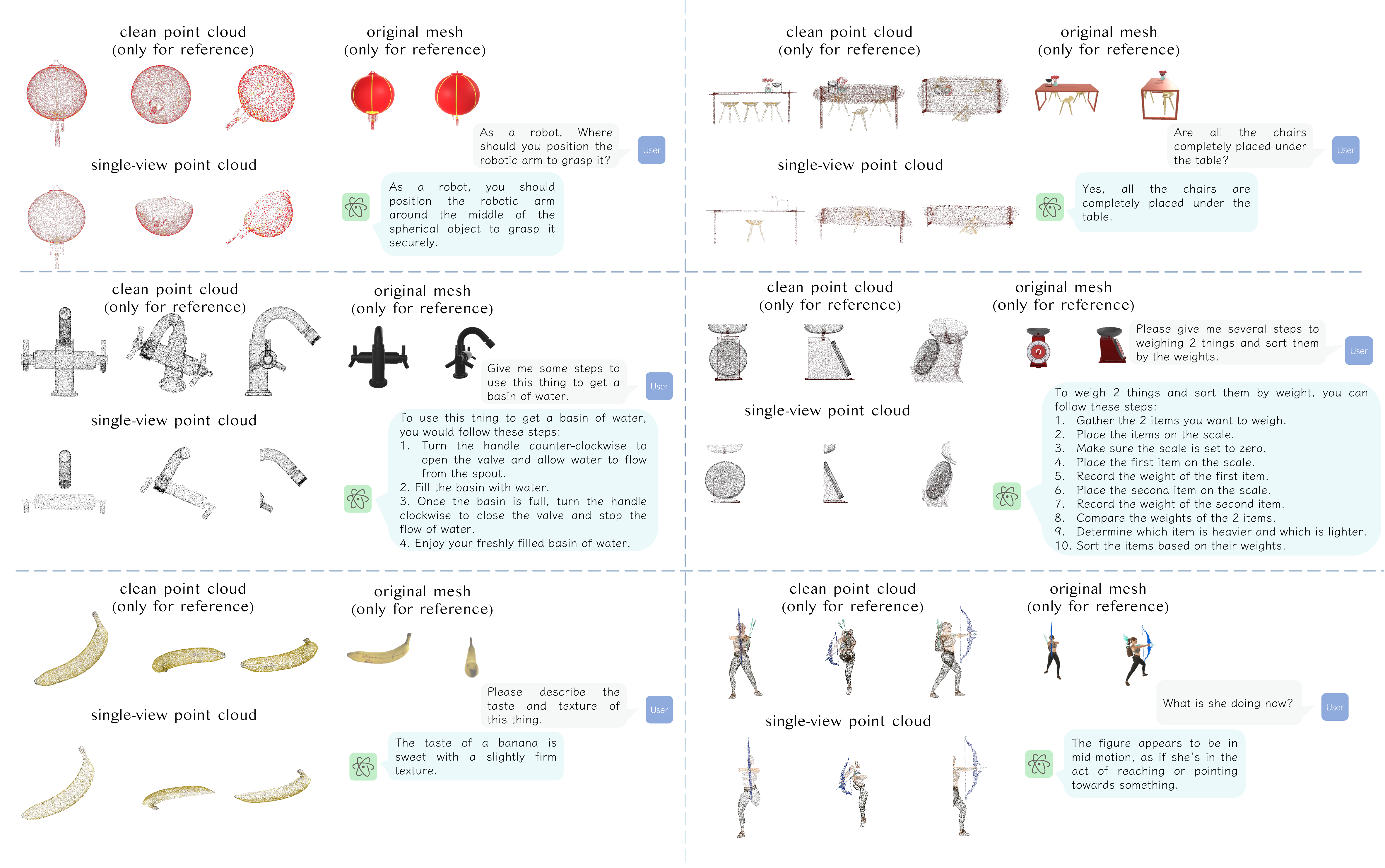}
\vspace{-22pt}
\captionof{figure}{\textbf{3D multimodal dialogue using \textit{single-view} point cloud inputs}. All answers are generated by \shapellm-13B with single-view occluded inputs. \shapellm\ achieves outstanding robustness against such commonly met occlusion in the real world.
}\label{fig:singleview_qa}
\vspace{-10pt}
\end{figure}

\vspace{5pt}
\begin{wrapfigure}{r}{0.55\textwidth}
    \vspace{-0.82cm}
    \makeatletter\def\@captype{table}\makeatother
    \caption{\textbf{Ablation study on baseline improvements}. Results are tested on 3D MM-Vet with the baseline model PointLLM-13B~\cite{pointllm23} using different point encoders and SFT data.
  }
  \label{tab:improvement}
  \vspace{-5pt}
  \centering
    \resizebox{\linewidth}{!}{
    \begin{tabular}{lcccccccc}
    \toprule[0.95pt]
        Encoder & SFT Data & \textbf{\texttt{Rec}} & \textbf{\texttt{Know}} & \textbf{\texttt{Gen}} & \textbf{\texttt{Spat}} & \textbf{\texttt{Emb}} & Total  \\ 
        \midrule[0.6pt]
        ULIP-2~\cite{ULIP2_23} & PointLLM & 46.6 & 48.3 & 38.8 & 45.2 & 50.9 & 46.6\\
        \cellcolor{linecolor4}\textbf{\recon} & PointLLM & 47.5 & 52.8 & 43.6 & 44.9 & 54.5 & 50.8\\
        \cellcolor{linecolor3}\textbf{\recon} & \cellcolor{linecolor3}\textbf{Ours} & 46.8 & 53.0 & 53.9 & 45.3 & 68.4 & 53.1\\
        \bottomrule[0.95pt]
    \end{tabular}
    }
\vspace{-0.75cm}
\end{wrapfigure}

\noindent\textbf{i) Improvement from encoder.}

\noindent First, by changing PointLLM's encoder to \recon, a significant improvement of +4.20\% is obtained. This demonstrates the significantly better 3D representation extraction of \recon\ compared to ULIP-2. 
It is consistent with previous findings in \cref{tab:cls} and \cref{tab:zeroshot} that \recon\ outperforms ULIP-2 by a large margin regarding 3D representation transferring learning and zero-shot learning.

\vspace{5pt}
\noindent\textbf{ii) Improvement from data.}~
As stated in \cref{sec:sft_data_gpt4v}, we have constructed instruction-following data for supervised fine-tuning (SFT) using GPT-4V involving diverse topics.
By further using the SFT data curated by us, PointLLM's performance gap to \shapellm\ has been fulfilled.
This demonstrates the superiority of our SFT data, where the decent quality comes from the advanced GPT4-V using multi-view images and the topics covered in the data.

\vspace{5pt}
\noindent\textbf{Qualitative Analysis}~
\cref{fig:dialogues} illustrates qualitative examples of \shapellm\ in \textit{multimodal dialogue}. \shapellm\ can support general VQA, embodied task and action planning, and 6-DoF pose estimation. Notably, LLMs easily grasp such patterns and consistently produce valid coordinates due to the strict spatial relationship inherent in 6-DoF bounding box coordinates. \cref{fig:singleview_qa} shows the examples of \shapellm-13B's response using \textit{single-view point cloud inputs}, demonstrating surprisingly outstanding robustness in processing such occlusion. This is crucial for the practical deployment of real machines, as single-view point clouds can be easily obtained from RGB-D cameras.

\vspace{-5pt}
\section{Discussions}\label{Discuss}
\vspace{-4pt}
\subsection{Is \textbf{\shapellm}\ grounded in physical worlds?}
\vspace{-2pt}
\cref{tab:gapartnet} compares \shapellm\ with image-only methods on 3D referring expression grounding (REG) of 6-DoF poses on GAPartNet~\cite{gapartnet23}.
The results show that: i) Image-only methods cannot perform zero-shot geometry-necessary 6-DoF pose estimation. 
ii) Compared to image-only methods with 2D to 6-DoF pose estimation fine-tuning or in-context prompting, \shapellm\ still performs significantly better. It demonstrates the necessity of geometry and the difficulty of the ill-posed 2D to 6-DoF pose estimation problem, as well as the importance of using 3D point clouds as input for spatial intelligence.
\begin{table}[t!]
\caption{\textbf{3D referring expression grounding} on GAPartNet~\cite{gapartnet23}. Accuracy with an IoU threshold of 0.25 is reported. $^\dagger$: Fine-tuned on GAPartNet images. $^\ddagger$: Inference with 3 in-context demonstrations.}
\vspace{-17pt}
\label{tab:gapartnet}
\setlength{\tabcolsep}{4pt}
\begin{center}
    \resizebox{0.84\linewidth}{!}{
    \begin{tabular}{lcccccccc}
    \toprule[0.95pt]
    \multirow{1}{*}[1.0ex]{Method} & \multirow{1}{*}[1.0ex]{Input} & \includegraphics[width=0.045\linewidth]{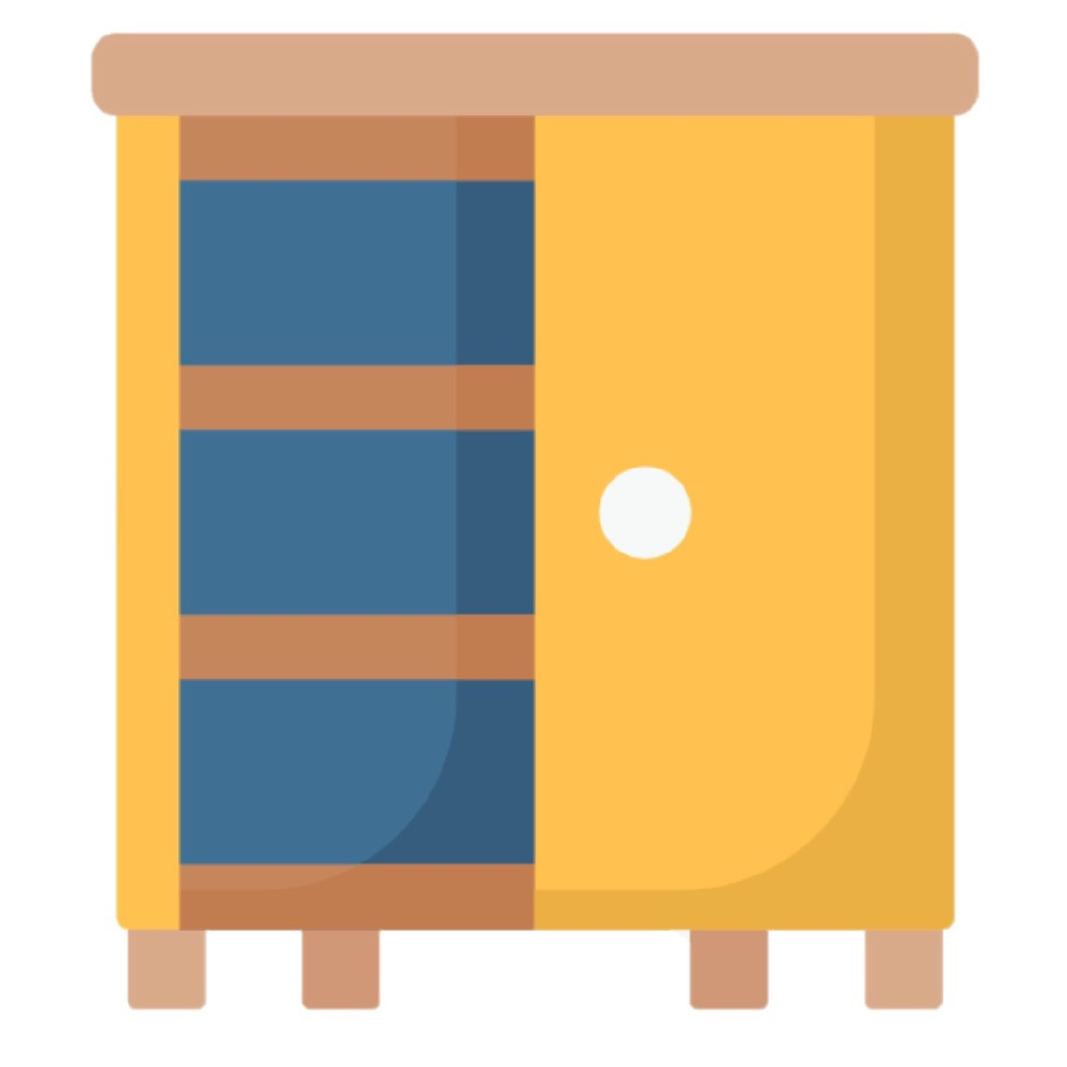} & \includegraphics[width=0.045\linewidth]{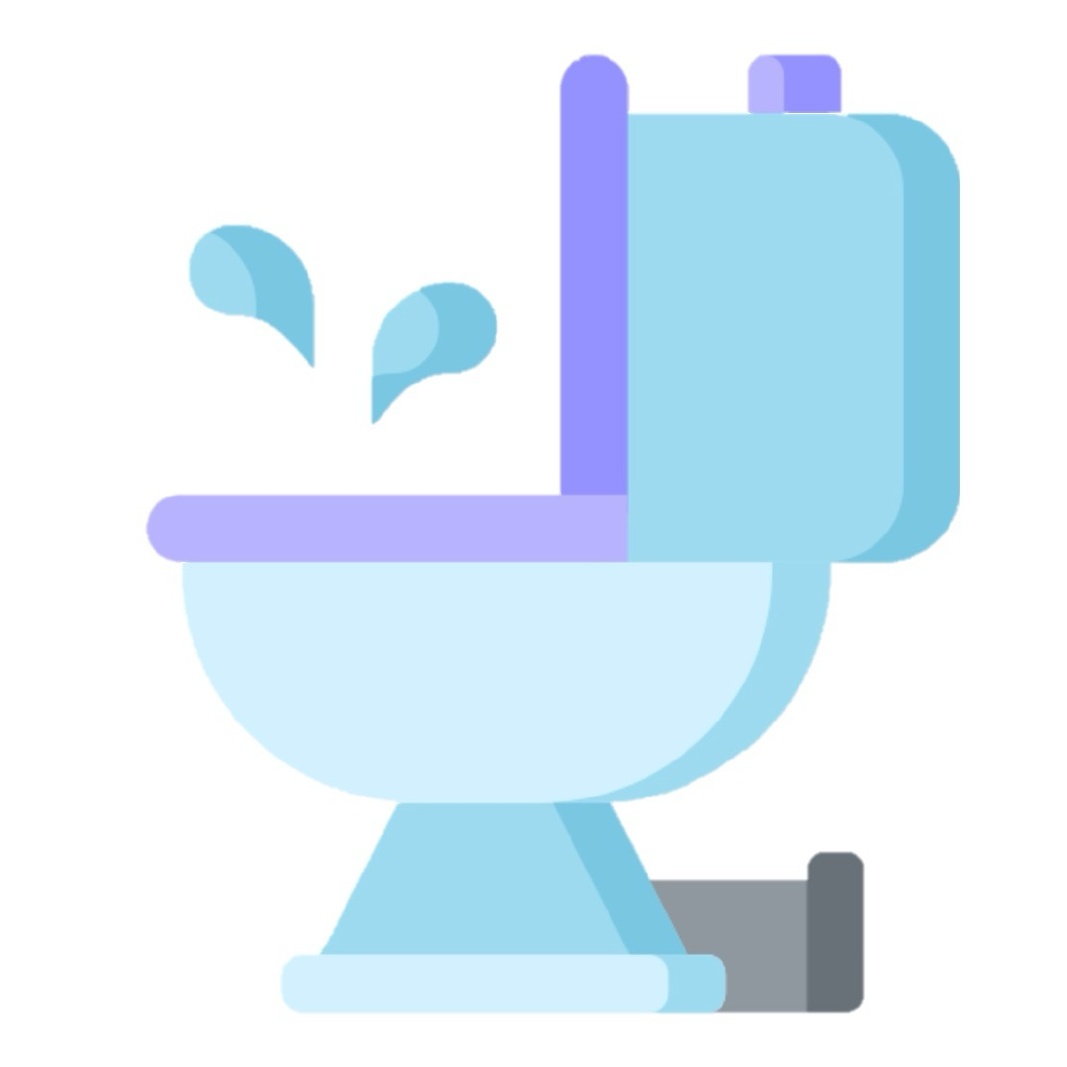} & \includegraphics[width=0.045\linewidth]{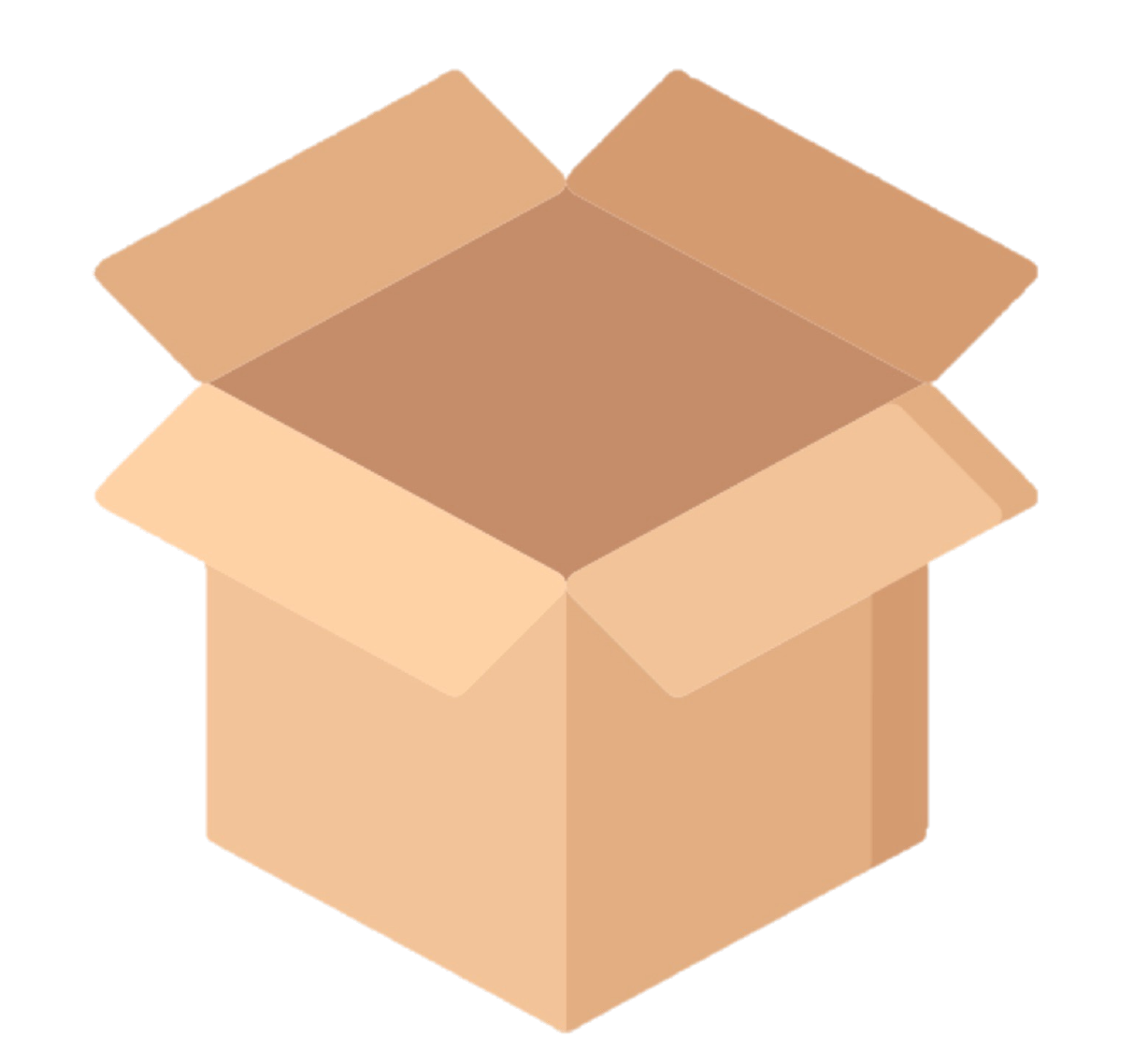} & \includegraphics[width=0.045\linewidth]{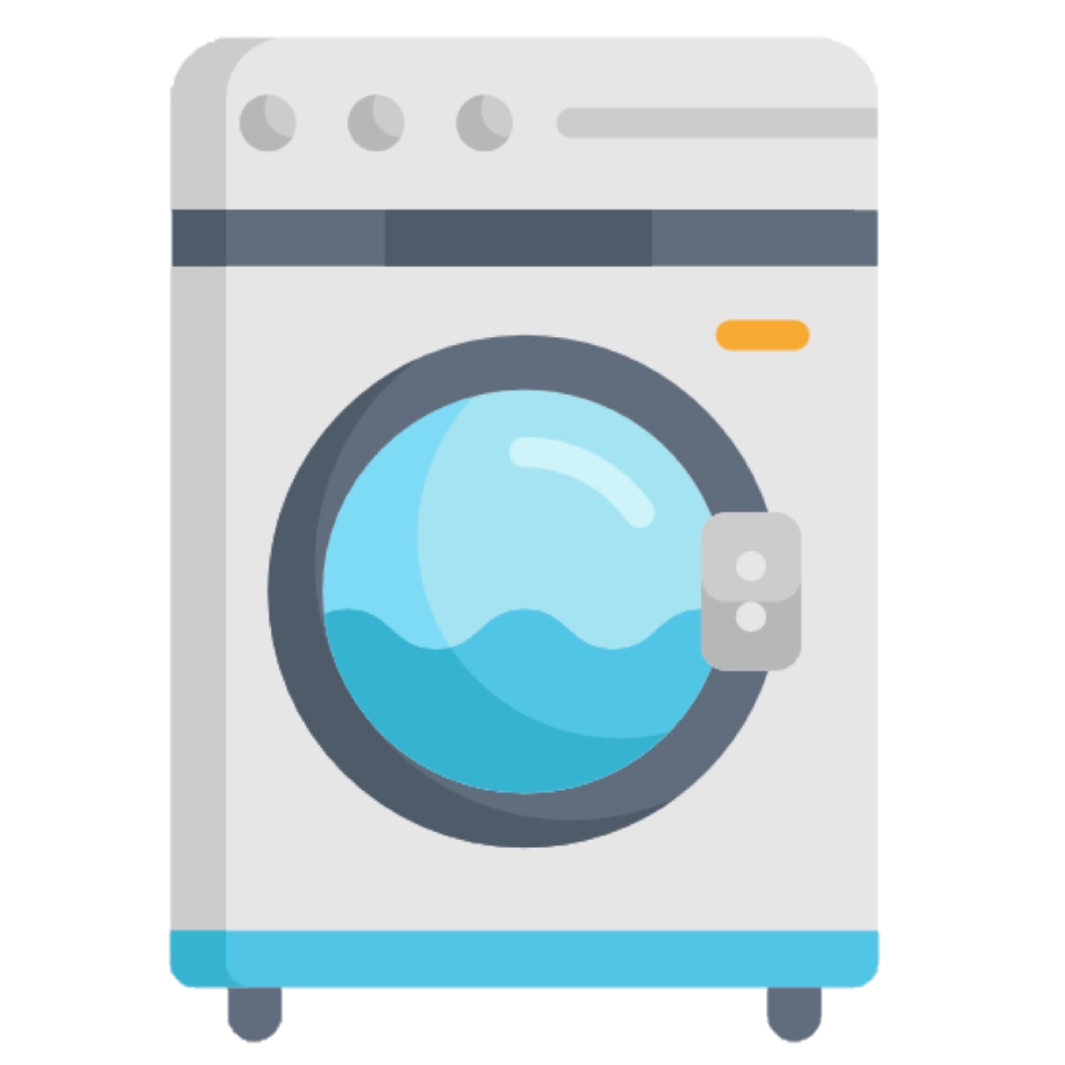} & \includegraphics[width=0.045\linewidth]{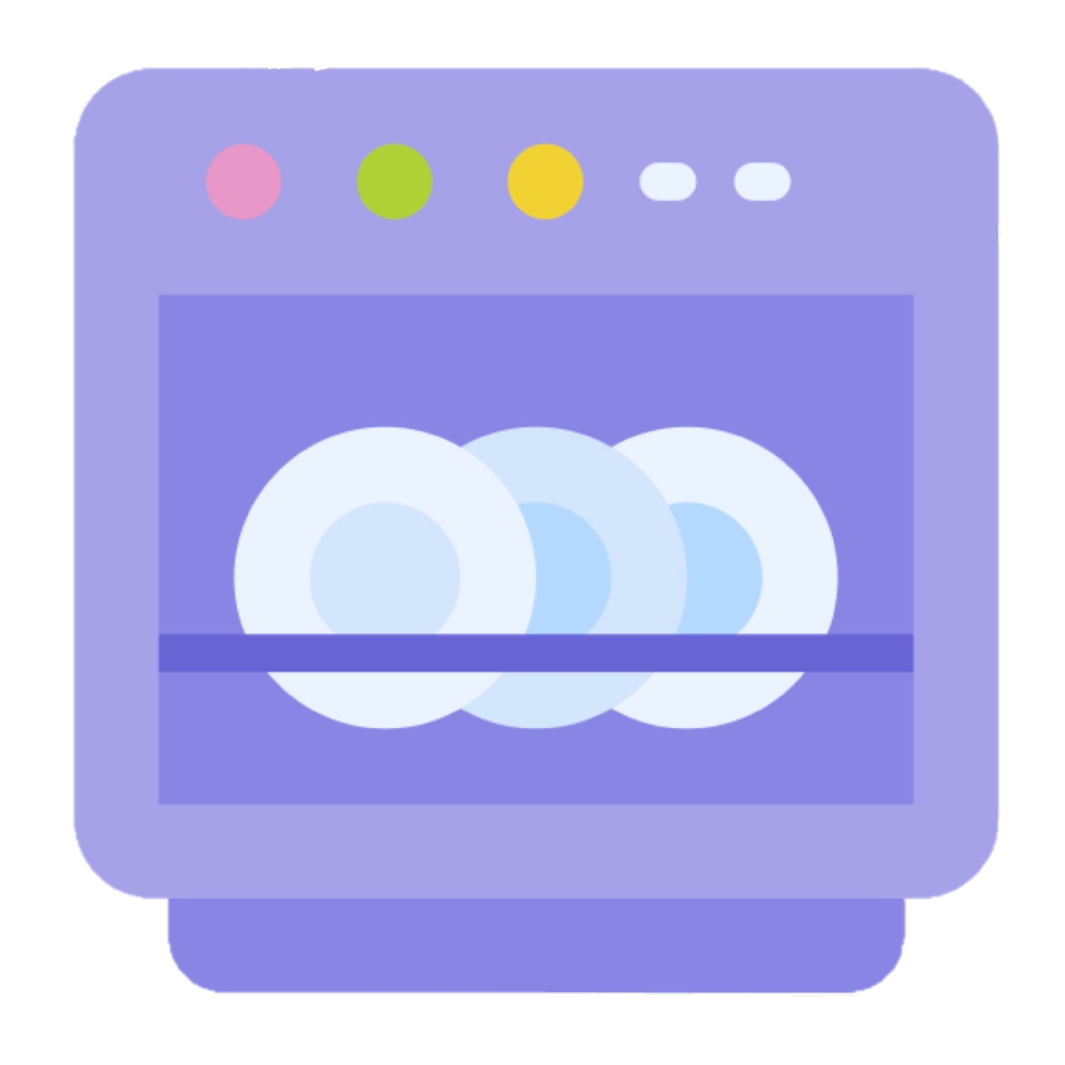} & \includegraphics[width=0.045\linewidth]{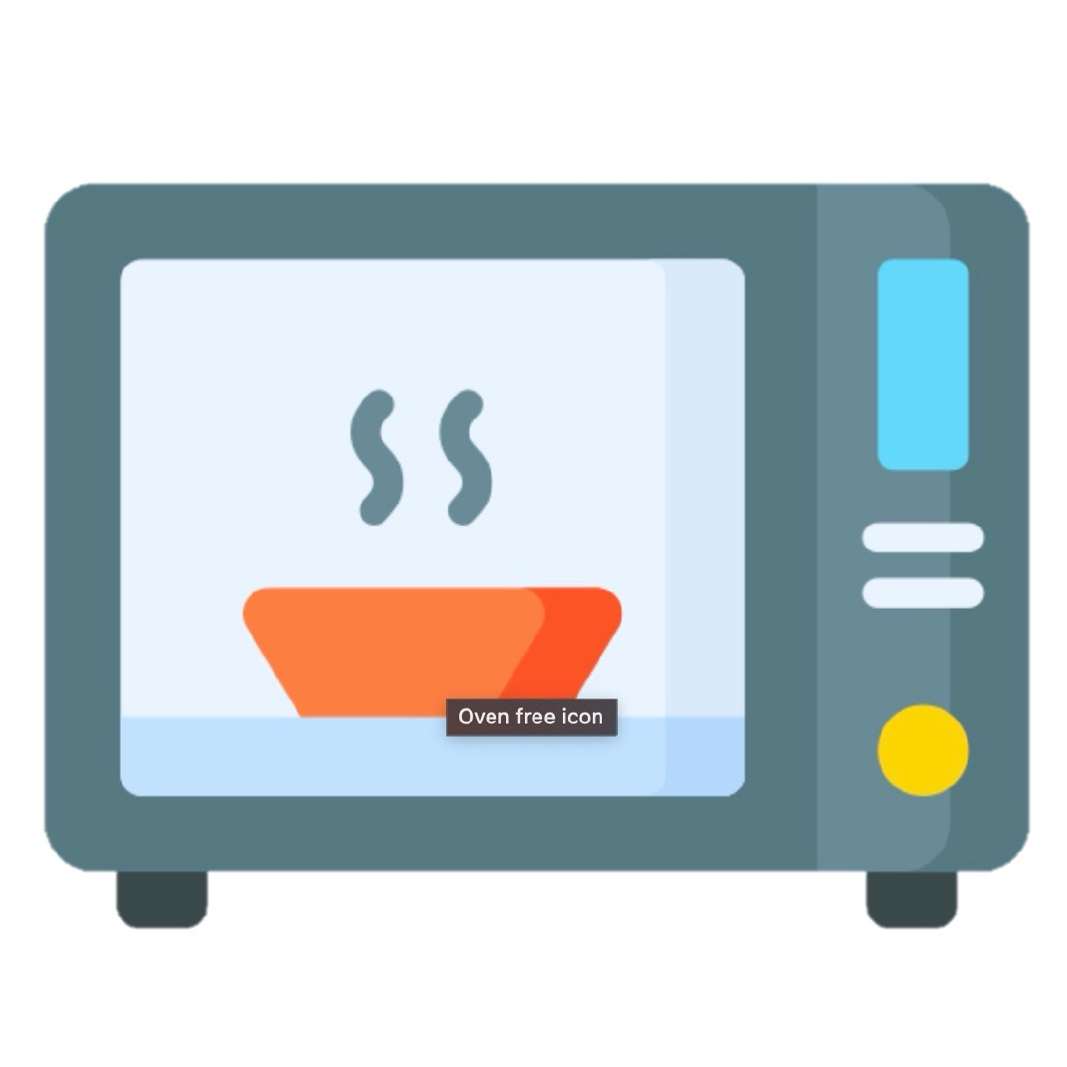} & \multirow{1}{*}[1.0ex]{Avg}\\
    \midrule[0.6pt]
    LLaVA-13B~\cite{LLaVA23} & 1-View 2D Image & 0.0 & 0.0 & 0.0 & 0.0 & 0.0 & 0.0 & 0.0\\
    
    LLaVA-13B~\cite{LLaVA23} & 4-View 2D Image & 0.0 & 0.0 & 0.0 & 0.0 & 0.0 & 0.0 & 0.0\\
    LLaVA-13B$^\dagger$~\cite{LLaVA23} & 1-View 2D Image & 1.8 & 9.3 & 3.8 & 0.0 & 2.1 & 11.1 & 4.4\\
    LLaVA-13B$^\dagger$~\cite{LLaVA23} & 4-View 2D Image & 2.5 & 13.7 & 7.7 & 0.0 & 4.3 & 11.1 & 6.2\\
    GPT-4V~\cite{GPT4Vision23} & 4-View 2D Image & 0.0 & 0.0 & 0.0 & 0.0 & 0.0 & 0.0 & 0.0\\
    GPT-4V$^\ddagger$~\cite{GPT4Vision23} & 4-View 2D Image & 0.1 & 1.6 & 0.0 & 0.0 & 0.0 & 0.0 & 0.3\\
    \midrule[0.6pt]
    \rowcolor{linecolor4}\textbf{\shapellm-7B} & 3D Point Cloud & \textbf{5.9} & \textbf{25.8} & \textbf{11.5} & \textbf{3.4} & \textbf{5.1} & \textbf{11.1} & \textbf{10.5}\\
    \rowcolor{linecolor3}\textbf{\shapellm-13B} & 3D Point Cloud & \textbf{7.6} & \textbf{26.7} & \textbf{11.5} & \textbf{6.7} & \textbf{6.8} & \textbf{11.1} & \textbf{11.7}\\
    \bottomrule[0.95pt]
    \end{tabular}
    }
\end{center}
\vspace{-16pt}
\end{table}

\vspace{-5pt}
\subsection{Can \textbf{\shapellm}\ generalize to unseen objects?}
\vspace{-2pt}
\begin{wrapfigure}{r}{0.53\textwidth}
\vspace{-0.8cm}
\centering
\includegraphics[width=0.99\linewidth]{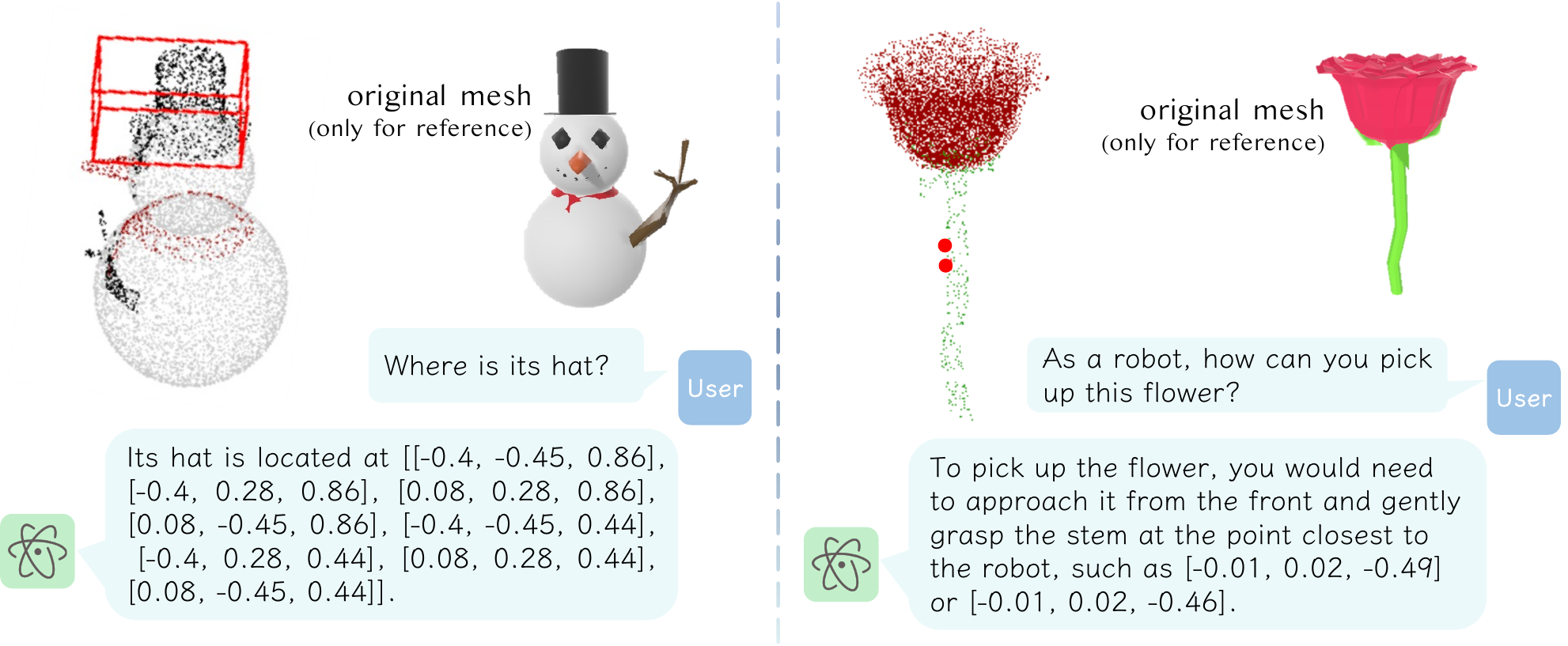}
\vspace{-20pt}
\captionof{figure}{\textbf{Part understanding examples} of unseen objects beyond GAPartNet.
}\label{fig:ov_part}
\vspace{-0.7cm}
\end{wrapfigure}
\cref{fig:ov_part} shows the part understanding examples of unseen objects.
While \shapellm's 6-DoF pose estimation and spatial awareness are trained on GAPartNet, which primarily consists of \textit{indoor articulated furniture} objects. 
It has demonstrated promising generalization potential of spatial understanding on the \textit{open-world objects}, paving ways for scaling up spatial awareness training.

\section{Related Works}
\vspace{-3pt}
\textbf{Interaction-Oriented 3D Understanding}~
Interaction with 3D objects typically involves concept-only interaction and physical-grounded interaction~\cite{SpatialVLM24}.
The former works focus on 3D perception and semantic parsing, such as 3D object recognition and scene perception~\cite{PointNet,PointNet++,DGCNN,PrimitiveTransformer22,Leaf23}.
By utilizing language for open-ended interaction in 3D, a number of works demonstrate successful 3D scene QA~\cite{ThreeDQA22,SQA3D23}, grounding~\cite{ScanRefer}, and captioning~\cite{Scan2Cap21}.
Recently, some works propose to utilize foundation models like LLMs or CLIP for open-ended 3D object recognition~\cite{PointCLIP22,pointclipv223,OpenShape23,ACT23} and scene segmentation~\cite{CLIPFO3D23,OpenScene23}.
Guo \& Zhang~\etal~\cite{pointbind23} utilizes ImageBind~\cite{imagebind23} and LLaMA-Adapter~\cite{LLaMA-Adapter23} to realize point cloud-based interactive QA. 
Following LLaVA, PointLLM~\cite{pointllm23} conducts supervised fine-tuning by constructing a visual instruction-following dataset.
Other works focus on scene-level tasks utilizing comprehensive 2D features~\cite{3DVista23,Leo23} or 3D features distilled from 2D images into LLMs~\cite{3DLLM23,3DVista23,Leo23}.
The second kind of interaction typically requires physical understanding in 3D, such as part understanding~\cite{ImagePartStates18,partnet19,gapartnet23,FewArtObjGen23}, 6-DoF pose estimation~\cite{DeepPartInduction18,NOCSPose19,NPCSCatPose20,PartLevelSE3Pose23,CAPTRA21}, particularly useful for human-object interaction (HOI) and robotic manipulation~\cite{GenOHDiffusion24,InHandRotate23,UniDexGrasp23,UniDexGrasp23++23,SAGE23,GengDonut23,GengARNOLD23,GengPartManip23,GengRLAfford23,DistillFeatureNeRFGrasp23,li2023manipllm,gapartnet23,EmbodiedGPT23,sugar24} and complex robotic planning~\cite{PaLME23,LLM-GROP23,VoxPoser23,Text2Motion23,GOAT23,RoboCook23}.
In this work, we focus on both physical and conceptual interactions with 3D shapes for embodied understanding.
\vspace{5pt}

\noindent\textbf{Multimodal Large Language Models}~
Multimodal comprehension, which allows human interaction with textual and visual elements, has witnessed significant advancements, particularly in extending LLMs like LLaMA~\cite{LLaMA23,Alpaca23,Vicuna23}.
The early efforts predominantly revolved around integrating LLMs with various downstream systems by employing it as an agent~\cite{VisualChatGPT23,VisualProgramming23,MM-REACT23, TaskMatrix23,HuggingGPT23,GPT4Tools23,ViperGPT23,DALLE323,Voyager23}. 
Significant success has been demonstrated within this plugin-style framework.
Due to the remarkable capabilities of LLMs, aligning the visual semantic space with language through parameter-efficient tuning~\cite{LoRA22,Flamingo22,BLIP2_23,LLaMA-Adapter23, MiniGPT4_23,mPLUG-Owl23} and instruction tuning~\cite{MultiInstruct23,LLaVA23,InstructBLIP23,DreamLLM23} has emerged as the prevailing approach in current research.
To further enhance interactive capabilities, some approaches have been developed towards visual-interactive multimodal comprehension by precisely referring to instruction tuning~\cite{ChatSpot23,Kosmos2_23,Shikra23,GPT4ROI23}.
Another family advances the developments of LLMs endowed with content creation beyond comprehension~\cite{DreamLLM23,GILL23,Emu223,Emu23,SEED23,NextGPT23,KosmosG23}.

\vspace{-2pt}
\section{Conclusions}
\vspace{-6pt}
This paper introduces \shapellm, the first 3D MLLM for embodied interaction, excelling in generalizable recognition and interaction comprehension. We present \recon, a novel 3D point cloud encoder leveraging multi-view distillation and advanced 3D representation learning, forming the basis for \shapellm. We perform 3D visual instruction tuning on curated instruction-following data for broad and embodied comprehension. Additionally, we establish 3D MM-Vet, a benchmark to evaluate four levels of capacity in embodied interaction scenarios, from fundamental recognition to control statement generation.

\section*{Acknowledgments}
The work was supported by the Dushi Program from Tsinghua University, the National Key R\&D Program of China (2022YFB2804103), and the National Science and Technology Major Project of China (2023ZD0121300).

{
    \bibliographystyle{splncs04}
    \bibliography{main}

\begin{thebibliography}{100}
\providecommand{\url}[1]{\texttt{#1}}
\providecommand{\urlprefix}{URL }
\providecommand{\doi}[1]{https://doi.org/#1}

\bibitem{RepGen3D18}
Achlioptas, P., Diamanti, O., Mitliagkas, I., Guibas, L.J.: Learning representations and generative models for 3d point clouds. In: Int. Conf. Mach. Learn. (ICML) (2018)

\bibitem{Flamingo22}
Alayrac, J., Donahue, J., Luc, P., Miech, A., Barr, I., Hasson, Y., Lenc, K., Mensch, A., Millican, K., Reynolds, M., Ring, R., Rutherford, E., Cabi, S., Han, T., Gong, Z., Samangooei, S., Monteiro, M., Menick, J., Borgeaud, S., Brock, A., Nematzadeh, A., Sharifzadeh, S., Binkowski, M., Barreira, R., Vinyals, O., Zisserman, A., Simonyan, K.: Flamingo: a visual language model for few-shot learning. In: Adv. Neural Inform. Process. Syst. (NeurIPS) (2022)

\bibitem{S3DIS16}
Armeni, I., Sener, O., Zamir, A.R., Jiang, H., Brilakis, I., Fischer, M., Savarese, S.: 3d semantic parsing of large-scale indoor spaces. In: IEEE/CVF Conf. Comput. Vis. Pattern Recog. (CVPR) (2016)

\bibitem{LVM23}
Bai, Y., Geng, X., Mangalam, K., Bar, A., Yuille, A.L., Darrell, T., Malik, J., Efros, A.A.: Sequential modeling enables scalable learning for large vision models. In: IEEE/CVF Conf. Comput. Vis. Pattern Recog. (CVPR) (2024)

\bibitem{METEOR05}
Banerjee, S., Lavie, A.: {METEOR:} an automatic metric for {MT} evaluation with improved correlation with human judgments. In: Proceedings of the Workshop on Intrinsic and Extrinsic Evaluation Measures for Machine Translation and/or Summarization@ACL 2005, Ann Arbor, Michigan, USA, June 29, 2005 (2005)

\bibitem{DALLE323}
Betker, J., Gabriel, G., Jing, L., Brooks, T., Wang, J., Li, L., Ouyang, L., Zhuang, J., Lee, J., Guo, Y., Manassra, W., Dhariwal, P., Chu, C., Jiao, Y., Ramesh, A.: Improving image generation with better captions  (2023)

\bibitem{FoundationModel21}
Bommasani, R., Hudson, D.A., Adeli, E., Altman, R., Arora, S., von Arx, S., Bernstein, M.S., Bohg, J., Bosselut, A., Brunskill, E., Brynjolfsson, E., Buch, S., Card, D., Castellon, R., Chatterji, N.S., Chen, A.S., Creel, K., Davis, J.Q., Demszky, D., Donahue, C., Doumbouya, M., Durmus, E., Ermon, S., Etchemendy, J., Ethayarajh, K., Fei{-}Fei, L., Finn, C., Gale, T., Gillespie, L., Goel, K., Goodman, N.D., Grossman, S., Guha, N., Hashimoto, T., Henderson, P., Hewitt, J., Ho, D.E., Hong, J., Hsu, K., Huang, J., Icard, T., Jain, S., Jurafsky, D., Kalluri, P., Karamcheti, S., Keeling, G., Khani, F., Khattab, O., Koh, P.W., Krass, M.S., Krishna, R., Kuditipudi, R., et~al.: On the opportunities and risks of foundation models. CoRR  \textbf{abs/2108.07258} (2021)

\bibitem{MarginSVM92}
Boser, B.E., Guyon, I., Vapnik, V.: A training algorithm for optimal margin classifiers. In: ACM Conf. Comput. Learn. Theory (COLT). pp. 144--152. {ACM} (1992)

\bibitem{MV3DRec94}
Bradski, G., Grossberg, S.: Recognition of 3-d objects from multiple 2-d views by a self-organizing neural architecture. In: From Statistics to Neural Networks: Theory and Pattern Recognition Applications, pp. 349--375. Springer (1994)

\bibitem{ShapeGoogle11}
Bronstein, A.M., Bronstein, M.M., Guibas, L.J., Ovsjanikov, M.: Shape google: Geometric words and expressions for invariant shape retrieval. {ACM} Trans. Graph.  \textbf{30}(1),  1:1--1:20 (2011)

\bibitem{GPT3_20}
Brown, T.B., Mann, B., Ryder, N., Subbiah, M., Kaplan, J., Dhariwal, P., Neelakantan, A., Shyam, P., Sastry, G., Askell, A., Agarwal, S., Herbert{-}Voss, A., Krueger, G., Henighan, T., Child, R., Ramesh, A., Ziegler, D.M., Wu, J., Winter, C., Hesse, C., Chen, M., Sigler, E., Litwin, M., Gray, S., Chess, B., Clark, J., Berner, C., McCandlish, S., Radford, A., Sutskever, I., Amodei, D.: Language models are few-shot learners. In: Adv. Neural Inform. Process. Syst. (NeurIPS) (2020)

\bibitem{DETR20}
Carion, N., Massa, F., Synnaeve, G., Usunier, N., Kirillov, A., Zagoruyko, S.: End-to-end object detection with transformers. In: Eur. Conf. Comput. Vis. (ECCV) (2020)

\bibitem{ShapeNet15}
Chang, A.X., Funkhouser, T.A., Guibas, L.J., Hanrahan, P., Huang, Q., Li, Z., Savarese, S., Savva, M., Song, S., Su, H., Xiao, J., Yi, L., Yu, F.: Shapenet: An information-rich 3d model repository. CoRR  \textbf{abs/1512.03012} (2015)

\bibitem{GOAT23}
Chang, M., Gervet, T., Khanna, M., Yenamandra, S., Shah, D., Min, S.Y., Shah, K., Paxton, C., Gupta, S., Batra, D., Mottaghi, R., Malik, J., Chaplot, D.S.: {GOAT:} {GO} to any thing. In: Robotics: Science and Systems (RSS) (2024)

\bibitem{SpatialVLM24}
Chen, B., Xu, Z., Kirmani, S., Ichter, B., Driess, D., Florence, P., Sadigh, D., Guibas, L., Xia, F.: Spatialvlm: Endowing vision-language models with spatial reasoning capabilities. In: IEEE/CVF Conf. Comput. Vis. Pattern Recog. (CVPR) (2024)

\bibitem{ScanRefer}
Chen, D.Z., Chang, A.X., Nie{\ss}ner, M.: Scanrefer: 3d object localization in {RGB-D} scans using natural language. In: Eur. Conf. Comput. Vis. (ECCV) (2020)

\bibitem{Scan2Cap21}
Chen, D.Z., Gholami, A., Nie{\ss}ner, M., Chang, A.X.: Scan2cap: Context-aware dense captioning in {RGB-D} scans. In: IEEE/CVF Conf. Comput. Vis. Pattern Recog. (CVPR) (2021)

\bibitem{pointgpt23}
Chen, G., Wang, M., Yang, Y., Yu, K., Yuan, L., Yue, Y.: Pointgpt: Auto-regressively generative pre-training from point clouds. In: Adv. Neural Inform. Process. Syst. (NeurIPS) (2023)

\bibitem{Shikra23}
Chen, K., Zhang, Z., Zeng, W., Zhang, R., Zhu, F., Zhao, R.: Shikra: Unleashing multimodal llm's referential dialogue magic. CoRR  \textbf{abs/2306.15195} (2023)

\bibitem{sugar24}
Chen, S., Garcia, R., Laptev, I., Schmid, C.: Sugar: Pre-training 3d visual representations for robotics. In: Proceedings of the IEEE/CVF Conference on Computer Vision and Pattern Recognition. pp. 18049--18060 (2024)

\bibitem{PALI-X23}
Chen, X., Djolonga, J., Padlewski, P., Mustafa, B., Changpinyo, S., Wu, J., Ruiz, C.R., Goodman, S., Wang, X., Tay, Y., Shakeri, S., Dehghani, M., Salz, D., Lucic, M., Tschannen, M., Nagrani, A., Hu, H., Joshi, M., Pang, B., Montgomery, C., Pietrzyk, P., Ritter, M., Piergiovanni, A.J., Minderer, M., Pavetic, F., Waters, A., Li, G., Alabdulmohsin, I., Beyer, L., Amelot, J., Lee, K., Steiner, A.P., Li, Y., Keysers, D., Arnab, A., Xu, Y., Rong, K., Kolesnikov, A., Seyedhosseini, M., Angelova, A., Zhai, X., Houlsby, N., Soricut, R.: Pali-x: On scaling up a multilingual vision and language model. In: Int. Conf. Learn. Represent. (ICLR) (2023)

\bibitem{Vicuna23}
Chiang, W.L., Li, Z., Lin, Z., Sheng, Y., Wu, Z., Zhang, H., Zheng, L., Zhuang, S., Zhuang, Y., Gonzalez, J.E., Stoica, I., Xing, E.P.: Vicuna: An open-source chatbot impressing gpt-4 with 90\%* chatgpt quality (March 2023), \url{https://lmsys.org/blog/2023-03-30-vicuna/}

\bibitem{ABO22}
Collins, J., Goel, S., Deng, K., Luthra, A., Xu, L., Gundogdu, E., Zhang, X., Vicente, T.F.Y., Dideriksen, T., Arora, H., et~al.: Abo: Dataset and benchmarks for real-world 3d object understanding. In: IEEE/CVF Conf. Comput. Vis. Pattern Recog. (CVPR) (2022)

\bibitem{ScanNet17}
Dai, A., Chang, A.X., Savva, M., Halber, M., Funkhouser, T., Nie{\ss}ner, M.: Scannet: Richly-annotated 3d reconstructions of indoor scenes. In: IEEE/CVF Conf. Comput. Vis. Pattern Recog. (CVPR) (2017)

\bibitem{InstructBLIP23}
Dai, W., Li, J., Li, D., Tiong, A.M.H., Zhao, J., Wang, W., Li, B., Fung, P., Hoi, S.C.H.: Instructblip: Towards general-purpose vision-language models with instruction tuning. In: Adv. Neural Inform. Process. Syst. (NeurIPS) (2023)

\bibitem{VLPHallucination23}
Dai, W., Liu, Z., Ji, Z., Su, D., Fung, P.: Plausible may not be faithful: Probing object hallucination in vision-language pre-training. In: Proceedings of the 17th Conference of the European Chapter of the Association for Computational Linguistics, {EACL} 2023, Dubrovnik, Croatia, May 2-6, 2023 (2023)

\bibitem{flashattention22}
Dao, T., Fu, D., Ermon, S., Rudra, A., R{\'e}, C.: Flashattention: Fast and memory-efficient exact attention with io-awareness. In: Adv. Neural Inform. Process. Syst. (NeurIPS) (2022)

\bibitem{VisDial19}
Das, A., Kottur, S., Gupta, K., Singh, A., Yadav, D., Lee, S., Moura, J.M.F., Parikh, D., Batra, D.: Visual dialog. IEEE Trans. Pattern Anal. Mach. Intell. (TPAMI)  \textbf{41}(5),  1242--1256 (2019)

\bibitem{CommonSenseLM19}
Davison, J., Feldman, J., Rush, A.M.: Commonsense knowledge mining from pretrained models. In: Proceedings of the 2019 Conference on Empirical Methods in Natural Language Processing and the 9th International Joint Conference on Natural Language Processing, {EMNLP-IJCNLP} 2019, Hong Kong, China, November 3-7, 2019 (2019)

\bibitem{objaverse23}
Deitke, M., Schwenk, D., Salvador, J., Weihs, L., Michel, O., VanderBilt, E., Schmidt, L., Ehsani, K., Kembhavi, A., Farhadi, A.: Objaverse: A universe of annotated 3d objects. In: IEEE/CVF Conf. Comput. Vis. Pattern Recog. (CVPR) (2023)

\bibitem{voxelrcnn21}
Deng, J., Shi, S., Li, P., Zhou, W., Zhang, Y., Li, H.: Voxel r-cnn: Towards high performance voxel-based 3d object detection. In: AAAI Conf. Artif. Intell. (AAAI) (2021)

\bibitem{BERT}
Devlin, J., Chang, M., Lee, K., Toutanova, K.: {BERT:} pre-training of deep bidirectional transformers for language understanding. In: Proceedings of the 2019 Conference of the North American Chapter of the Association for Computational Linguistics: Human Language Technologies, {NAACL-HLT} 2019, Minneapolis, MN, USA, June 2-7, 2019, Volume 1 (Long and Short Papers). pp. 4171--4186. Association for Computational Linguistics (2019)

\bibitem{PLA23}
Ding, R., Yang, J., Xue, C., Zhang, W., Bai, S., Qi, X.: {PLA:} language-driven open-vocabulary 3d scene understanding. In: IEEE/CVF Conf. Comput. Vis. Pattern Recog. (CVPR) (2023)

\bibitem{Lowis3D23}
Ding, R., Yang, J., Xue, C., Zhang, W., Bai, S., Qi, X.: Lowis3d: Language-driven open-world instance-level 3d scene understanding. IEEE Trans. Pattern Anal. Mach. Intell. (TPAMI) pp. 1--16 (2024)

\bibitem{LLM-GROP23}
Ding, Y., Zhang, X., Paxton, C., Zhang, S.: Task and motion planning with large language models for object rearrangement. In: IEEE/RSJ Int. Conf. Intell. Robot. and Syst. (IROS) (2023)

\bibitem{DreamLLM23}
Dong, R., Han, C., Peng, Y., Qi, Z., Ge, Z., Yang, J., Zhao, L., Sun, J., Zhou, H., Wei, H., Kong, X., Zhang, X., Ma, K., Yi, L.: Dream{LLM}: Synergistic multimodal comprehension and creation. In: Int. Conf. Learn. Represent. (ICLR) (2024)

\bibitem{ACT23}
Dong, R., Qi, Z., Zhang, L., Zhang, J., Sun, J., Ge, Z., Yi, L., Ma, K.: Autoencoders as cross-modal teachers: Can pretrained 2d image transformers help 3d representation learning? In: Int. Conf. Learn. Represent. (ICLR) (2023)

\bibitem{DGMS23}
Dong, R., Tan, Z., Wu, M., Zhang, L., Ma, K.: Finding the task-optimal low-bit sub-distribution in deep neural networks. In: Int. Conf. Mach. Learn. (ICML) (2022)

\bibitem{ViT}
Dosovitskiy, A., Beyer, L., Kolesnikov, A., Weissenborn, D., Zhai, X., Unterthiner, T., Dehghani, M., Minderer, M., Heigold, G., Gelly, S., Uszkoreit, J., Houlsby, N.: An image is worth 16x16 words: Transformers for image recognition at scale. In: Int. Conf. Learn. Represent. (ICLR) (2021)

\bibitem{PaLME23}
Driess, D., Xia, F., Sajjadi, M.S.M., Lynch, C., Chowdhery, A., Ichter, B., Wahid, A., Tompson, J., Vuong, Q., Yu, T., Huang, W., Chebotar, Y., Sermanet, P., Duckworth, D., Levine, S., Vanhoucke, V., Hausman, K., Toussaint, M., Greff, K., Zeng, A., Mordatch, I., Florence, P.: Palm-e: An embodied multimodal language model. In: Int. Conf. Mach. Learn. (ICML) (2023)

\bibitem{PointTrans21}
Engel, N., Belagiannis, V., Dietmayer, K.: Point transformer. {IEEE} Access  \textbf{9},  134826--134840 (2021)

\bibitem{PointGCC23}
Fan, G., Qi, Z., Shi, W., Ma, K.: Point-gcc: Universal self-supervised 3d scene pre-training via geometry-color contrast. CoRR  \textbf{abs/2305.19623} (2023)

\bibitem{ChamferDistance17}
Fan, H., Su, H., Guibas, L.J.: A point set generation network for 3d object reconstruction from a single image. In: IEEE/CVF Conf. Comput. Vis. Pattern Recog. (CVPR) (2017)

\bibitem{3d-future21}
Fu, H., Jia, R., Gao, L., Gong, M., Zhao, B., Maybank, S., Tao, D.: 3d-future: 3d furniture shape with texture. International Journal of Computer Vision  \textbf{129},  3313--3337 (2021)

\bibitem{SimCSE21}
Gao, T., Yao, X., Chen, D.: Simcse: Simple contrastive learning of sentence embeddings. In: Proceedings of the 2021 Conference on Empirical Methods in Natural Language Processing, {EMNLP} 2021, Virtual Event / Punta Cana, Dominican Republic, 7-11 November, 2021 (2021)

\bibitem{MixCon3D24}
Gao, Y., Wang, Z., Zheng, W.S., Xie, C., Zhou, Y.: Sculpting holistic 3d representation in contrastive language-image-3d pre-training. In: IEEE/CVF Conf. Comput. Vis. Pattern Recog. (CVPR) (2024)

\bibitem{SEED23}
Ge, Y., Ge, Y., Zeng, Z., Wang, X., Shan, Y.: Planting a {SEED} of vision in large language model. In: Int. Conf. Learn. Represent. (ICLR) (2024)

\bibitem{GengPartManip23}
Geng, H., Li, Z., Geng, Y., Chen, J., Dong, H., Wang, H.: Partmanip: Learning cross-category generalizable part manipulation policy from point cloud observations. In: IEEE/CVF Conf. Comput. Vis. Pattern Recog. (CVPR) (2023)

\bibitem{SAGE23}
Geng, H., Wei, S., Deng, C., Shen, B., Wang, H., Guibas, L.: Sage: Bridging semantic and actionable parts for generalizable articulated-object manipulation under language instructions. In: Robotics: Science and Systems (RSS) (2024)

\bibitem{gapartnet23}
Geng, H., Xu, H., Zhao, C., Xu, C., Yi, L., Huang, S., Wang, H.: Gapartnet: Cross-category domain-generalizable object perception and manipulation via generalizable and actionable parts. In: IEEE/CVF Conf. Comput. Vis. Pattern Recog. (CVPR) (2023)

\bibitem{GengRLAfford23}
Geng, Y., An, B., Geng, H., Chen, Y., Yang, Y., Dong, H.: Rlafford: End-to-end affordance learning for robotic manipulation. In: IEEE Int. Conf. Robot. Autom. (ICRA) (2023)

\bibitem{imagebind23}
Girdhar, R., El-Nouby, A., Liu, Z., Singh, M., Alwala, K.V., Joulin, A., Misra, I.: Imagebind: One embedding space to bind them all. In: Proceedings of the IEEE/CVF Conference on Computer Vision and Pattern Recognition. pp. 15180--15190 (2023)

\bibitem{GengARNOLD23}
Gong, R., Huang, J., Zhao, Y., Geng, H., Gao, X., Wu, Q., Ai, W., Zhou, Z., Terzopoulos, D., Zhu, S., Jia, B., Huang, S.: {ARNOLD:} {A} benchmark for language-grounded task learning with continuous states in realistic 3d scenes. In: Int. Conf. Comput. Vis. (ICCV) (2023)

\bibitem{BenchmarkSSL19}
Goyal, P., Mahajan, D., Gupta, A., Misra, I.: Scaling and benchmarking self-supervised visual representation learning. In: Int. Conf. Comput. Vis. (ICCV). pp. 6390--6399. {IEEE} (2019)

\bibitem{WhatMakesAChair11}
Grabner, H., Gall, J., Gool, L.V.: What makes a chair a chair? In: IEEE/CVF Conf. Comput. Vis. Pattern Recog. (CVPR) (2011)

\bibitem{FDPO23}
Gunjal, A., Yin, J., Bas, E.: Detecting and preventing hallucinations in large vision language models. In: AAAI Conf. Artif. Intell. (AAAI) (2024)

\bibitem{pointbind23}
Guo, Z., Zhang, R., Zhu, X., Tang, Y., Ma, X., Han, J., Chen, K., Gao, P., Li, X., Li, H., Heng, P.: Point-bind {\&} point-llm: Aligning point cloud with multi-modality for 3d understanding, generation, and instruction following. CoRR  \textbf{abs/2309.00615} (2023)

\bibitem{LVIS19}
Gupta, A., Dollar, P., Girshick, R.: Lvis: A dataset for large vocabulary instance segmentation. In: IEEE/CVF Conf. Comput. Vis. Pattern Recog. (CVPR) (2019)

\bibitem{SGRGBD14}
Gupta, S., Girshick, R.B., Arbel{\'{a}}ez, P.A., Malik, J.: Learning rich features from {RGB-D} images for object detection and segmentation. In: Eur. Conf. Comput. Vis. (ECCV) (2014)

\bibitem{VisualProgramming23}
Gupta, T., Kembhavi, A.: Visual programming: Compositional visual reasoning without training. In: IEEE/CVF Conf. Comput. Vis. Pattern Recog. (CVPR) (2023)

\bibitem{MVTN}
Hamdi, A., Giancola, S., Ghanem, B.: {MVTN:} multi-view transformation network for 3d shape recognition. In: Int. Conf. Comput. Vis. (ICCV). pp. 1--11. {IEEE} (2021)

\bibitem{UnifedPETuning21}
He, J., Zhou, C., Ma, X., Berg-Kirkpatrick, T., Neubig, G.: Towards a unified view of parameter-efficient transfer learning. In: Int. Conf. Learn. Represent. (ICLR) (2021)

\bibitem{MAE}
He, K., Chen, X., Xie, S., Li, Y., Doll{\'{a}}r, P., Girshick, R.B.: Masked autoencoders are scalable vision learners. In: IEEE/CVF Conf. Comput. Vis. Pattern Recog. (CVPR) (2022)

\bibitem{GLEU16}
Hendrycks, D., Gimpel, K.: Gaussian error linear units (gelus). CoRR  \textbf{abs/1606.08415} (2016)

\bibitem{3DLLM23}
Hong, Y., Zhen, H., Chen, P., Zheng, S., Du, Y., Chen, Z., Gan, C.: 3d-llm: Injecting the 3d world into large language models. In: Adv. Neural Inform. Process. Syst. (NeurIPS) (2023)

\bibitem{Pri3D21}
Hou, J., Xie, S., Graham, B., Dai, A., Nie{\ss}ner, M.: Pri3d: Can 3d priors help 2d representation learning? In: Int. Conf. Comput. Vis. (ICCV). pp. 5673--5682. {IEEE} (2021)

\bibitem{LoRA22}
Hu, E.J., Shen, Y., Wallis, P., Allen{-}Zhu, Z., Li, Y., Wang, S., Wang, L., Chen, W.: Lora: Low-rank adaptation of large language models. In: Int. Conf. Learn. Represent. (ICLR) (2022)

\bibitem{LearnObjectFunction16}
Hu, R., van Kaick, O., Wu, B., Huang, H., Shamir, A., Zhang, H.: Learning how objects function via co-analysis of interactions. {ACM} Trans. Graph.  \textbf{35}(4),  47:1--47:13 (2016)

\bibitem{SnapshotPartMobility17}
Hu, R., Li, W., van Kaick, O., Shamir, A., Zhang, H., Huang, H.: Learning to predict part mobility from a single static snapshot. {ACM} Trans. Graph.  \textbf{36}(6),  227:1--227:13 (2017)

\bibitem{InteractionContext15}
Hu, R., Zhu, C., van Kaick, O., Liu, L., Shamir, A., Zhang, H.: Interaction context {(ICON):} towards a geometric functionality descriptor. {ACM} Trans. Graph.  \textbf{34}(4),  83:1--83:12 (2015)

\bibitem{Leo23}
Huang, J., Yong, S., Ma, X., Linghu, X., Li, P., Wang, Y., Li, Q., Zhu, S., Jia, B., Huang, S.: An embodied generalist agent in 3d world. In: Int. Conf. Mach. Learn. (ICML) (2024)

\bibitem{CLIP2Point22}
Huang, T., Dong, B., Yang, Y., Huang, X., Lau, R.W.H., Ouyang, W., Zuo, W.: Clip2point: Transfer {CLIP} to point cloud classification with image-depth pre-training. In: Int. Conf. Comput. Vis. (ICCV) (2023)

\bibitem{OnePolicyControlAll20}
Huang, W., Mordatch, I., Pathak, D.: One policy to control them all: Shared modular policies for agent-agnostic control. In: Int. Conf. Mach. Learn. (ICML) (2020)

\bibitem{VoxPoser23}
Huang, W., Wang, C., Zhang, R., Li, Y., Wu, J., Fei{-}Fei, L.: Voxposer: Composable 3d value maps for robotic manipulation with language models. In: Annu. Conf. Robot. Learn. (CoRL) (2023)

\bibitem{EmbodiedReasoningLM22}
Huang, W., Xia, F., Xiao, T., Chan, H., Liang, J., Florence, P., Zeng, A., Tompson, J., Mordatch, I., Chebotar, Y., Sermanet, P., Jackson, T., Brown, N., Luu, L., Levine, S., Hausman, K., Ichter, B.: Inner monologue: Embodied reasoning through planning with language models. In: Annu. Conf. Robot. Learn. (CoRL) (2022)

\bibitem{SayCan22}
Ichter, B., Brohan, A., Chebotar, Y., Finn, C., Hausman, K., Herzog, A., Ho, D., Ibarz, J., Irpan, A., Jang, E., Julian, R., Kalashnikov, D., Levine, S., Lu, Y., Parada, C., Rao, K., Sermanet, P., Toshev, A., Vanhoucke, V., Xia, F., Xiao, T., Xu, P., Yan, M., Brown, N., Ahn, M., Cortes, O., Sievers, N., Tan, C., Xu, S., Reyes, D., Rettinghouse, J., Quiambao, J., Pastor, P., Luu, L., Lee, K., Kuang, Y., Jesmonth, S., Joshi, N.J., Jeffrey, K., Ruano, R.J., Hsu, J., Gopalakrishnan, K., David, B., Zeng, A., Fu, C.K.: Do as {I} can, not as {I} say: Grounding language in robotic affordances. In: Annu. Conf. Robot. Learn. (CoRL) (2022)

\bibitem{OpenCLIP21}
Ilharco, G., Wortsman, M., Wightman, R., Gordon, C., Carlini, N., Taori, R., Dave, A., Shankar, V., Namkoong, H., Miller, J., Hajishirzi, H., Farhadi, A., Schmidt, L.: Openclip (Jul 2021)

\bibitem{CompressLLMs23}
Jaiswal, A., Gan, Z., Du, X., Zhang, B., Wang, Z., Yang, Y.: Compressing llms: The truth is rarely pure and never simple. In: Int. Conf. Learn. Represent. (ICLR) (2024)

\bibitem{VPT22}
Jia, M., Tang, L., Chen, B., Cardie, C., Belongie, S.J., Hariharan, B., Lim, S.: Visual prompt tuning. In: Eur. Conf. Comput. Vis. (ECCV) (2022)

\bibitem{VIMA23}
Jiang, Y., Gupta, A., Zhang, Z., Wang, G., Dou, Y., Chen, Y., Fei{-}Fei, L., Anandkumar, A., Zhu, Y., Fan, L.: {VIMA:} general robot manipulation with multimodal prompts. In: Annu. Conf. Robot. Learn. (CoRL) (2023)

\bibitem{LLMKnowledge20}
Jiang, Z., Xu, F.F., Araki, J., Neubig, G.: How can we know what language models know. Trans. Assoc. Comput. Linguistics  \textbf{8},  423--438 (2020)

\bibitem{StereoMatching94}
Kanade, T., Okutomi, M.: A stereo matching algorithm with an adaptive window: Theory and experiment. {IEEE} Trans. Pattern Anal. Mach. Intell.  \textbf{16}(9),  920--932 (1994)

\bibitem{Shape2Pose14}
Kim, V.G., Chaudhuri, S., Guibas, L.J., Funkhouser, T.A.: Shape2pose: human-centric shape analysis. {ACM} Trans. Graph.  \textbf{33}(4),  120:1--120:12 (2014)

\bibitem{GILL23}
Koh, J.Y., Fried, D., Salakhutdinov, R.: Generating images with multimodal language models. In: Adv. Neural Inform. Process. Syst. (NeurIPS) (2023)

\bibitem{HungarianAlogorithm55}
Kuhn, H.W.: The hungarian method for the assignment problem. Naval research logistics quarterly  \textbf{2}(1-2),  83--97 (1955)

\bibitem{BLIP2_23}
Li, J., Li, D., Savarese, S., Hoi, S.C.H.: {BLIP-2:} bootstrapping language-image pre-training with frozen image encoders and large language models. In: Int. Conf. Mach. Learn. (ICML) (2023)

\bibitem{PrefixTuning21}
Li, X.L., Liang, P.: Prefix-tuning: Optimizing continuous prompts for generation. In: Proceedings of the 59th Annual Meeting of the Association for Computational Linguistics and the 11th International Joint Conference on Natural Language Processing (Volume 1: Long Papers) (2021)

\bibitem{NPCSCatPose20}
Li, X., Wang, H., Yi, L., Guibas, L.J., Abbott, A.L., Song, S.: Category-level articulated object pose estimation. In: IEEE/CVF Conf. Comput. Vis. Pattern Recog. (CVPR) (2020)

\bibitem{li2023manipllm}
Li, X., Zhang, M., Geng, Y., Geng, H., Long, Y., Shen, Y., Zhang, R., Liu, J., Dong, H.: Manipllm: Embodied multimodal large language model for object-centric robotic manipulation (2023)

\bibitem{POPE23}
Li, Y., Du, Y., Zhou, K., Wang, J., Zhao, X., Wen, J.R.: Evaluating object hallucination in large vision-language models. In: Proceedings of the 2023 Conference on Empirical Methods in Natural Language Processing. pp. 292--305. Association for Computational Linguistics, Singapore (2023)

\bibitem{TaskMatrix23}
Liang, Y., Wu, C., Song, T., Wu, W., Xia, Y., Liu, Y., Ou, Y., Lu, S., Ji, L., Mao, S., et~al.: Taskmatrix. ai: Completing tasks by connecting foundation models with millions of apis. Intelligent Computing  \textbf{3}, ~0063 (2024)

\bibitem{ROUGH04}
Lin, C.Y.: Rouge: A package for automatic evaluation of summaries. In: Proc. Workshop on Text Summariation Branches Out, Post-Conference Workshop of ACL 2004 (2004)

\bibitem{Text2Motion23}
Lin, K., Agia, C., Migimatsu, T., Pavone, M., Bohg, J.: Text2motion: From natural language instructions to feasible plans. Autonomous Robots  \textbf{47}(8),  1345--1365 (2023)

\bibitem{GAVIE23}
Liu, F., Lin, K., Li, L., Wang, J., Yacoob, Y., Wang, L.: Aligning large multi-modal model with robust instruction tuning. CoRR  \textbf{abs/2306.14565} (2023)

\bibitem{LLaVA1.523}
Liu, H., Li, C., Li, Y., Lee, Y.J.: Improved baselines with visual instruction tuning. In: IEEE/CVF Conf. Comput. Vis. Pattern Recog. (CVPR) (2024)

\bibitem{LLaVA23}
Liu, H., Li, C., Wu, Q., Lee, Y.J.: Visual instruction tuning. In: Adv. Neural Inform. Process. Syst. (NeurIPS) (2023)

\bibitem{OpenShape23}
Liu, M., Shi, R., Kuang, K., Zhu, Y., Li, X., Han, S., Cai, H., Porikli, F., Su, H.: Openshape: Scaling up 3d shape representation towards open-world understanding. In: Adv. Neural Inform. Process. Syst. (NeurIPS) (2023)

\bibitem{FewArtObjGen23}
Liu, X., Wang, B., Wang, H., Yi, L.: Few-shot physically-aware articulated mesh generation via hierarchical deformation. In: Int. Conf. Comput. Vis. (ICCV) (2023)

\bibitem{GenOHDiffusion24}
Liu, X., Yi, L.: Gene{OH} diffusion: Towards generalizable hand-object interaction denoising via denoising diffusion. In: Int. Conf. Learn. Represent. (ICLR) (2024)

\bibitem{PartLevelSE3Pose23}
Liu, X., Zhang, J., Hu, R., Huang, H., Wang, H., Yi, L.: Self-supervised category-level articulated object pose estimation with part-level {SE(3)} equivariance. In: Int. Conf. Learn. Represent. (ICLR) (2023)

\bibitem{RSCNN}
Liu, Y., Fan, B., Xiang, S., Pan, C.: Relation-shape convolutional neural network for point cloud analysis. In: IEEE/CVF Conf. Comput. Vis. Pattern Recog. (CVPR) (2019)

\bibitem{MMBench23}
Liu, Y., Duan, H., Zhang, Y., Li, B., Zhang, S., Zhao, W., Yuan, Y., Wang, J., He, C., Liu, Z., Chen, K., Lin, D.: Mmbench: Is your multi-modal model an all-around player? CoRR  \textbf{abs/2307.06281} (2023)

\bibitem{SyncDreamer23}
Liu, Y., Lin, C., Zeng, Z., Long, X., Liu, L., Komura, T., Wang, W.: Syncdreamer: Generating multiview-consistent images from a single-view image. In: Int. Conf. Learn. Represent. (ICLR) (2024)

\bibitem{Leaf23}
Liu, Y., Chen, J., Zhang, Z., Huang, J., Yi, L.: Leaf: Learning frames for 4d point cloud sequence understanding. In: Int. Conf. Comput. Vis. (ICCV) (2023)

\bibitem{groupfree21}
Liu, Z., Zhang, Z., Cao, Y., Hu, H., Tong, X.: Group-free 3d object detection via transformers. In: Int. Conf. Comput. Vis. (ICCV) (2021)

\bibitem{CosineLRSGDR}
Loshchilov, I., Hutter, F.: {SGDR:} stochastic gradient descent with warm restarts. In: Int. Conf. Learn. Represent. (ICLR) (2017)

\bibitem{AdamW19}
Loshchilov, I., Hutter, F.: Decoupled weight decay regularization. In: Int. Conf. Learn. Represent. (ICLR) (2019)

\bibitem{ImagePartStates18}
Lu, C., Su, H., Li, Y., Lu, Y., Yi, L., Tang, C., Guibas, L.J.: Beyond holistic object recognition: Enriching image understanding with part states. In: IEEE/CVF Conf. Comput. Vis. Pattern Recog. (CVPR) (2018)

\bibitem{OVIR3D23}
Lu, S., Chang, H., Jing, E.P., Boularias, A., Bekris, K.E.: {OVIR-3D:} open-vocabulary 3d instance retrieval without training on 3d data. In: Annu. Conf. Robot. Learn. (CoRL) (2023)

\bibitem{Cap3D23}
Luo, T., Rockwell, C., Lee, H., Johnson, J.: Scalable 3d captioning with pretrained models. In: Adv. Neural Inform. Process. Syst. (NeurIPS) (2023)

\bibitem{SQA3D23}
Ma, X., Yong, S., Zheng, Z., Li, Q., Liang, Y., Zhu, S., Huang, S.: {SQA3D:} situated question answering in 3d scenes. In: Int. Conf. Learn. Represent. (ICLR) (2023)

\bibitem{PointMLP}
Ma, X., Qin, C., You, H., Ran, H., Fu, Y.: Rethinking network design and local geometry in point cloud: {A} simple residual {MLP} framework. In: Int. Conf. Learn. Represent. (ICLR). OpenReview.net (2022)

\bibitem{BlindSafety17}
MacLeod, H., Bennett, C.L., Morris, M.R., Cutrell, E.: Understanding blind people's experiences with computer-generated captions of social media images. In: Proceedings of the 2017 CHI Conference on Human Factors in Computing Systems. p. 5988–5999. CHI '17, Association for Computing Machinery, New York, NY, USA (2017)

\bibitem{voxeltransformer21}
Mao, J., Xue, Y., Niu, M., Bai, H., Feng, J., Liang, X., Xu, H., Xu, C.: Voxel transformer for 3d object detection. In: Int. Conf. Comput. Vis. (ICCV) (2021)

\bibitem{voxelnet15}
Maturana, D., Scherer, S.A.: Voxnet: {A} 3d convolutional neural network for real-time object recognition. In: IEEE/RSJ Int. Conf. Intell. Robot. and Syst. (IROS). pp. 922--928. {IEEE} (2015)

\bibitem{partnet19}
Mo, K., Zhu, S., Chang, A.X., Yi, L., Tripathi, S., Guibas, L.J., Su, H.: Partnet: A large-scale benchmark for fine-grained and hierarchical part-level 3d object understanding. In: IEEE/CVF Conf. Comput. Vis. Pattern Recog. (CVPR) (2019)

\bibitem{EmbodiedGPT23}
Mu, Y., Zhang, Q., Hu, M., Wang, W., Ding, M., Jin, J., Wang, B., Dai, J., Qiao, Y., Luo, P.: Embodiedgpt: Vision-language pre-training via embodied chain of thought. In: Adv. Neural Inform. Process. Syst. (NeurIPS) (2023)

\bibitem{ChatGPT22}
OpenAI: Introducing chatgpt  (2022), \url{https://openai.com/blog/chatgpt}

\bibitem{GPT4_23}
OpenAI: {GPT-4} technical report. CoRR  \textbf{abs/2303.08774} (2023), \url{https://openai.com/research/gpt-4}

\bibitem{GPT4Vision23}
OpenAI: Gpt-4v(ision) system card (2023), \url{https://openai.com/research/gpt-4v-system-card}

\bibitem{GPT4o24}
OpenAI: Introducing gpt-4o and more tools to chatgpt free users  (2024), \url{https://openai.com/index/gpt-4o-and-more-tools-to-chatgpt-free/}

\bibitem{InstructGPT22}
Ouyang, L., Wu, J., Jiang, X., Almeida, D., Wainwright, C.L., Mishkin, P., Zhang, C., Agarwal, S., Slama, K., Ray, A., Schulman, J., Hilton, J., Kelton, F., Miller, L., Simens, M., Askell, A., Welinder, P., Christiano, P.F., Leike, J., Lowe, R.: Training language models to follow instructions with human feedback. In: Adv. Neural Inform. Process. Syst. (NeurIPS) (2022)

\bibitem{KosmosG23}
Pan, X., Dong, L., Huang, S., Peng, Z., Chen, W., Wei, F.: Kosmos-g: Generating images in context with multimodal large language models. In: Int. Conf. Learn. Represent. (ICLR) (2024)

\bibitem{PointMAE}
Pang, Y., Wang, W., Tay, F.E.H., Liu, W., Tian, Y., Yuan, L.: Masked autoencoders for point cloud self-supervised learning. In: Eur. Conf. Comput. Vis. (ECCV) (2022)

\bibitem{BLEU02}
Papineni, K., Roukos, S., Ward, T., Zhu, W.: Bleu: a method for automatic evaluation of machine translation (2002)

\bibitem{InstructGPT4_23}
Peng, B., Li, C., He, P., Galley, M., Gao, J.: Instruction tuning with {GPT-4}. CoRR  \textbf{abs/2304.03277} (2023)

\bibitem{OpenScene23}
Peng, S., Genova, K., Jiang, C.M., Tagliasacchi, A., Pollefeys, M., Funkhouser, T.A.: Openscene: 3d scene understanding with open vocabularies. In: IEEE/CVF Conf. Comput. Vis. Pattern Recog. (CVPR) (2023)

\bibitem{DreamBenchPlus24}
Peng, Y., Cui, Y., Tang, H., Qi, Z., Dong, R., Bai, J., Han, C., Ge, Z., Zhang, X., Xia, S.T.: Dreambench++: A human-aligned benchmark for personalized image generation. CoRR  \textbf{abs/2406.16855} (2024)

\bibitem{Kosmos2_23}
Peng, Z., Wang, W., Dong, L., Hao, Y., Huang, S., Ma, S., Wei, F.: Kosmos-2: Grounding multimodal large language models to the world. CoRR  \textbf{abs/2306.14824} (2023)

\bibitem{LMKnowledgeBase19}
Petroni, F., Rockt{\"{a}}schel, T., Riedel, S., Lewis, P.S.H., Bakhtin, A., Wu, Y., Miller, A.H.: Language models as knowledge bases? In: Proceedings of the 2019 Conference on Empirical Methods in Natural Language Processing and the 9th International Joint Conference on Natural Language Processing, {EMNLP-IJCNLP} 2019, Hong Kong, China, November 3-7, 2019 (2019)

\bibitem{InteractionLandScapes17}
Pirk, S., Krs, V., Hu, K., Rajasekaran, S.D., Kang, H., Yoshiyasu, Y., Benes, B., Guibas, L.J.: Understanding and exploiting object interaction landscapes. {ACM} Trans. Graph.  \textbf{36}(3),  31:1--31:14 (2017)

\bibitem{PointNet}
Qi, C.R., Su, H., Mo, K., Guibas, L.J.: Pointnet: Deep learning on point sets for 3d classification and segmentation. In: IEEE/CVF Conf. Comput. Vis. Pattern Recog. (CVPR). pp. 77--85 (2017)

\bibitem{PointNet++}
Qi, C.R., Yi, L., Su, H., Guibas, L.J.: Pointnet++: Deep hierarchical feature learning on point sets in a metric space. In: Adv. Neural Inform. Process. Syst. (NIPS). pp. 5099--5108 (2017)

\bibitem{InHandRotate23}
Qi, H., Kumar, A., Calandra, R., Ma, Y., Malik, J.: In-hand object rotation via rapid motor adaptation. In: Annu. Conf. Robot. Learn. (CoRL) (2023)

\bibitem{ReCon23}
Qi, Z., Dong, R., Fan, G., Ge, Z., Zhang, X., Ma, K., Yi, L.: Contrast with reconstruct: Contrastive 3d representation learning guided by generative pretraining. In: Int. Conf. Mach. Learn. (ICML) (2023)

\bibitem{VPP23}
Qi, Z., Yu, M., Dong, R., Ma, K.: {VPP:} efficient conditional 3d generation via voxel-point progressive representation. In: Adv. Neural Inform. Process. Syst. (NeurIPS) (2023)

\bibitem{PointNext}
Qian, G., Li, Y., Peng, H., Mai, J., Hammoud, H.A.A.K., Elhoseiny, M., Ghanem, B.: Pointnext: Revisiting pointnet++ with improved training and scaling strategies. In: Adv. Neural Inform. Process. Syst. (NeurIPS) (2022)

\bibitem{CLIP}
Radford, A., Kim, J.W., Hallacy, C., Ramesh, A., Goh, G., Agarwal, S., Sastry, G., Askell, A., Mishkin, P., Clark, J., Krueger, G., Sutskever, I.: Learning transferable visual models from natural language supervision. In: Int. Conf. Mach. Learn. (ICML). Proceedings of Machine Learning Research, vol.~139, pp. 8748--8763. {PMLR} (2021)

\bibitem{GPT1_18}
Radford, A., Narasimhan, K., Salimans, T., Sutskever, I.: Improving language understanding by generative pre-training  (2018)

\bibitem{GPT2_19}
Radford, A., Wu, J., Child, R., Luan, D., Amodei, D., Sutskever, I.: Language models are unsupervised multitask learners. OpenAI blog  \textbf{1}(8), ~9 (2019)

\bibitem{SBERT19}
Reimers, N., Gurevych, I.: Sentence-bert: Sentence embeddings using siamese bert-networks. In: Proceedings of the 2019 Conference on Empirical Methods in Natural Language Processing and the 9th International Joint Conference on Natural Language Processing, {EMNLP-IJCNLP} 2019, Hong Kong, China, November 3-7, 2019 (2019)

\bibitem{ModelNetC22}
Ren, J., Pan, L., Liu, Z.: Benchmarking and analyzing point cloud classification under corruptions. In: Int. Conf. Mach. Learn. (ICML) (2022)

\bibitem{VisualHallucination18}
Rohrbach, A., Hendricks, L.A., Burns, K., Darrell, T., Saenko, K.: Object hallucination in image captioning. In: Proceedings of the 2018 Conference on Empirical Methods in Natural Language Processing, Brussels, Belgium, October 31 - November 4, 2018 (2018)

\bibitem{ObjectHallucination18}
Rohrbach, A., Hendricks, L.A., Burns, K., Darrell, T., Saenko, K.: Object hallucination in image captioning. In: Proceedings of the 2018 Conference on Empirical Methods in Natural Language Processing (2018)

\bibitem{DistillFeatureNeRFGrasp23}
Shen, W., Yang, G., Yu, A., Wong, J., Kaelbling, L.P., Isola, P.: Distilled feature fields enable few-shot language-guided manipulation. In: Annu. Conf. Robot. Learn. (CoRL) (2023)

\bibitem{HuggingGPT23}
Shen, Y., Song, K., Tan, X., Li, D., Lu, W., Zhuang, Y.: Hugginggpt: Solving {AI} tasks with chatgpt and its friends in huggingface. In: Adv. Neural Inform. Process. Syst. (NeurIPS) (2023)

\bibitem{RoboCook23}
Shi, H., Xu, H., Clarke, S., Li, Y., Wu, J.: Robocook: Long-horizon elasto-plastic object manipulation with diverse tools. In: Annu. Conf. Robot. Learn. (CoRL) (2023)

\bibitem{Turbosquid}
Shutterstock: Turbosquid. \url{https://www.turbosquid.com/}

\bibitem{MVCNN3D15}
Su, H., Maji, S., Kalogerakis, E., Learned{-}Miller, E.G.: Multi-view convolutional neural networks for 3d shape recognition. In: Int. Conf. Comput. Vis. (ICCV) (2015)

\bibitem{ModelNet40C22}
Sun, J., Zhang, Q., Kailkhura, B., Yu, Z., Xiao, C., Mao, Z.M.: Modelnet40-c: A robustness benchmark for 3d point cloud recognition under corruption. In: ICLR 2022 Workshop on Socially Responsible Machine Learning

\bibitem{Emu223}
Sun, Q., Cui, Y., Zhang, X., Zhang, F., Yu, Q., Luo, Z., Wang, Y., Rao, Y., Liu, J., Huang, T., Wang, X.: Generative multimodal models are in-context learners. In: IEEE/CVF Conf. Comput. Vis. Pattern Recog. (CVPR) (2024)

\bibitem{EVACLIP23}
Sun, Q., Fang, Y., Wu, L., Wang, X., Cao, Y.: {EVA-CLIP:} improved training techniques for {CLIP} at scale. CoRR  \textbf{abs/2303.15389} (2023)

\bibitem{Emu23}
Sun, Q., Yu, Q., Cui, Y., Zhang, F., Zhang, X., Wang, Y., Gao, H., Liu, J., Huang, T., Wang, X.: Emu: Generative pretraining in multimodality. In: Int. Conf. Learn. Represent. (ICLR) (2024)

\bibitem{ViperGPT23}
Sur{\'{\i}}s, D., Menon, S., Vondrick, C.: Vipergpt: Visual inference via python execution for reasoning. In: Int. Conf. Comput. Vis. (ICCV) (2023)

\bibitem{Alpaca23}
Taori, R., Gulrajani, I., Zhang, T., Dubois, Y., Li, X., Guestrin, C., Liang, P., Hashimoto, T.B.: Stanford alpaca: An instruction-following llama model. \url{https://github.com/tatsu-lab/stanford_alpaca} (2023)

\bibitem{LLaMA23}
Touvron, H., Lavril, T., Izacard, G., Martinet, X., Lachaux, M., Lacroix, T., Rozi{\`{e}}re, B., Goyal, N., Hambro, E., Azhar, F., Rodriguez, A., Joulin, A., Grave, E., Lample, G.: Llama: Open and efficient foundation language models. CoRR  \textbf{abs/2302.13971} (2023)

\bibitem{ScanObjectNN19}
Uy, M.A., Pham, Q.H., Hua, B.S., Nguyen, T., Yeung, S.K.: Revisiting point cloud classification: A new benchmark dataset and classification model on real-world data. In: IEEE/CVF Conf. Comput. Vis. Pattern Recog. (CVPR). pp. 1588--1597 (2019)

\bibitem{SLTheory98}
Vapnik, V.: Statistical learning theory. Wiley (1998)

\bibitem{AttentionIsAllYouNeed}
Vaswani, A., Shazeer, N., Parmar, N., Uszkoreit, J., Jones, L., Gomez, A.N., Kaiser, L., Polosukhin, I.: Attention is all you need. In: Adv. Neural Inform. Process. Syst. (NIPS). pp. 5998--6008 (2017)

\bibitem{UniDexGrasp23++23}
Wan, W., Geng, H., Liu, Y., Shan, Z., Yang, Y., Yi, L., Wang, H.: Unidexgrasp++: Improving dexterous grasping policy learning via geometry-aware curriculum and iterative generalist-specialist learning. In: Int. Conf. Comput. Vis. (ICCV) (2023)

\bibitem{Voyager23}
Wang, G., Xie, Y., Jiang, Y., Mandlekar, A., Xiao, C., Zhu, Y., Fan, L., Anandkumar, A.: Voyager: An open-ended embodied agent with large language models. T. Mach. Learn. Res. (TMLR)  (2024)

\bibitem{NOCSPose19}
Wang, H., Sridhar, S., Huang, J., Valentin, J., Song, S., Guibas, L.J.: Normalized object coordinate space for category-level 6d object pose and size estimation. In: IEEE/CVF Conf. Comput. Vis. Pattern Recog. (CVPR) (2019)

\bibitem{DGCNN}
Wang, Y., Sun, Y., Liu, Z., Sarma, S.E., Bronstein, M.M., Solomon, J.M.: Dynamic graph {CNN} for learning on point clouds. ACM Trans. Graph.  \textbf{38}(5),  146:1--146:12 (2019)

\bibitem{takeaphoto23}
Wang, Z., Yu, X., Rao, Y., Zhou, J., Lu, J.: Take-a-photo: 3d-to-2d generative pre-training of point cloud models. In: Int. Conf. Comput. Vis. (ICCV) (2023)

\bibitem{PrimitiveTransformer22}
Wen, H., Liu, Y., Huang, J., Duan, B., Yi, L.: Point primitive transformer for long-term 4d point cloud video understanding. In: Eur. Conf. Comput. Vis. (ECCV) (2022)

\bibitem{CAPTRA21}
Weng, Y., Wang, H., Zhou, Q., Qin, Y., Duan, Y., Fan, Q., Chen, B., Su, H., Guibas, L.J.: {CAPTRA:} category-level pose tracking for rigid and articulated objects from point clouds. In: Int. Conf. Comput. Vis. (ICCV) (2021)

\bibitem{VisualChatGPT23}
Wu, C., Yin, S., Qi, W., Wang, X., Tang, Z., Duan, N.: Visual chatgpt: Talking, drawing and editing with visual foundation models. CoRR  \textbf{abs/2303.04671} (2023)

\bibitem{NextGPT23}
Wu, S., Fei, H., Qu, L., Ji, W., Chua, T.: Next-gpt: Any-to-any multimodal {LLM}. In: Int. Conf. Mach. Learn. (ICML) (2024)

\bibitem{GPT4V3DEval24}
Wu, T., Yang, G., Li, Z., Zhang, K., Liu, Z., Guibas, L.J., Lin, D., Wetzstein, G.: Gpt-4v(ision) is a human-aligned evaluator for text-to-3d generation. In: IEEE/CVF Conf. Comput. Vis. Pattern Recog. (CVPR) (2024)

\bibitem{ModelNet15}
Wu, Z., Song, S., Khosla, A., Yu, F., Zhang, L., Tang, X., Xiao, J.: 3d shapenets: A deep representation for volumetric shapes. In: IEEE/CVF Conf. Comput. Vis. Pattern Recog. (CVPR). pp. 1912--1920 (2015)

\bibitem{PointContrast20}
Xie, S., Gu, J., Guo, D., Qi, C.R., Guibas, L.J., Litany, O.: Pointcontrast: Unsupervised pre-training for 3d point cloud understanding. In: Eur. Conf. Comput. Vis. (ECCV). Lecture Notes in Computer Science, vol. 12348, pp. 574--591. Springer (2020)

\bibitem{pointllm23}
Xu, R., Wang, X., Wang, T., Chen, Y., Pang, J., Lin, D.: Pointllm: Empowering large language models to understand point clouds. CoRR  \textbf{abs/2308.16911} (2023)

\bibitem{UniDexGrasp23}
Xu, Y., Wan, W., Zhang, J., Liu, H., Shan, Z., Shen, H., Wang, R., Geng, H., Weng, Y., Chen, J., Liu, T., Yi, L., Wang, H.: Unidexgrasp: Universal robotic dexterous grasping via learning diverse proposal generation and goal-conditioned policy. In: IEEE/CVF Conf. Comput. Vis. Pattern Recog. (CVPR) (2023)

\bibitem{MultiInstruct23}
Xu, Z., Shen, Y., Huang, L.: Multiinstruct: Improving multi-modal zero-shot learning via instruction tuning. In: Proceedings of the 61st Annual Meeting of the Association for Computational Linguistics (ACL) (Volume 1: Long Papers) (2023)

\bibitem{ULIP22}
Xue, L., Gao, M., Xing, C., Mart{\'{\i}}n{-}Mart{\'{\i}}n, R., Wu, J., Xiong, C., Xu, R., Niebles, J.C., Savarese, S.: {ULIP:} learning unified representation of language, image and point cloud for 3d understanding. In: IEEE/CVF Conf. Comput. Vis. Pattern Recog. (CVPR) (2023)

\bibitem{ULIP2_23}
Xue, L., Yu, N., Zhang, S., Li, J., Mart{\'{\i}}n{-}Mart{\'{\i}}n, R., Wu, J., Xiong, C., Xu, R., Niebles, J.C., Savarese, S.: {ULIP-2:} towards scalable multimodal pre-training for 3d understanding. In: IEEE/CVF Conf. Comput. Vis. Pattern Recog. (CVPR) (2024)

\bibitem{GPT4Tools23}
Yang, R., Song, L., Li, Y., Zhao, S., Ge, Y., Li, X., Shan, Y.: Gpt4tools: Teaching large language model to use tools via self-instruction. In: Adv. Neural Inform. Process. Syst. (NeurIPS) (2023)

\bibitem{MM-REACT23}
Yang, Z., Li, L., Wang, J., Lin, K., Azarnasab, E., Ahmed, F., Liu, Z., Liu, C., Zeng, M., Wang, L.: {MM-REACT:} prompting chatgpt for multimodal reasoning and action. CoRR  \textbf{abs/2303.11381} (2023)

\bibitem{mPLUG-Owl23}
Ye, Q., Xu, H., Xu, G., Ye, J., Yan, M., Zhou, Y., Wang, J., Hu, A., Shi, P., Shi, Y., Li, C., Xu, Y., Chen, H., Tian, J., Qi, Q., Zhang, J., Huang, F.: mplug-owl: Modularization empowers large language models with multimodality. CoRR  \textbf{abs/2304.14178} (2023)

\bibitem{ThreeDQA22}
Ye, S., Chen, D., Han, S., Liao, J.: 3d question answering. IEEE Transactions on Visualization and Computer Graphics  (2022)

\bibitem{DeepPartInduction18}
Yi, L., Huang, H., Liu, D., Kalogerakis, E., Su, H., Guibas, L.J.: Deep part induction from articulated object pairs. {ACM} Trans. Graph.  \textbf{37}(6), ~209 (2018)

\bibitem{ShapeNetPart16}
Yi, L., Kim, V.G., Ceylan, D., Shen, I.C., Yan, M., Su, H., Lu, C., Huang, Q., Sheffer, A., Guibas, L.: A scalable active framework for region annotation in 3d shape collections. ACM Trans. Graph.  \textbf{35}(6),  1--12 (2016)

\bibitem{SyncSpecCNN17}
Yi, L., Su, H., Guo, X., Guibas, L.J.: Syncspeccnn: Synchronized spectral {CNN} for 3d shape segmentation. In: IEEE/CVF Conf. Comput. Vis. Pattern Recog. (CVPR) (2017)

\bibitem{GengDonut23}
You, Y., Shen, B., Deng, C., Geng, H., Wang, H., Guibas, L.J.: Make a donut: Language-guided hierarchical emd-space planning for zero-shot deformable object manipulation. CoRR  \textbf{abs/2311.02787} (2023)

\bibitem{MMVet23}
Yu, W., Yang, Z., Li, L., Wang, J., Lin, K., Liu, Z., Wang, X., Wang, L.: Mm-vet: Evaluating large multimodal models for integrated capabilities. In: Int. Conf. Mach. Learn. (ICML) (2024)

\bibitem{PointBERT}
Yu, X., Tang, L., Rao, Y., Huang, T., Zhou, J., Lu, J.: Point-bert: Pre-training 3d point cloud transformers with masked point modeling. In: IEEE/CVF Conf. Comput. Vis. Pattern Recog. (CVPR) (2022)

\bibitem{point2vec23}
Zeid, K.A., Schult, J., Hermans, A., Leibe, B.: Point2vec for self-supervised representation learning on point clouds. In: DAGM German Conference on Pattern Recognition. pp. 131--146. Springer (2023)

\bibitem{CLIPFO3D23}
Zhang, J., Dong, R., Ma, K.: {CLIP-FO3D:} learning free open-world 3d scene representations from 2d dense {CLIP}. In: Int. Conf. Comput. Vis. Worksh. (ICCV Workshop) (2023)

\bibitem{SelfKDPAMI22}
Zhang, L., Bao, C., Ma, K.: Self-distillation: Towards efficient and compact neural networks. {IEEE} Trans. Pattern Anal. Mach. Intell.  \textbf{44}(8),  4388--4403 (2022)

\bibitem{ReKo23}
Zhang, L., Chen, X., Dong, R., Ma, K.: Region-aware knowledge distillation for efficient image-to-image translation. In: Brit. Mach. Vis. Conf. (BMVC) (2023)

\bibitem{PointDistiller22}
Zhang, L., Dong, R., Tai, H., Ma, K.: Pointdistiller: Structured knowledge distillation towards efficient and compact 3d detection. In: IEEE/CVF Conf. Comput. Vis. Pattern Recog. (CVPR) (2023)

\bibitem{PointM2AE22}
Zhang, R., Guo, Z., Gao, P., Fang, R., Zhao, B., Wang, D., Qiao, Y., Li, H.: Point-m2{AE}: Multi-scale masked autoencoders for hierarchical point cloud pre-training. In: Adv. Neural Inform. Process. Syst. (NeurIPS) (2022)

\bibitem{PointCLIP22}
Zhang, R., Guo, Z., Zhang, W., Li, K., Miao, X., Cui, B., Qiao, Y., Gao, P., Li, H.: Pointclip: Point cloud understanding by {CLIP}. In: IEEE/CVF Conf. Comput. Vis. Pattern Recog. (CVPR) (2022)

\bibitem{LLaMA-Adapter23}
Zhang, R., Han, J., Zhou, A., Hu, X., Yan, S., Lu, P., Li, H., Gao, P., Qiao, Y.: Llama-adapter: Efficient fine-tuning of language models with zero-init attention. In: Int. Conf. Learn. Represent. (ICLR) (2024)

\bibitem{I2PMAE23}
Zhang, R., Wang, L., Qiao, Y., Gao, P., Li, H.: Learning 3d representations from 2d pre-trained models via image-to-point masked autoencoders. In: IEEE/CVF Conf. Comput. Vis. Pattern Recog. (CVPR) (2023)

\bibitem{GPT4ROI23}
Zhang, S., Sun, P., Chen, S., Xiao, M., Shao, W., Zhang, W., Chen, K., Luo, P.: Gpt4roi: Instruction tuning large language model on region-of-interest. CoRR  \textbf{abs/2307.03601} (2023)

\bibitem{HallucinationSurvey23}
Zhang, Y., Li, Y., Cui, L., Cai, D., Liu, L., Fu, T., Huang, X., Zhao, E., Zhang, Y., Chen, Y., Wang, L., Luu, A.T., Bi, W., Shi, F., Shi, S.: Siren's song in the {AI} ocean: {A} survey on hallucination in large language models. CoRR  \textbf{abs/2309.01219} (2023)

\bibitem{TAMM24}
Zhang, Z., Cao, S., Wang, Y.: {TAMM:} triadapter multi-modal learning for 3d shape understanding. In: IEEE/CVF Conf. Comput. Vis. Pattern Recog. (CVPR) (2024)

\bibitem{ChatSpot23}
Zhao, L., Yu, E., Ge, Z., Yang, J., Wei, H., Zhou, H., Sun, J., Peng, Y., Dong, R., Han, C., Zhang, X.: Chatspot: Bootstrapping multimodal llms via precise referring instruction tuning. In: Int. Joint Conf. Artif. Intell. (IJCAI) (2024)

\bibitem{BisectorInteraction3D14}
Zhao, X., Wang, H., Komura, T.: Indexing 3d scenes using the interaction bisector surface. {ACM} Trans. Graph.  \textbf{33}(3),  22:1--22:14 (2014)

\bibitem{CAMS23}
Zheng, J., Zheng, Q., Fang, L., Liu, Y., Yi, L.: {CAMS:} canonicalized manipulation spaces for category-level functional hand-object manipulation synthesis. In: IEEE/CVF Conf. Comput. Vis. Pattern Recog. (CVPR) (2023)

\bibitem{ShareGPT23}
Zheng, L., Chiang, W., Sheng, Y., Zhuang, S., Wu, Z., Zhuang, Y., Lin, Z., Li, Z., Li, D., Xing, E.P., Zhang, H., Gonzalez, J.E., Stoica, I.: Judging llm-as-a-judge with mt-bench and chatbot arena. In: Adv. Neural Inform. Process. Syst. (NeurIPS) (2024)

\bibitem{Uni3D23}
Zhou, J., Wang, J., Ma, B., Liu, Y., Huang, T., Wang, X.: Uni3d: Exploring unified 3d representation at scale. In: Int. Conf. Learn. Represent. (ICLR) (2024)

\bibitem{ZhouHall24}
Zhou, Y., Cui, C., Yoon, J., Zhang, L., Deng, Z., Finn, C., Bansal, M., Yao, H.: Analyzing and mitigating object hallucination in large vision-language models. In: Int. Conf. Learn. Represent. (ICLR) (2024)

\bibitem{MiniGPT4_23}
Zhu, D., Chen, J., Shen, X., Li, X., Elhoseiny, M.: Minigpt-4: Enhancing vision-language understanding with advanced large language models. In: Int. Conf. Learn. Represent. (ICLR) (2024)

\bibitem{pointclipv223}
Zhu, X., Zhang, R., He, B., Guo, Z., Zeng, Z., Qin, Z., Zhang, S., Gao, P.: Pointclip v2: Prompting clip and gpt for powerful 3d open-world learning. In: Int. Conf. Comput. Vis. (ICCV) (2023)

\bibitem{3DVista23}
Zhu, Z., Ma, X., Chen, Y., Deng, Z., Huang, S., Li, Q.: 3d-vista: Pre-trained transformer for 3d vision and text alignment. In: Int. Conf. Comput. Vis. (ICCV) (2023)

\end{thebibliography}
}

\newpage
\appendix
\section{Additional Experiments}\label{app:add_exp}
\subsection{\shapellm\ Architecture}
Let $\mathcal{F}_{\theta}$ be the multimodal LLM parameterized by $\theta$, we use a \recon\ encoder $\mathcal{H_{\phi}}$ as \shapellm's 3D point cloud encoder, followed by three MLP projection layers $\mathcal{M}_{\zeta^{\text{local}}}$ and $\mathcal{M}_{\zeta^{\text{global}}}$ for 3D embedding projection of \recon's local and global representations, respectively.
To facilitate geometry-necessary tasks like 6-DoF pose estimation, we use absolute position encoding (APE) with an MLP projection $\mathcal{M}_{\zeta^{\text{APE}}}$ to provide additional precise low-level geometric information.
Given the original 3D point cloud inputs $\mathcal{P}=\{\mathbf{p}_i | i=1,2,\dots,N\}\in \mathbb{R}^{N\times3}$ with $N$ coordinates encoded in a $(x,y,z)$ Cartesian space.
Following previous works~\cite{PointBERT,ACT23,ReCon23}, $N_s$ seed points are first sampled using farthest point sampling (FPS). 
The point cloud $\mathcal{P}$ is then grouped into $N_s$ neighborhoods $\mathcal{N} = \{\mathcal{N}_i|i=1,2,\dots,N_s\}\in\mathbb{R}^{N_s\times K\times 3}$ with group centroids from the seed point set $\mathcal{P}^s$.
The APE representation can be written as
\begin{equation}
    \mathbf{E}_{\text{APE}} = \mathcal{M}_{\zeta^{\text{APE}}} \circ \mathcal{P}^s.
\end{equation}
The local and transformation-invariant 3D embeddings $\mathbf{x}_i = \mathop{\text{MAX}}\limits_{\mathbf{p}_{i,j}\in{\mathcal{N}_i}}\big(\Phi_{\gamma}\left(\xi_{i,j}\right)\big )$ for $\mathcal{P}_i^s, i=1,2,\dots,N_s$ is used as 3D token embeddings of \recon, where $\Phi_\gamma$ is a per-point MLP point feature extractor~\cite{PointNet,PointNet++} and $\xi_{i,j}$ is the feature of $j$-th neighbour point $\mathbf{p}_{i,j}$ in the neighbourhood $\mathcal{N}_i$.
Let $\{\mathbf{g}_q^{\text{image}}\}_{q=1}^G$ be $G$ multi-view image global queries and $\mathbf{g}^{\text{text}}$ be the global text query.
\recon\ outputs the local and global 3D point cloud representations by taking 3D embeddings and global queries as inputs:
\begin{equation}
    \Big[\mathbf{e}_{\text{local}}, \mathbf{e}_{\text{global}}\Big] = \Big[\mathcal{H}_{\phi}\left([\mathcal{P}^s, \{\mathbf{g}_q^{\text{image}}\}_{q=1}^G, \mathbf{g}^{\text{text}}]\right)\Big],
\end{equation}
and the representation to \shapellm\ is:
\begin{equation}
    [\mathbf{E}_{\text{local}}, \mathbf{E}_{\text{global}}] = [\mathcal{M}_{\zeta^{\text{local}}} \circ \mathbf{e}_{\text{local}}, \mathcal{M}_{\zeta^{\text{global}}} \circ \mathbf{e}_{\text{global}}].
\end{equation}
In addition, inspired by prefix-tuning~\cite{PrefixTuning21} and dream queries~\cite{DreamLLM23}, we append $Q$-length learnable embeddings $\{\mathbf{d}^{\text{APE}}_q\}_{q=1}^Q$, $\{\mathbf{d}^{\text{local}}_q\}_{q=1}^Q$, $\{\mathbf{d}^{\text{global}}_q\}_{q=1}^Q$ as visual prompts representation~\cite{VPP23} $\mathbf{E}_{\text{prompt}}$ for adaptively modulating different semantic information encoded in APE, local and global \recon\ representations, respectively. 

Formally, the encoded 3D representations to \shapellm\ can be written as:
\begin{equation}
    \left[\{\mathbf{d}^{\text{APE}}_q\}_{q=1}^Q,\mathbf{E}_{\text{APE}},\{\mathbf{d}^{\text{local}}_q\}_{q=1}^Q,\mathbf{E}_{\text{local}},\{\mathbf{d}^{\text{global}}_q\}_{q=1}^Q,\mathbf{E}_{\text{global}}\right].
\end{equation}

\subsubsection{Input Components}~
\cref{tab:ablation_shapellm} shows the ablation study of each input component by supervised fine-tuning with different input representations, demonstrating that it is necessary to employ all designs for achieving decent performance on both 3D comprehension and real-world grounding.
\begin{table}[t!]
\setlength{\tabcolsep}{6pt}
  \caption{\textbf{Ablation study on the dedicated designs of \shapellm\ architecture}. 
  The performance of multimodal comprehension on 3D MM-Vet and referring expression grounding on GAPartNet with \shapellm-13B is reported. Note that $\mathbf{E}_{\text{global}}$ is calculated with both global queries and cross-attention with local 3D embeddings.
  }
  \label{tab:ablation_shapellm}
  \centering
    \vspace{-2pt}
    \resizebox{0.77\linewidth}{!}{
    \begin{tabular}{cccccc}
    \toprule[0.95pt]
        $\mathbf{E}_{\text{APE}}$ & $\mathbf{E}_{\text{prompt}}$ & $\mathbf{E}_{\text{local}}$ & $\mathbf{E}_{\text{global}}$ & \textbf{3D MM-Vet} & \textbf{GAPartNet} \\
        \midrule[0.6pt]
        \cmark & \xmark & \xmark & \xmark & 30.8 & \textbf{12.3}\\
        \cmark & \cmark & \xmark & \xmark & 32.0 & 11.4\\
        \xmark & \xmark & \cmark & \xmark & 42.2 & 10.0\\
        \xmark & \cmark & \xmark & \cmark & 50.3 & 10.5\\
        \cmark & \xmark & \cmark & \cmark & 52.3 & 10.5\\
        \cmark & \cmark & \xmark & \cmark & 50.3 & 11.7\\
        \xmark & \xmark & \xmark & \cmark & 52.4 & 11.7\\
        \xmark & \xmark & \cmark & \cmark & 49.6 & 10.1\\
        \xmark & \cmark & \cmark & \cmark & 51.7 & 10.1\\
        \cmark & \cmark & \cmark & \cmark & \textbf{53.1} & 11.7\\
        \bottomrule[0.95pt]
    \end{tabular}
    }
\vspace{-5pt}
\end{table}

\subsubsection{Visual Prompt Number}
\begin{figure}[h!]
\centering
\vspace{-5pt}
\includegraphics[width=0.5\linewidth]{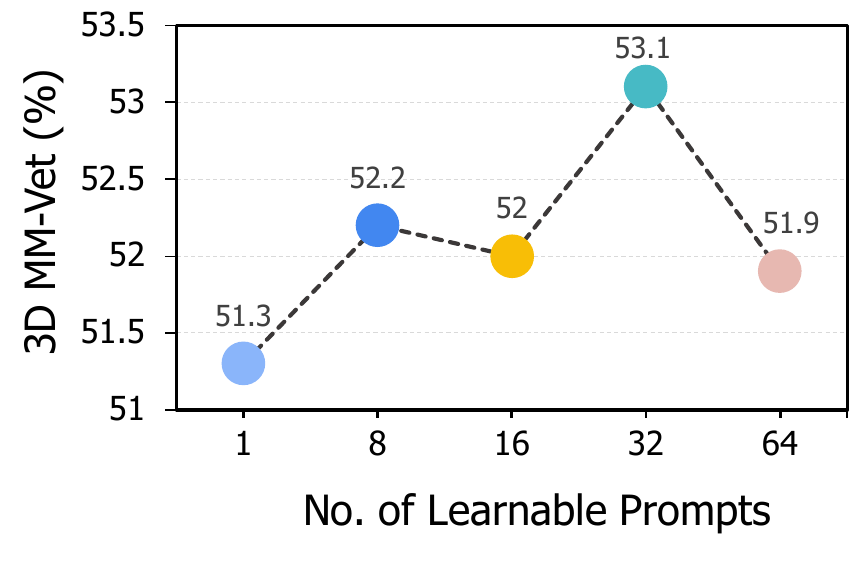}
\vspace{-7pt}
\captionof{figure}{\textbf{Ablation study on visual prompt number}. 
The performance of \shapellm-13B on 3D MM-Vet is reported.
}
\label{fig:number_prompts}
\vspace{-10pt}
\end{figure}

~
\cref{fig:number_prompts} shows the performance of \shapellm\ using different numbers of prompts, including 1, 8, 16, 32, and 64. 
This ablation study has shown that a different number of prompts leads to varied improvements, and the optimal setting is 32.
This observation is similar to VPT~\cite{VPT22} where the prompts used to modulate Transformer attention should be studied~\cite{UnifedPETuning21}.

\subsection{Multimodal Comprehension with \textbf{\shapellm}}
\noindent\textbf{Generative 3D Object Recognition \& Captioning}~
Following PointLLM~\cite{pointllm23}, we conduct generative 3D recognition and captioning experiments.
\cref{tab:generative_classification} shows 3D object classification overall accuracy (\%) and captioning performance evaluated by GPT-4 and data-driven metrics: Sentence-BERT (S-BERT)~\cite{SBERT19} and SimCSE~\cite{SimCSE21}.
It can be observed that \shapellm\ consistently outperforms other methods across all metrics, demonstrating robust recognition and instruction-following capabilities.

Note that similar to PointLLM's findings, we also notice that the 3D captioning performance evaluated by traditional metrics like BLEU-1~\cite{BLEU02}, ROUGE-L~\cite{ROUGH04}, and METEIOR~\cite{METEOR05} are highly unreliable in accurately revealing the response quality.
This is further demonstrated by human-oriented evaluation, such as the preference win rate comparison presented next.
\begin{table}[t!]
\centering
\caption{\textbf{Generative 3D recognition and captioning}. 
The accuracy (\%) averaged under the instruction-typed prompt ``What is this?'' and the completion-typed prompt ``This is an object of'' is reported.
}
\label{tab:generative_classification}
\resizebox{\linewidth}{!}{
\begin{tabular}{lcccccc}
\toprule[0.95pt]
\multirow{2}{*}[-0.5ex]{Method} & \multirow{2}{*}[-0.5ex]{Input} & \multicolumn{2}{c}{\textbf{Classification}} & \multicolumn{3}{c}{\textbf{Captioning}}\\
\cmidrule(lr){3-4}
\cmidrule(lr){5-7}
 &  & MN-40 & Objaverse & GPT-4 & S-BERT & SimCSE \\ 
\midrule[0.6pt]
InstructBLIP-7B~\cite{InstructBLIP23} & 1-View 2D Image & 25.51 & 43.50 & 45.34 & 47.41 & 48.48 \\
InstructBLIP-13B~\cite{InstructBLIP23} & 1-View 2D Image & 28.69 & 34.25 & 44.97 & 45.90 & 48.86\\
LLaVA-7B~\cite{LLaVA23} & 1-View 2D Image & 39.71 &  50.00 & 46.71 & 45.61 & 47.10\\
LLaVA-13B~\cite{LLaVA23} & 1-View 2D Image & 36.59 &  51.75 & 38.28 & 46.37 & 45.90\\ 
\midrule[0.6pt]
\multirow{2}{*}{3D-LLM~\cite{3DLLM23}} & \multirow{2}*{\shortstack{3D Object +\\ Multi-View 2D Image}} & \multirow{2}{*}{-} & \multirow{2}{*}{45.25} & \multirow{2}{*}{33.42} & \multirow{2}{*}{44.48} & \multirow{2}{*}{43.68} \\
& & & & & & \\
\midrule[0.6pt]
PointLLM-7B~\cite{pointllm23} & 3D Point Cloud & 52.63 &  53.00 & 44.85 & 47.47 & 48.55\\
PointLLM-13B~\cite{pointllm23} & 3D Point Cloud & 52.78 &  54.00 & 48.15 & 47.91 & 49.12 \\
\rowcolor{linecolor4}\textbf{\shapellm-7B} & 3D Point Cloud & \textbf{53.08} & \textbf{54.50} & \textbf{46.92} & \textbf{48.20} & \textbf{49.23}\\
\rowcolor{linecolor3}\textbf{\shapellm-13B} & 3D Point Cloud & \textbf{52.96} & \textbf{54.00} & \textbf{48.94} & \textbf{48.52} & \textbf{49.98}\\ 
\bottomrule[0.95pt]
\end{tabular}
}
\end{table}

\subsubsection{Human Win Rate Comparison}
\begin{figure}[h!]
\centering
\includegraphics[width=\linewidth]{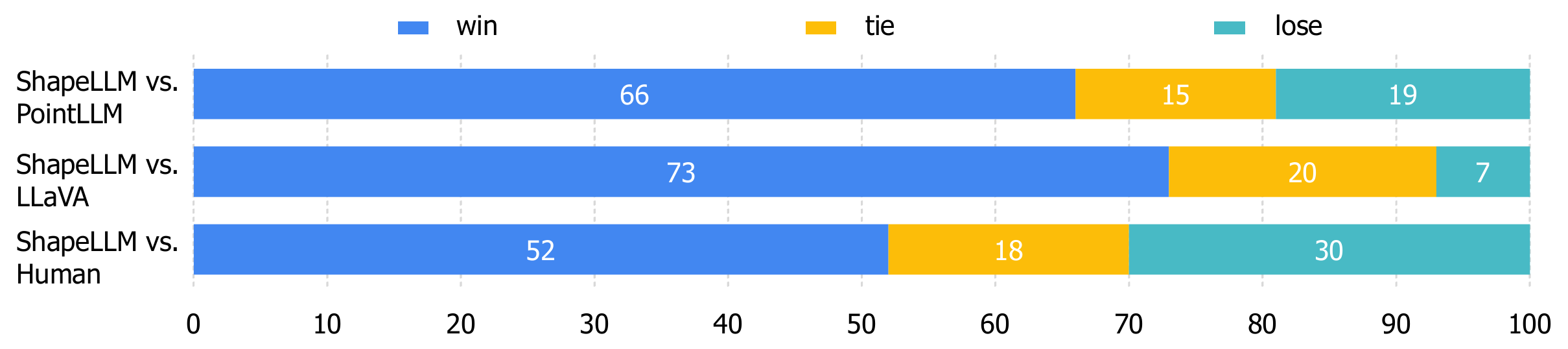}
\captionof{figure}{\textbf{Win rate comparison.}
}\label{fig:winrate}
\end{figure}

~
GPT-4~\cite{GPT4_23} is widely used as an evaluator in natural language and vision language processing, as seen in recent modern benchmarks like MM-Bench and MM-Vet. Recent studies~\cite{GPT4V3DEval24} have demonstrated that ChatGPT-based evaluation is more closely \textit{aligned with human preferences} compared to traditional metrics. With GPT4-turbo, the standard deviation of 3D MM-Vet is less than 0.1. To further verify the soundness of the models' response, we also conduct human evaluation and report the win rate in \cref{fig:winrate}, where \shapellm~demonstrates superior preference by humans.

\begin{figure}[t!]
\begin{center}
\centering
\includegraphics[width=\linewidth]{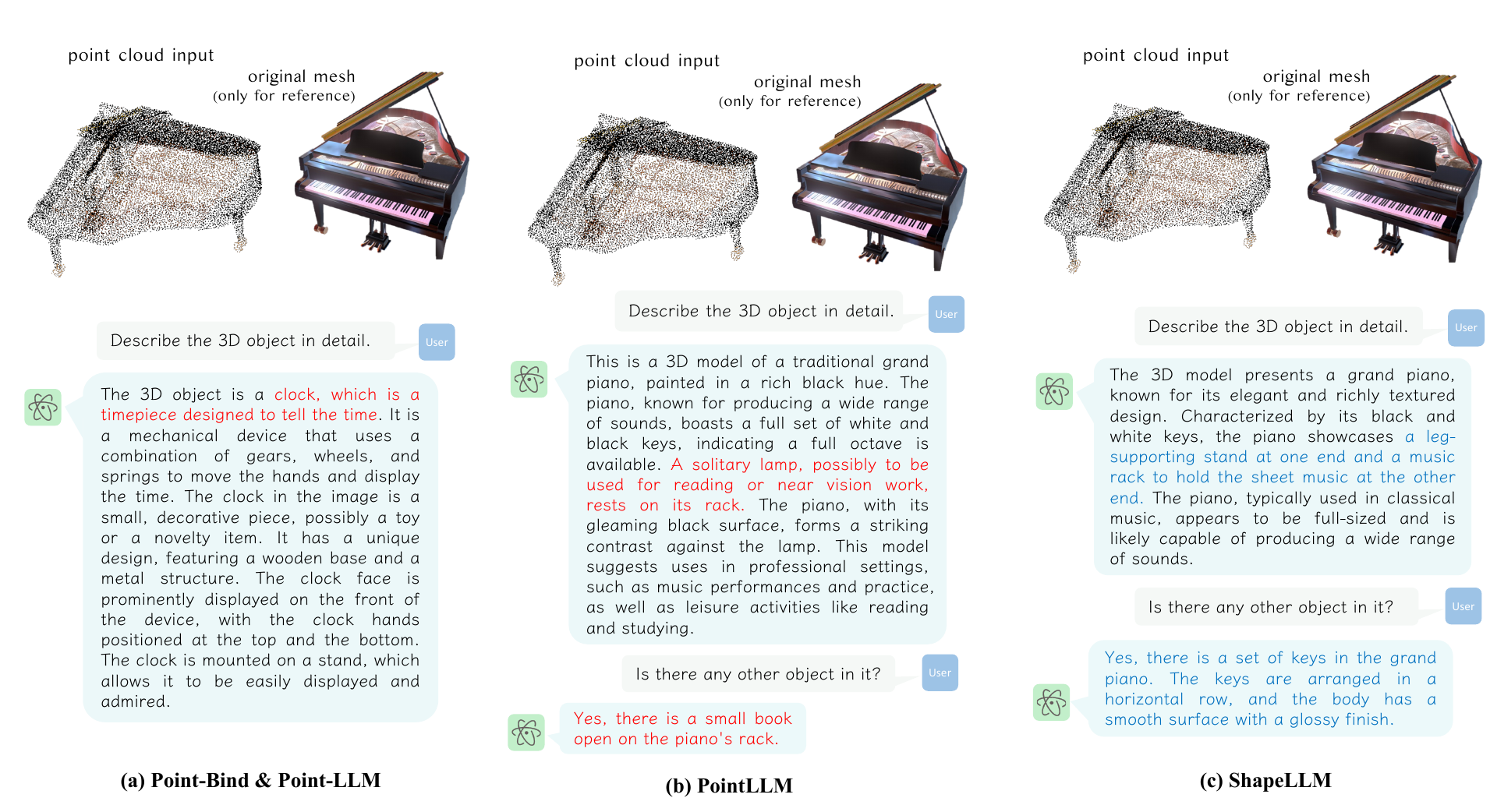}
\vspace{-15pt}
\captionof{figure}{\textbf{Qualitative comparison examples of visual hallucination}.}
\vspace{-25pt}
\label{fig:hallucination}
\end{center}
\end{figure}

\subsubsection{Visual Hallucination}
Visual hallucination is a well-known issue in LLMs and MLLMs that generate non-existent objects or identities from the input data, significantly compromising their multimodal comprehension capabilities~\cite{VLPHallucination23,GAVIE23,FDPO23,ZhouHall24} and may pose safety risks~\cite{BlindSafety17, ObjectHallucination18}. 
Recent research~\cite{HallucinationSurvey23} suggests that hallucination may stem from biases in training data, particularly within supervised fine-tuning data, or inappropriate generation strategies. In \cref{fig:hallucination}, we qualitatively demonstrate the illusion evaluation of \shapellm\ compared to other methods. We assess the model's ability to counteract illusions by prompting it with detailed captions and misleading questions.
The results in \cref{fig:hallucination} demonstrate that previous methods Point-Bind\&Point-LLM~\cite{pointbind23} and PointLLM~\cite{pointllm23} suffer from the problems of mis-recognition and mis-associating non-existing identities.
\vspace{-12pt}

\begin{table}[b]
\vspace{-12pt}
\setlength{\tabcolsep}{7pt}
  \caption{\textbf{Language-only baseline results on 3D MM-Vet}.}
  \vspace{-5pt}
  \label{tab:language_baseline}
  \centering
    \resizebox{0.74\linewidth}{!}{
    \begin{tabular}{lccccccc}
    \toprule[0.95pt]
        Method & \textbf{\texttt{Rec}} & \textbf{\texttt{Know}} & \textbf{\texttt{Gen}} & \textbf{\texttt{Spat}} & \textbf{\texttt{Emb}} & Total  \\ 
        \midrule[0.6pt]
        LLaMA2-7B-Chat & 11.8 & 10.6 & 22.1 & 14.6 & 25.8 & 16.2
        \\
        GPT-3.5-Turbo & 2.9 & 5.4 & 17.3 & 10.2 & 27.8 & 11.7\\
        GPT-4-Turbo & 1.7 & 3.6 & 16.1 & 6.6 & 26.0 & 9.8\\
        \midrule[0.6pt]
        \rowcolor{linecolor4}\textbf{\shapellm-7B} & \textbf{36.5} & \textbf{35.1} & \textbf{36.9} & \textbf{35.1} & \textbf{52.5} & \textbf{39.0}\\
        \rowcolor{linecolor3}\textbf{\shapellm-13B} & \textbf{46.8} & \textbf{53.0} & \textbf{53.9} & \textbf{45.3} & \textbf{68.4} & \textbf{53.1}\\
        \bottomrule[0.95pt]
    \end{tabular}
    }
\end{table}
\subsubsection{Language-Only Baseline}
\begin{figure}[t!]
\centering
\includegraphics[width=1\linewidth]{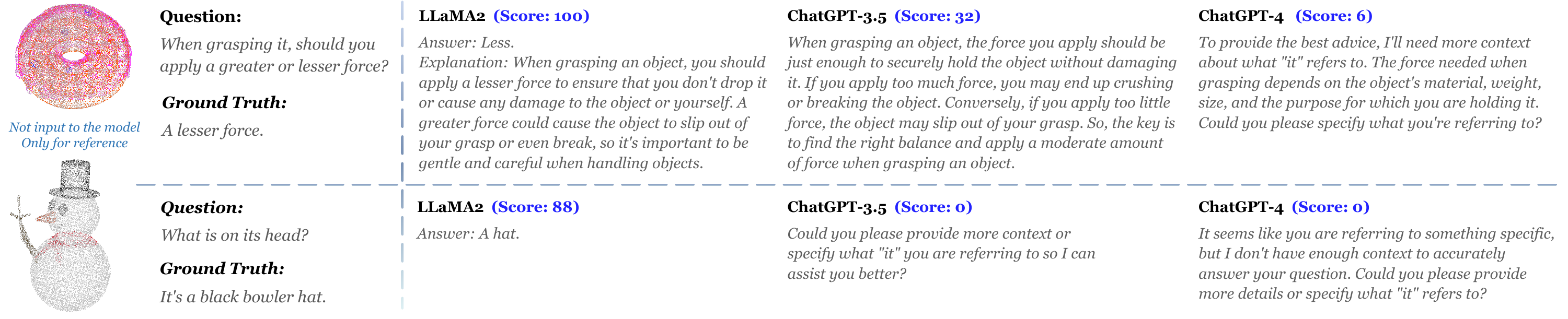}
\vspace{-4pt}
\captionof{figure}{\textbf{Language-only baseline analysis}.}
\label{fig:language_baseline}
\vspace{-15pt}
\end{figure}

To verify whether the capabilities of our model stem solely from large language models rather than from 3D perception abilities, we also conducted a Language-Only Baseline evaluation. This involved testing the performance of 3D MMVet on large language models such as ChatGPT~\cite{ChatGPT22} and LLaMA~\cite{LLaMA23}.
Results is shown in \cref{tab:language_baseline}. 
Our intentional usage of pronouns (e.g., ``it'') for referring to unknown objects in questions leads to poor performance,
and LLaMA hallucinates but scores higher than ChatGPT, which rejects to answer without specific inputs~\cref{fig:language_baseline}.
\vspace{-4pt}

\subsection{Representation Learning with \textbf{\recon}}
\vspace{-2pt}
\textbf{Linear SVM}~
Linear SVM evaluation~\cite{MarginSVM92,SLTheory98} can be used to evaluate the discriminative quality of pretrained features~\cite{BenchmarkSSL19}. The results on ModelNet40 are shown in \cref{tab:linear}. 
The results show that our \recon~outperforms both plain and hierachical Transformer methods by a clear margin.
\begin{table}[h!]
\vspace{-15pt}
\caption{\textbf{Linear SVM classification on ModelNet40}. Overall accuracy (\%) without voting is reported.} \label{tab:linear}
\vspace{-15pt}
\begin{center}
    \resizebox{0.52\linewidth}{!}{
    \begin{tabular}{lcc}
    \toprule[0.95pt]
    Method & Hierachical & \textbf{ModelNet40}\\
    \midrule[0.6pt]
    Point-BERT~\cite{PointBERT} & \xmark & 87.4\\
    PointMAE~\cite{PointMAE} & \xmark & 91.0\\
    PointM2AE~\cite{PointM2AE22} & \cmark & 92.9\\
    ACT~\cite{ACT23} & \xmark & 93.1\\
    I2P-MAE~\cite{I2PMAE23} & \cmark & 93.4\\
    {\scshape ReCon}~\cite{ReCon23} & \xmark & 93.4\\
    \rowcolor{linecolor3}\textbf{\recon} & \xmark & \textbf{93.6}\\
    \bottomrule[0.95pt]
    \end{tabular}
    }
\end{center}
\vspace{-20pt}
\end{table}

\noindent\textbf{Few-Shot 3D Object Recognition}~
Few-shot learning is critical for evaluating the representation transferring capabilities in data and training efficiency.
We conduct few-shot 3D object recognition experiments on ModelNet40, and the results are shown in \cref{tab:few-shot}. Our \recon~achieves state-of-the-art performance in all the benchmarks compared to previous works.
\begin{center}
\vspace{-15pt}
\begin{table}[h!]
    \centering
    \setlength\tabcolsep{3pt}
    \caption{\textbf{Few-shot classification results on ModelNet40}. Overall accuracy (\%) without voting is reported.}\label{tab:few-shot}
    \vspace{-5pt}
    \resizebox{0.78\linewidth}{!}{
    \begin{tabular}{lcccc}
    \toprule[0.95pt]
    \multirow{2}{*}[-0.5ex]{Method}& \multicolumn{2}{c}{\textbf{5-way}} & \multicolumn{2}{c}{\textbf{10-way}}\\
    \cmidrule(lr){2-3}\cmidrule(lr){4-5} & 10-shot & 20-shot & 10-shot & 20-shot\\
    \midrule[0.6pt]
    Transformer~\cite{AttentionIsAllYouNeed} & 87.8 $\pm$ 5.2& 93.3 $\pm$ 4.3 & 84.6 $\pm$ 5.5 & 89.4 $\pm$ 6.3\\
    Point-BERT~\cite{PointBERT} & 94.6 $\pm$ 3.1 & 96.3 $\pm$ 2.7 &  91.0 $\pm$ 5.4 & 92.7 $\pm$ 5.1\\
    Point-MAE~\cite{PointMAE} & 96.3 $\pm$ 2.5&97.8 $\pm$ 1.8 & 92.6 $\pm$ 4.1 & 95.0 $\pm$ 3.0\\
    Point-M2AE~\cite{PointM2AE22} & 96.8 $\pm$ 1.8&98.3 $\pm$ 1.4 & 92.3 $\pm$ 4.5 & 95.0 $\pm$ 3.0\\
    ACT~\cite{ACT23} & 96.8 $\pm$ 2.3 & 98.0 $\pm$ 1.4 & 93.3 $\pm$ 4.0 & 95.6 $\pm$ 2.8\\
    VPP~\cite{VPP23} & 96.9 $\pm$ 1.9 & 98.3 $\pm$ 1.5 & 93.0 $\pm$ 4.0 & 95.4 $\pm$ 3.1\\
    {\scshape ReCon}~\cite{ReCon23} & 97.3 $\pm$ 1.9 & 98.9 $\pm$ 1.2 & 93.3 $\pm$ 3.9 & 95.8 $\pm$ 3.0 \\
    PointGPT~\cite{pointgpt23} & 98.0 $\pm$ 1.9 & 99.0 $\pm$ 1.0 & 94.1 $\pm$ 3.3 & 96.1 $\pm$ 2.8 \\
    \rowcolor{linecolor3}\textbf{\recon} & \textbf{98.0 $\pm$ 2.3} & \textbf{99.5 $\pm$ 0.8} & \textbf{94.5 $\pm$ 4.1} & \textbf{96.5 $\pm$ 3.0} \\
    \bottomrule[0.95pt]
    \end{tabular}
    }
\end{table}
\vspace{-10pt}
\end{center}

\subsubsection{Multi-view Alignment visualization analysis.}
\begin{figure}[h!]
\centering
\includegraphics[width=0.64\linewidth]{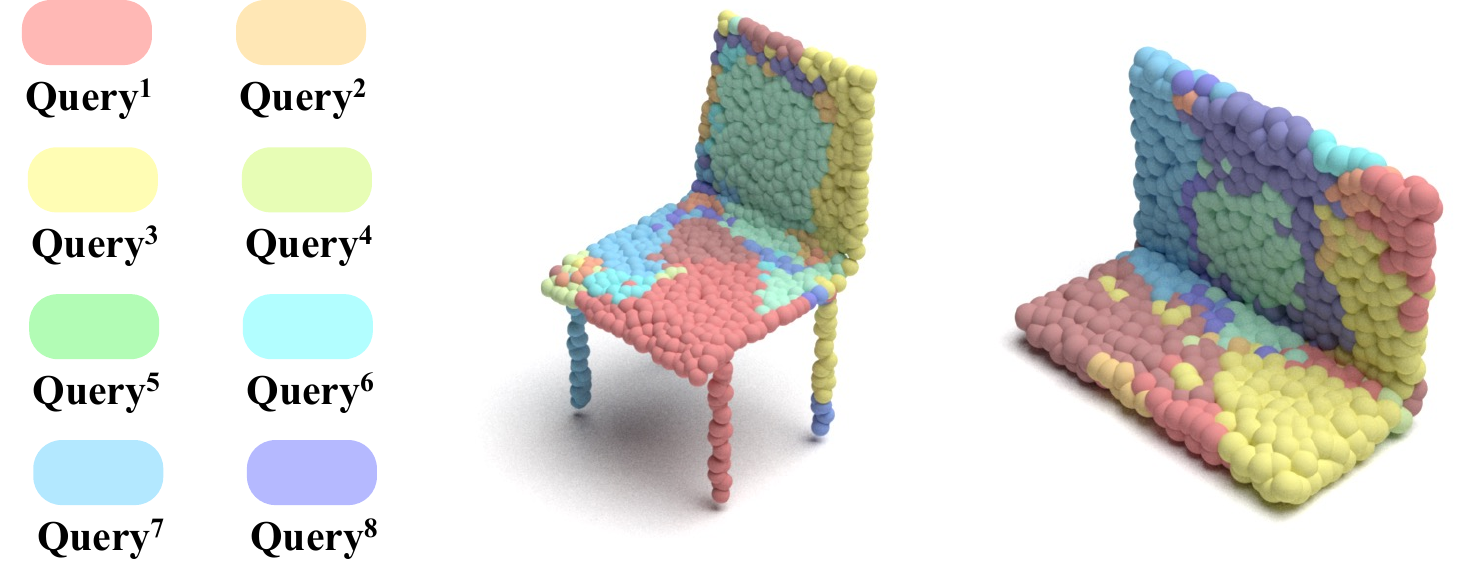}
\captionof{figure}{\textbf{Visualization of multi-view query results.} The distinct colors serve to denote distinct image queries.}
\label{fig:query}
\end{figure}

\cref{fig:query} illustrates the visualization of the attention maps in the last cross-attention layer, documenting the image query to which each local patch primarily attends.
It provides evidence that multi-view alignment achieves geometrically informed spatial understanding, which may implicitly encompass the estimation of the object pose and a more profound knowledge of 3D spatial relationships.

\subsubsection{ReCon++ Key Modifications Analysis}
We conduct an ablation study on the two key modifications of \recon, namely scaling up and multi-view alignment, and the results are presented in \cref{tab:recon_ablation}.
The results demonstrate that:
i) scaling up 3D representation is critical for both 3D representation learning, and stonger 3D representation understanding brought by \recon~consistently yields better 3D multimodal comprehension;
ii) the proposed multi-view distillation further leads to significant improvement.
\begin{table}[h!]
\setlength{\tabcolsep}{7pt}
\caption{\textbf{Ablation study on scaling and multi-view alignment}.} \label{tab:recon_ablation}
\vspace{-5pt}
\begin{center}
    \resizebox{0.62\linewidth}{!}{
    \begin{tabular}{cccc}
    \toprule[0.95pt]
    scaling & multi-view & \textbf{Zero-Shot} & \textbf{3D MM-Vet}\\
    \midrule[0.6pt]
    \xmark & \xmark & 6.7 & 15.8\\
    \xmark & \cmark & 10.3 & 21.9\\
    \cmark & \xmark & 51.5 & 48.2\\
    \rowcolor{linecolor3}\cmark & \cmark & \textbf{53.7} & \textbf{53.1}\\
    \bottomrule[0.6pt]
    \end{tabular}
    }
\end{center}
\vspace{-5pt}
\end{table}

\section{Additional Information about 3D MM-vet}\label{app:mmvet_detail}
\subsection{Evaluation System}
Unlike classification or regression tasks, language generation tasks lack a definitive ground truth that can comprehensively cover diverse real-life scenarios. Therefore, evaluating the alignment of model-generated results with the question and assessing their appropriateness becomes a challenging problem, requiring a reasonable quantitative score. Fortunately, we have observed the recent surge in the popularity of GPT, providing us with a dependable tool for conducting open-ended evaluations.

 To enhance the performance of GPT, we employ a few-shot style in-context prompt. This involves feeding GPT with prompts from evaluative examples and instructing it to generate scores. 
 Specifically, we present prompts to obtain a score ranging from 0 to 1, indicating the degree of similarity between the model-generated answers and the ground truths we provided. 
 When implementing this approach, we observed that results generated multiple times may vary a lot. 
 To address it, we apply the same evaluation setting to a single answer for $K$ iterations, obtaining the average result as the final score for a precise answer. 
 The score of an answer $S_a$ and the total score $S_t$ of answer set $A$ are calculated by:
 $$S_a = \frac{\sum\limits_{i=1}^{K} s_{a_i}}{K}, \quad S_t=\frac{\sum\limits_{a\in A} S_{a}}{N}.$$ 
 Here we set $K=5$, and $s_{a_i}$ is the score of the $i_{th}$ test of answer $a$. The average score for a specific capability is the sum of scores in category $C$ answer set $A_C$:
 $$S_c = \frac{\sum\limits_{a\in A_C} S_a}{N_c},$$
 where $N_c$ is the number of answers in each capability set.

 To mitigate excessive standard deviation, we opt for GPT-4 in a series of $K$ scoring rounds to get rounds of outputs with a standard deviation below 0.1. 
 This choice is motivated by the enhanced stability offered by GPT-4~\cite{GPT4_23}, in contrast to GPT-3.5~\cite{ChatGPT22}, where scores across different rounds exhibit significant variability.
 
\subsection{Analysis}
\begin{figure}[t!]
\centering
\includegraphics[width=0.5\linewidth]{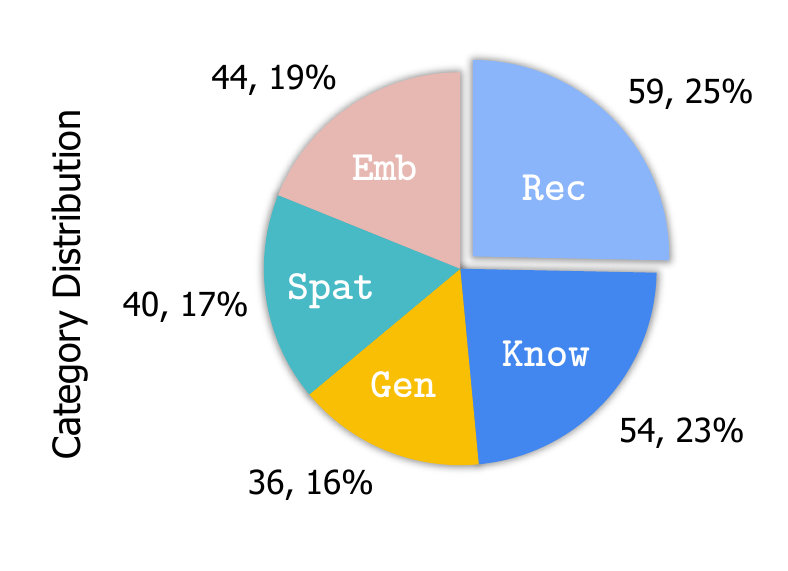}
\vspace{-12pt}
\captionof{figure}{\textbf{The number of diverse questions} of \textit{core VL capabilities} on 3D MM-Vet.
  \textbf{\texttt{Rec}}: General Visual Recognition, \textbf{\texttt{Know}}: Knowledge, \textbf{\texttt{Gen}}: Language Generation, \textbf{\texttt{Spat}}: Spatial Awareness, \textbf{\texttt{Emb}}: Embodied Interaction.}
\label{fig:num}
\end{figure}

The 3D MM-Vet evaluation benchmark consists of 5 different categories of questions. In \cref{fig:num} we report the distribution of problem categories.

\begin{figure}[t!]
\centering
\includegraphics[width=1\linewidth]{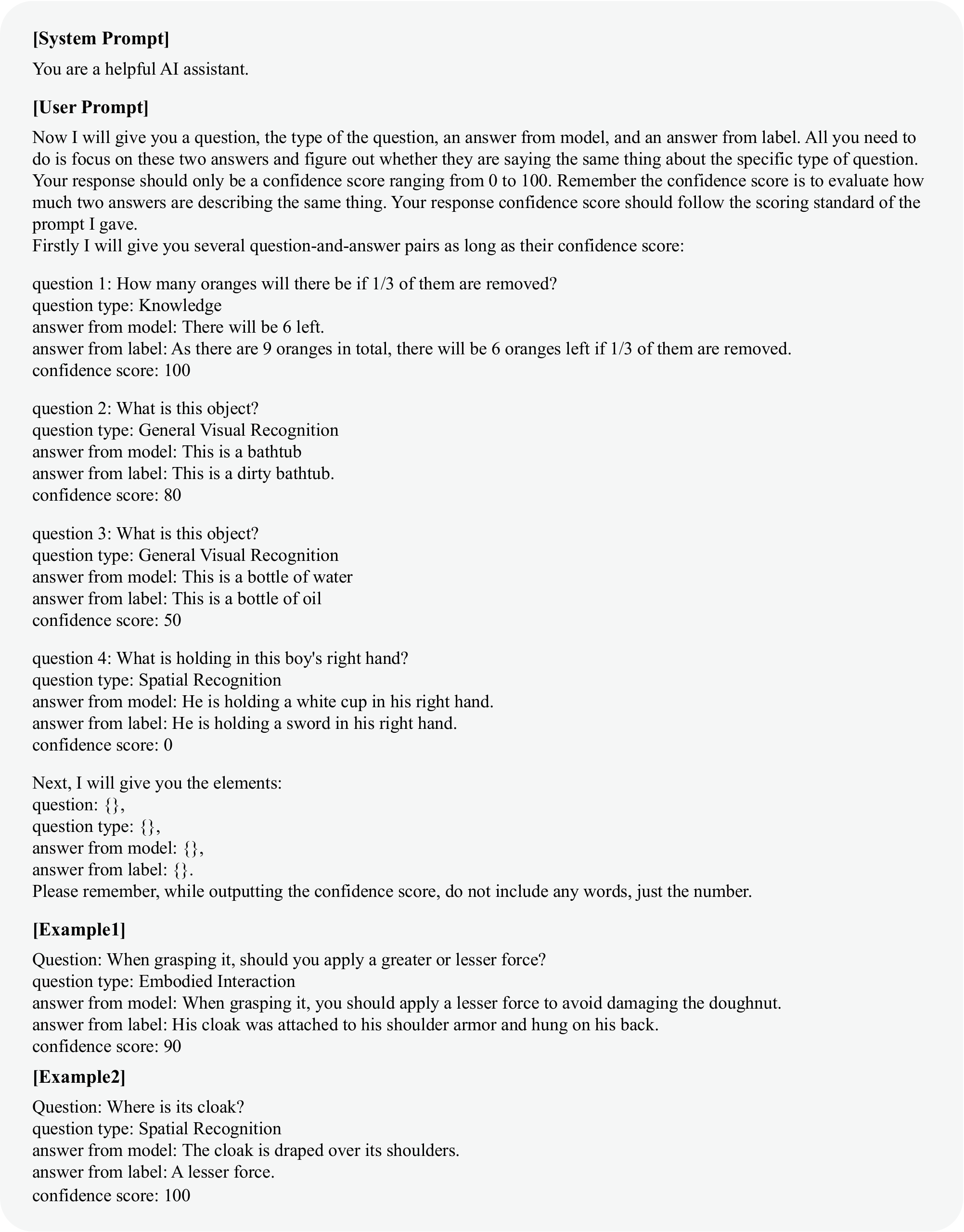}
\captionof{figure}{\textbf{GPT-4 evaluation template and examples of our 3D MM-Vet benchmark.}
}
\vspace{-5pt}
\label{fig:mmvet_eval_temp}
\end{figure}

\begin{figure}[!ht]
\centering
\includegraphics[width=\linewidth]{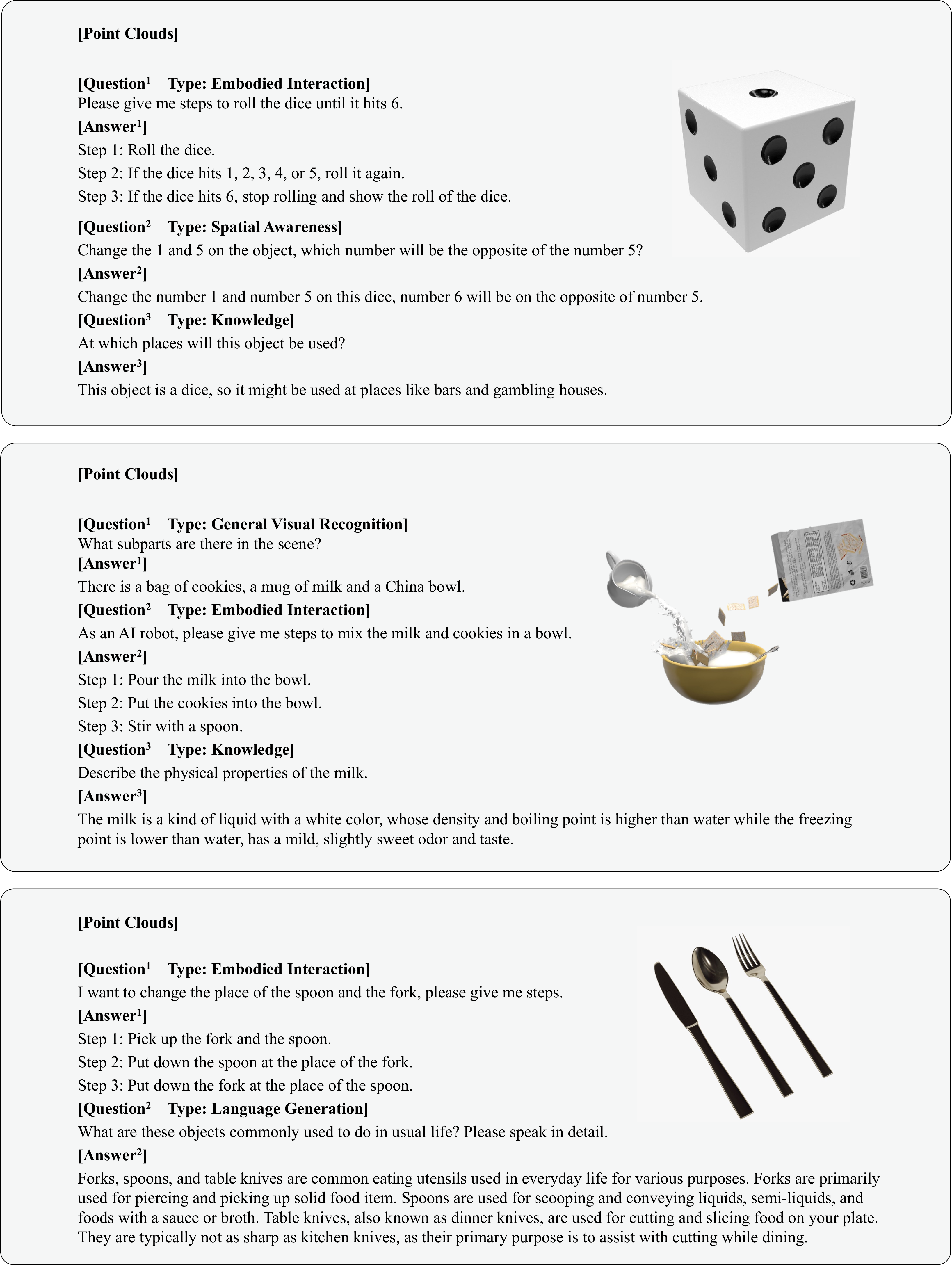}
\captionof{figure}{\textbf{Additional Visualization example of 3D MM-Vet Q\&A pairs.}}
\label{fig:show}
\end{figure}

\begin{figure}[!ht]
\centering
\includegraphics[width=\linewidth]{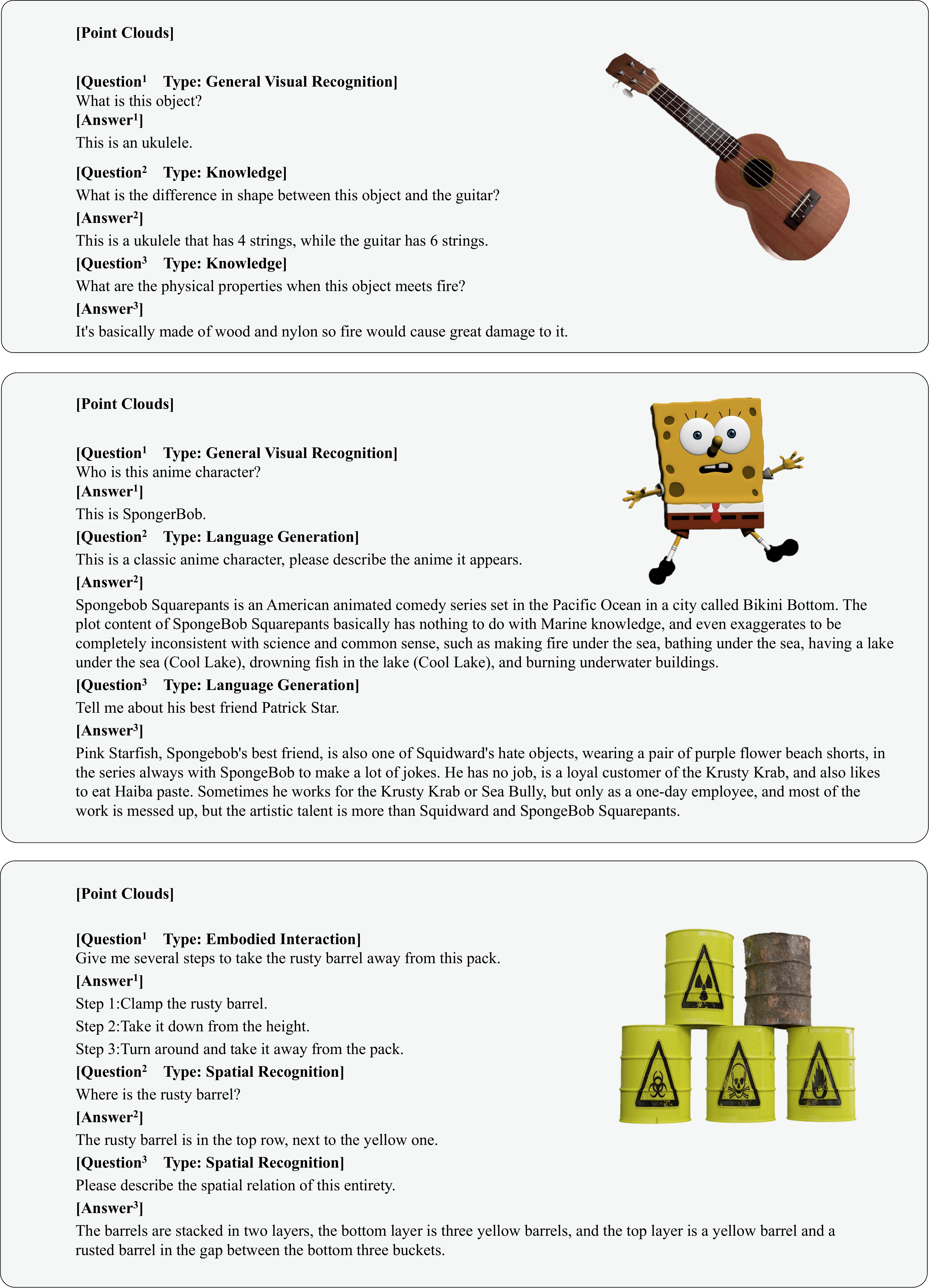}
\captionof{figure}{\textbf{Additional Visualization example of 3D MM-Vet Q\&A pairs.}}
\label{fig:show2}
\end{figure}

The knowledge and General Visual Recognition parts contain multiple subparts that comprehensively evaluate these capacities and thus hold higher proportions. 
\cref{fig:mmvet_eval_temp} shows an example of how we prompt GPT-4 for 3D MM-Vet evaluation. 
\cref{fig:show} and \cref{fig:show2} illustrate additional examples of 3D MM-Vet Q\&As.
\begin{table}[t!]
\setlength{\tabcolsep}{6pt}
\caption{\textbf{Sample categories of 3D MM-Vet data}.} 
\vspace{-17pt}
\label{tab:mmvet_category}
\begin{center}
    \resizebox{0.74\linewidth}{!}{
    \begin{tabular}{lccccc}
    \toprule[0.95pt]
    \textbf{Category} & Characters & Life & Art & Architecture & Animals\\
    \midrule[0.6pt]
    \textbf{Number}& 11 & 16 & 10 & 13 & 9\\
    \bottomrule[0.6pt]
    \end{tabular}
    }
\end{center}
\vspace{-10pt}
\end{table}

\vspace{-3pt}
\subsection{ChatGPT Costs}
In constructing the Supervised Finetuning dataset for ShapeLLM and conducting inference on 3D MMVet using the GPT-4 or GPT-4V API, we have roughly estimated the costs. For ShapeLLM's training data, which contains over 50k Q\&A pairs, with each request yielding 5-6 Q\&A pairs, the estimated cost is approximately \$900. As for inference on 3D MMVet, with only 232 samples and averaging five requests per sample, the cost is estimated to be around \$12.
\vspace{-3pt}

\begin{table}[b!]
\caption{\textbf{\recon\ model variants}, which follow ViT~\cite{ViT}.} \label{tab:recon_variants}
\vspace{-15pt}
\begin{center}
    \resizebox{0.6\linewidth}{!}{
    \begin{tabular}{lcccc}
    \toprule[0.95pt]
    Model & Layers & Hidden size & MLP size & Heads\\
    \midrule[0.6pt]
    \recon-S & 12 & 384 & 1536 & 6\\
    \recon-B & 12 & 768 & 3072 & 12 \\
    \recon-L & 24 & 1024 & 4096 & 16 \\
    \bottomrule[0.95pt]
    \end{tabular}
    }
\end{center}
\end{table}
\section{Implementation details}
\vspace{-2pt}
\noindent\textbf{\recon~}
Following the standard ViT~\cite{ViT} architecture, we design four different model structures consistent with prior work~\cite{ReCon23,OpenShape23,Uni3D23}. The model parameters are shown in \cref{tab:recon_variants}. Following OpenShape~\cite{OpenShape23}, we employ four datasets as pretraining data, namely Objaverse~\cite{objaverse23}, ShapeNet~\cite{ShapeNet15}, ABO~\cite{ABO22}, and 3D-FUTURE~\cite{3d-future21}. Each point cloud sample has a size of 10,000$\times$6, where the first three dimensions represent $xyz$ coordinates, and the latter three dimensions represent $rgb$ values. 

\begin{table}[t!]
\caption{\textbf{Ablation study on mask type \& stop gradient}. transfer: fine-tuned 3D recognition on ScanObjectNN~\cite{ScanObjectNN19}. zero-shot: zero-shot 3D recognition on Objaverse-LVIS~\cite{objaverse23}. All experiments are conducted on \recon-L and \shapellm-13B.} \label{tab:mask_type}
\vspace{-20pt}
\begin{center}
    \resizebox{0.68\linewidth}{!}{
    \begin{tabular}{ccccc}
    \toprule[0.95pt]
    Mask Type & Stop Grad & Fine-Tune & Zero-Shot & \textbf{3D MM-Vet}\\
    \midrule[0.6pt]
    Random & \cmark & 92.5 & 52.8 & \textbf{53.1} \\
    Random & \xmark & 93.6 & \textbf{53.7} & 52.9 \\
    Causal & \cmark & \textbf{95.3} & 49.8 & 50.7 \\
    Causal & \xmark & 92.8 & 51.0 & 51.6 \\
    \bottomrule[0.6pt]
    \end{tabular}
    }
\end{center}
\vspace{-25pt}
\end{table}
Regarding the masked modeling strategy, we experimented with both random masking strategies and the latest causal masking strategy. Using causal masking as initialization significantly improves transfer learning capability, as shown in the ablation experiments in \cref{tab:mask_type}. Specifically, the point encoder of \shapellm~still employs the original local-guided stop-gradient strategy~\cite{ReCon23}. Additionally, to enhance global classification and retrieval capabilities, we backpropagate gradients from the global branch to the local branch in open vocabulary zero-shot experiments, as demonstrated in the ablation experiments in \cref{tab:mask_type}.

\vspace{2pt}
\noindent\textbf{\shapellm}~
We use the LLaMA model~\cite{LLaMA23} as our LLM backbone, with the 7B and 13B Vicuna-1.1~\cite{Vicuna23} checkpoint as the default settings. We partitioned the point clouds into 512 patches using furthest point sampling and k-nearest neighbors.
Similar to other MLLMs~\cite{LLaVA23,pointllm23,DreamLLM23}, we employ a 3-layer MLP with GELU~\cite{GLEU16} as the projector, with hidden layer sizes of 1,024 and 2,048, respectively. Note that different projector parameters are utilized for absolute positional encoding, local, and global features. Through training the projector, multi-scale and multi-mode features of the point cloud are mapped into the text space. After adding two special tokens, the vocabulary size becomes 32,003.

\vspace{-10pt}
\section{Training details}
\vspace{-5pt}
\noindent\textbf{\recon~}
Due to the sensitivity of the Chamfer Distance~\cite{ChamferDistance17} loss to accuracy, all experiments were conducted at FP32 precision using 8 × 80G A800 GPUS. We still use the strategy of \textit{contrast with reconstruct}~\cite{ReCon23}. To save parameter tuning time and improve performance, we divide the training process into two stages: the reconstruction stage based on mask modeling and the cross-modal alignment stage based on knowledge distillation. For transfer learning classification tasks, \recon~is pretrained on 1,024 points. For zero-shot tasks and \shapellm~tasks, \recon~is pretrained on 10,000 points. Further details regarding the hyperparameter settings are documented in \cref{tab:hyper_params}.
\begin{table}[t!]
\caption{\textbf{Training recipes for \recon~and \shapellm}.}
\label{tab:hyper_params}
\centering
\resizebox{\linewidth}{!}{
\begin{tabular}{lccccc}
 & \multicolumn{3}{c}{\textbf{\recon}} & \multicolumn{2}{c}{\textbf{\shapellm}}\\
 \toprule[0.95pt]
 Config & HyBrid/Ensembled & ScanObjectNN & ModelNet & Cap3D & LVIS/GAPartNet\\
 \midrule[0.6pt]
 optimizer & AdamW & AdamW & AdamW & AdamW & AdamW\\
 learning rate & 5e-5 & 2e-5 & 1e-5 & 2e-3 & 2e-5 \\
 weight decay & 5e-2 & 5e-2 & 5e-2 & - & - \\
 learning rate scheduler & cosine & cosine & cosine & cosine & cosine \\
 training epochs & 300 & 300 & 300 & 3 & 1 \\
 warmup epochs & 10 & 10 & 10 & 0.03 & 0.03 \\
 batch size & 512 & 32 & 32 & 256 & 128 \\
 drop path rate & 0.1 & 0.2 & 0.2 & - & - \\
 \midrule[0.6pt]
 number of points & 1024/10000 & 2048 & 1024/10000 & 10000 & 10000 \\
 number of point patches & 64/512 & 128 & 64/512 & 512 & 512 \\
 point patch size & 32 & 32 & 32 & 32 & 32 \\
 \midrule[0.6pt]
 augmentation & Rot\&Scale\&Trans & Rot & Scale\&Trans & - & - \\
 \midrule[0.6pt]
 GPU device & 8$\times$A800 & 1$\times$A800 & 1$\times$A800 & 8$\times$A800 & 8$\times$A800 \\
\bottomrule[0.95pt]
\end{tabular}
}
\end{table}

\vspace{2pt}
\noindent\textbf{\shapellm~}
All experiments were conducted using 8 $\times$ 80G A800 GPUs with a BF16 data type. During the multimodal alignment stage, we train our model for one epoch with a batch size 256 and a learning rate 2e-3. During the instruction tuning stage, we train our model for one epoch with a batch size of 128 and a learning rate 2e-5. Throughout both stages, we employ flash-attention~\cite{flashattention22}, the AdamW~\cite{AdamW19} optimizer, and a cosine learning rate scheduler~\cite{CosineLRSGDR}. For the entire training process, the 7B and 13B models require approximately 10 and 20 hours, respectively. Further hyper-parameters are documented in \cref{tab:hyper_params}.

\section{Additional Related Work}\label{app:related_work}
\subsection{3D Representation Learning}
Various methods have been proposed to tackle 3D Representation Learning, including point-based~\cite{PointNet,PointNet++}, voxel-based~\cite{voxelnet15}, and multiview-based approaches~\cite{MVCNN3D15,MVTN}. 
Point-based methods~\cite{PointNext,PointTrans21} have gained prominence in object classification~\cite{ModelNet15,ScanObjectNN19} due to their sparsity yet geometry-informative representation. On the other hand, voxel-based methods~\cite{voxelrcnn21,SyncSpecCNN17,VPP23} offer dense representation and translation invariance, leading to a remarkable performance in object detection~\cite{ScanNet17} and segmentation~\cite{ShapeNetPart16, S3DIS16}.
The evolution of attention mechanisms~\cite{AttentionIsAllYouNeed} has also contributed to the development of effective representations for downstream tasks, as exemplified by the emergence of 3D Transformers~\cite{PointTrans21,groupfree21, voxeltransformer21}. Notably, 3D self-supervised representation learning has garnered significant attention in recent studies. PointContrast~\cite{PointContrast20} utilizes contrastive learning across different views to acquire discriminative 3D scene representations. 
Innovations such as Point-BERT~\cite{PointBERT} and Point-MAE~\cite{PointMAE} introduce masked modeling~\cite{MAE,BERT} pretraining into the 3D domain. 
ACT~\cite{ACT23} pioneers cross-modal geometry understanding through 2D or language foundation models such as CLIP~\cite{CLIP} or BERT~\cite{BERT}. 
Following ACT, {\scshape ReCon}~\cite{ReCon23} further proposes a learning paradigm that unifies generative and contrastive learning, which can be applied to both single-modal or cross-modal settings.
Additionally, leveraging foundation vision-language models like CLIP~\cite{ACT23,CLIP} has spurred the exploration of a new direction in open-world 3D representation learning. 
This line of work seeks to extend the applicability and adaptability of 3D representations in diverse and open-world/vocabulary scenarios by distilling the open-world knowledge within foundation models~\cite{OpenScene23,CLIPFO3D23,PLA23,Lowis3D23,OVIR3D23,PointGCC23}, with which it is now possible to perceive the 3D physical scenes using human languages.

\section{Future Works}
\shapellm~has made significant progress in advancing 3D shape understanding and embodied perception through MLLMs. Future endeavors aim to scale up embodied understanding training using datasets larger than GAPartNet, potentially leading to open-vocabulary part-level comprehension, including 6-DoF pose estimation. 
To this end, the first possibility is to empower the training data and benchmarking data with more advanced MLLMs such as GPT4-o~\cite{GPT4o24}, which are more human-aligned intelligent agents~\cite{GPT4V3DEval24,DreamBenchPlus24}.
Excitingly, there is a vision to establish a unified framework capable of comprehending not only 3D shapes but also entire 3D scenes. To enhance real-world applications on robots, a promising approach involves a robotics co-design that effectively connects 3D representations with downstream language-based tasks~\cite{VoxPoser23,VIMA23,sugar24}. Additionally, addressing efficiency for real-time deployment is crucial, emphasizing techniques like model compression~\cite{DGMS23,CompressLLMs23,SelfKDPAMI22,PointDistiller22,ReKo23}.


\end{document}